# A Discriminative Approach to Bayesian Filtering with Applications to Human Neural Decoding

by

Michael C. Burkhart

B.Sc.'s, Honors Mathematics, Honors Statistics, and Economics, Purdue University, 2011

M.Sc., Mathematics, Rutgers University, 2013

A dissertation submitted in partial fulfillment of the
requirements for the Degree of Doctor of Philosophy
in the Division of Applied Mathematics at Brown University

Providence, Rhode Island

May 2019

This dissertation by Michael C. Burkhart is accepted in its present form by the Division of Applied Mathematics as satisfying the dissertation requirement for the degree of Doctor of Philosophy.

Date _______________          _________________________________________
                                              Matthew T. Harrison, Director

Recommended to the Graduate Council

Date _______________          _________________________________________
                                              Basilis Gidas, Reader

Date _______________          _________________________________________
                                              Jerome Darbon, Reader

Approved by the Graduate Council

Date _______________          _________________________________________
                                              Andrew G. Campbell
                                              Dean of the Graduate School

# VITA

## Education

Brown University    Ph.D. Applied Mathematics    2013–2018

Rutgers University    M.Sc. Mathematics    2011–2013

Purdue University    B.Sc.'s Honors Mathematics, Honors Statistics, & Economics    2007–2011

## Research Experience

BrainGate Clinical Trial    Ph.D. candidate advised by Prof. Matthew T. Harrison    2014–2018

- developed and implemented novel nonlinear filters for online neural decoding (Matlab and Python)

- enabled participants with quadriplegia to communicate and interact with their environments in real time using mental imagery alone

- experimented with Bayesian solutions to provide robustness against common nonstationarities for online decoding in Brain Computer Interfaces

Spotify USA, Inc.   Data Research Intern   2017

- implemented online stochastic variational inference for topic models (Latent Dirichlet Allocation & Hierarchical Dirichlet Processes) on playlist data

- scaled training to 500M playlists using Google Cloud Platform's BigQuery (SQL) and cloudML (for parallel cloud computing)

Brown–Kobe Summer School   Team Leader, High–Performance Computing   2016

- designed and supervised a project to create a parallelized particle filter for neural decoding

- taught topics in Bayesian filtering and Tensorflow/Cython (compiled Python) to graduate students from Brown and Kobe Universities

Argonne
National Lab

Graduate Research Aide to Dr. Victor M.
Zavala

2013

- propagated variance in a multistep
  Gaussian process prediction model to
  better estimate prediction error
  (Matlab and R)

- used Monte Carlo Expectation
  Maximization to learn
  hyperparameters

## Publications

- D. Brandman, M. Burkhart, J. Kelemen, B. Franco, M. Harrison*, & L. Hochberg*. Robust closed-loop control of a cursor in a person with tetraplegia using Gaussian process regression, *Neural Computation* 30 (2018).

- D. Brandman, T. Hosman, J. Saab, M. Burkhart, B. Shanahan, J. Ciancibello, et al. Rapid calibration of an intracortical brain computer interface for people with tetraplegia. *Journal of Neural Engineering* 15 (2018).

- M. Burkhart, Y. Heo, and V. Zavala. Measurement and verification of building systems under uncertain data: A Gaussian process modeling approach, *Energy and Buildings* 75 (2014).

## Pre-print

- M. Burkhart*, D. Brandman*, C. Vargas-Irwin, & M. Harrison. The discriminative Kalman filter for nonlinear and non-Gaussian sequential Bayesian filtering.

# Invited Talks

- M. Burkhart, D. Brandman, C. Vargas-Irwin, & M. Harrison. Nonparametric discriminative filtering for neural decoding. 2016 ICSA Applied Statistics Symposium. Atlanta, GA, 2016.

- D. Knott, U. Walther, & M. Burkhart. Finding the Non-reconstructible Locus. SIAM Conference on Applied Algebraic Geometry. Raleigh, NC, 2011.

# Conference Presentations

- M. Burkhart, D. Brandman, & M. Harrison. The discriminative Kalman filter for nonlinear and non-Gaussian sequential Bayesian filtering. The 31st New England Statistics Symposium, Storrs, CT, 2017.

- D. Brandman, M. Burkhart, ..., M. Harrison, & L. Hochberg. Noise-robust closed-loop neural decoding using an intracortical brain computer interface in a person with paralysis. Society for Neuroscience, Washington, DC, 2017.

- —. Closed loop intracortical brain computer interface cursor control in people using a continuously updating Gaussian process decoder. Society for Neuroscience, San Diego, CA, 2016.

- —. Closed Loop Intracortical Brain Computer Interface Control in a Person with ALS Using a Filtered Gaussian Process Decoder. American Neurological Assoc. Annual Meeting, Baltimore, MD, 2016.

- —. Intracortical brain computer interface control using Gaussian processes. Dalhousie University Surgery Research Day, Halifax, NS, 2016.

- —. Closed loop intracortical brain computer interface control using Gaussian processes in a nonlinear, discriminative version of the Kalman filter. 9th World Congress for Neurorehabilitation, Philadelphia, PA, 2016.

# Community Involvement

| | | |
|---|---|---|
| Brown SIAM Student Chapter | Vice President, Chapter Records | 2016–2017 |

- organized events within the applied math community

Interdepartmental Liaison Officer     2015–2016

- founding officer of Brown's student chapter

Rutgers Math Department     Member, Graduate Liaison Committee     2012–2013

- responsible for expressing graduate student concerns to department administration

- helped to orient new graduate students in the department

Purdue Student Publishing Foundation     Member, Corporate Board of Directors     2009–2011

- oversaw the *Exponent*, Purdue's Independent Daily Student Newspaper

- served on Editor in Chief Search Committee; interviewed candidates and helped a diverse committee of professors, community members, and students come to a consensus

|                           | Chairman, Finance Committee | 2010–2011 |

- oversaw >$1 million annual budget, set student and faculty salaries, approved capital expenditures

- worked to ensure the paper's long-term financial stability with investment accounts

|                           | Investigative Justice | 2007–2008 |

Purdue Student
Supreme Court

- heard parking and moving violation appeals to cases that occurred on university property

- served on a grade appeals committee

## Awards and Honors

| | |
|---|---|
| Brown Institute for Brain Science Graduate Research Award | 2016 |
| Brown International Travel Award (Arequipa, Peru) | 2016 |
| Brown Conference Travel Award (Arequipa, Peru) | 2016 |
| Brown-IMPA Partnership Travel Award (Rio de Janeiro, Brazil) | 2015 |
| Brown-Kobe Exchange Travel Award (Kobe, Japan) | 2014, 2016 |
| Rutgers Graduate Assistantship in Areas of National Need | 2012 |
| Valedictorian, Roncalli High School, Indianapolis, IN | 2007 |
| National Merit Scholar Finalist | 2007 |
| Eagle Scout | 2003 |

# DEDICATION

This thesis is dedicated to BrainGate volunteer "T9", who upon learning I studied statistics, kindly chided me that statisticians had predicted he would already be dead. I hope he would be pleased with our work. May he rest in peace.

# PREFACE

Suppose there is some underlying process $Z_1, \ldots, Z_T$ about which we are very interested, but that we cannot observe. Instead, we are sequentially presented with observations or measurements $X_{1:T}$ related to $Z_{1:T}$. At each time step $1 \le t \le T$, *filtering* is the process by which we use the observations $X_1, \ldots, X_t$ to form our best guess for the current hidden state $Z_t$.

Under the *Bayesian* approach to filtering, $X_{1:T}, Z_{1:T}$ are endowed with a joint probability distribution. The process by which we generate $X_{1:T}, Z_{1:T}$ can be described using the following graphical model. This particular form is variously known as a dynamic state-space or hidden Markov model.

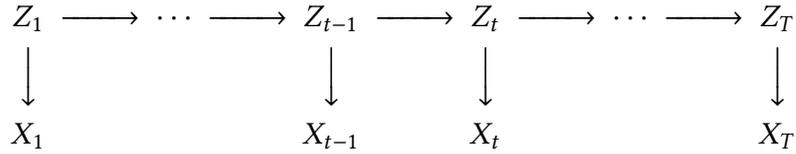

We start by drawing $Z_1$ from its marginal distribution $p(z_1)$. We then generate an observation $X_1$ that depends only on $Z_1$ using the distribution $p(x_1|z_1)$. At each subsequent time step $t$, we draw $Z_t$ from the distribution $p(z_t|z_{t-1})$ and $X_t$ from the distribution $p(x_t|z_t)$. These two conditional distributions are very important and characterize the generative process up to initialization of $Z_1$. The first, $p(z_t|z_{t-1})$, relates the state at time $t$ to the state at time $t-1$ and is often called the state or prediction model. The second, $p(x_t|z_t)$, relates the current observation to the current state and is called the measurement or observation model, or the likelihood. The Bayesian solution to the filtering problem returns the conditional distribution of $Z_t$ given that $X_1, \ldots, X_t$ have been observed to be $x_1, \ldots, x_t$. We refer to this distribution $p(z_t|x_{1:t})$ as the posterior.

A key observation is that the current posterior $p(z_t|x_{1:t})$ can be expressed recursively in terms of the previous posterior $p(z_{t-1}|x_{1:t-1})$, the state model $p(z_t|z_{t-1})$, and the measurement model $p(x_t|z_t)$ using the following relation:

$$p(z_t|x_{1:t}) \propto p(x_t|z_t) \int p(z_t|z_{t-1}) \, p(z_{t-1}|x_{1:t-1}) \, dz_{t-1}. \tag{1}$$

Through this relation, the Bayesian solution from the previous time step can be updated with a new observation $x_t$ to obtain the Bayesian solution for the current time step.

We refer to any method that inputs probabilistic state and measurement models and returns the posterior or some approximation to it as a *Bayesian filter* or filtering algorithm. There are a host of ways to perform Bayesian filtering, loosely corresponding to methods by which one can compute the integrals in equation 1 (both the explicit integral and the integral required for re-normalization). We describe them in detail in Chapter 1.

The research question that Professor Harrison proposed was *"how would one perform Bayesian filtering using a model for $p(z_t|x_t)$ instead of $p(x_t|z_t)$?"* Neural decoding provides an application where the dimensionality of the hidden variable (latent intentions, 2- or 3-d cursor control) tends to be much lower than that of the observations (observed neural firing patterns, made increasingly detailed by recent technological advances). A model for $p(z_t|x_t)$ could prove more accurate than a model for $p(x_t|z_t)$ when $\dim(Z_t) \ll \dim(X_t)$, especially if such models need to be learned from data. Bayes' rule relates these two quantities as

$$p(z_t|x_t) = \frac{p(z_t|x_t)\, p(x_t)}{p(z_t)} \propto \frac{p(z_t|x_t)}{p(z_t)}$$

up to a constant in $x_t$.

Under the further restriction that $p(z_t|x_t)$ and $p(z_t)$ are approximated as Gaussians satisfying some conditions on their covariance structure, I showed that the posterior $p(z_t|x_{1:t})$ would also be Gaussian and easily computable. This is, in essence, what we call the *Discriminative Kalman Filter* (DKF). We explore it in detail in Chapter 2. Modeling $p(z_t|x_t)$ as Gaussian proves fundamentally different than modeling $p(x_t|z_t)$ as Gaussian. In particular, we are no longer specifying a complete generative model for $X_{1:T}, Z_{1:T}$. However, if we consider a limit where $\dim(X_t) \to \infty$, the Bernstein–von Mises theorem states that under mild conditions, $p(z_t|x_t)$ becomes Gaussian in the total variation metric. We show in Chapter 3 that, under this condition, the DKF estimate will converge in total variation to the true posterior. This proof is due in a great part to Prof. Harrison.

Prof. Leigh Hochberg and Dr. David Brandman, along with a talented team including Dr. John Simeral, Jad Saab, Tommy Hosman, among others, implemented the DKF as part of the BrainGate2 clinical trial, and Dr. David Brandman, Brittany Sorice, Jessica Kelemen, Brian Franco, and myself visited the homes of three volunteers to collect data and help them use this filter within the BrainGate system to control an on-screen cursor with mental imagery alone.

After some preliminary experiments comparing the DKF and Kalman filters, Dr. David Brandman suggested we design a version of the DKF to be robust to certain nonstationarities in neural data. By nonstationarity, we mean that the underlying statistical relationship between measured neural signals $X_t$ and intended control $Z_t$ (characterized by the measurement model) changes over time. In practice, this is due to both neural plasticity (the brain is changing, learning) and mechanical variability (the intracortical array may drop the signal from a particular neuron, or detect a new neuron). In Chapter 4, we describe how we successfully designed, implemented, and tested a Gaussian process model for $p(z_t|x_t)$ that worked in conjunction with the DKF to mitigate nonstationarities occurring in a single neuron.

# ACKNOWLEDGEMENTS

I thank my parents for their continued love and support. I thank Prof. Matthew Harrison for his patience, Prof. Basilis Gidas for his dedication to graduate-level instruction, and Prof. Jerome Darbon for his willingness to discuss mathematics. My office mates and rock climbing partners Michael Snarski, Ian Alevy, and Sameer Iyer have been great sources of encouragement. Outside the office, Cat Munro, Clark Bowman, Richard Kenyon, and Johnny Guzmán greatly enriched my life in Providence. I'm grateful to our division's intramural soccer team (and our captains Dan Johnson and Guo-Jhen Wu) and hope that possibly without me they will manage a championship in the near future. I am indebted to Dan Keating, and the entire Brown Polo Club family for teaching me the game of polo. There is nothing quite so refreshing after a long day cooped up at a computer than riding around on a horse thwacking at grapefruit-sized ball.

My collaborators in neuroscience, particularly Dr. David Brandman and Prof. Leigh Hochberg, have given my work meaning. Before coming to Brown, I had never imagined I would have the opportunity to work on such a practical and impactful project as BrainGate. As a group, we are grateful to B. Travers, Y. Mironovas, D. Rosler, Dr. John Simeral, J. Saab, T. Hosman, D. Milstein, B. Shanahan, B. Sorice, J. Kelemen, B. Franco, B. Jarosiewicz, C. Vargas-Irwin, and the many others whose work has made this project possible over the years. I was honored to help Chris Grimm complete his undergraduate thesis work in Bayesian filtering. Most importantly, we honor those who have dedicated some of the final years of their lives to participating in the project, especially T9 and T10, along with their families.

I thank the Brown–Kobe Exchange in High Performance Computing, and particularly Prof. Nobuyuki Kaya (賀谷 信幸) for his generosity, the Brown–IMPA Partnership travel grant, and the Brown International and Conference travel awards for the opportunities to learn and share knowledge all around the world. I also thank the Brown Institute for Brain Science graduate research award for a semester's respite from teaching.

I thank again those who championed me before coming to Brown, including Mrs. Kathleen Helbing who taught me the value and richness of history, Prof. Burgess Davis who first taught me analysis and probability, Prof. Robert Zink who introduced me to abstract algebra and served with me on the Purdue *Exponent's* Board of Directors, Prof. Hans Uli Walther who guided me through my first research experience, Prof. Daniel Ocone who introduced me to stochastic differential equations, and Prof. Eduardo Sontag who guided my first graduate research experience. I'm indebted to Prof. Victor Zavala for introducing me to Gaussian processes and piloting my first research paper.

In that vein, I'm grateful to John Wiegand and Chris O'Neil, and their families, with whom I went to church and elementary school, played soccer, went on adventures in Boy Scouts, and whom I continue to be fortunate to count as friends. John and the Wiegands have made coming home to Indiana something to look forward to. The O'Neils were my first friends from Rhode Island, before I even knew what Rhode Island was. I'm grateful to Ed Chien, my first climbing partner and a great friend from my time in New Jersey. My friends have been a constant source of support and encouragement during my time in grad school, greatly enriching my life outside of academia, including of course S. Thais, V. Hallock, B. Dewenter, A. Bacoyanis, A. Johnson, S. Boutwell, L. Walton, N. Meyers, R. Parker, B. Whitney, N. Trask, L. Jia, E. Solomon, L. Roesler, M. Montgomery, E. Knox, L. Appel, L. Akers, and so many others!

I thank the Jim, Susan, Margaret, and Monica Hershberger, and the Cassidy family, D. Ruskaup, D. Steger, M. Christman, and all lifelong friends of our family. And of course, I thank my own family; aunts, uncles, and cousins, and especially my grandpa James Q. Beatty.

My work benefited immensely from open source projects and developers, including the makers and maintainers of LaTeX, Python, Numpy, Tensorflow, Git, the Library Genesis Project, and Sci-Hub, among others.

I'm extremely grateful to our warm-hearted and supportive staff including J. Radican, S. Han, C. Dickson, E. Fox, C. Hansen-Decelles, T. Saotome, J. Robinson, and R. Wertheimer.

I credit my cat Bobo with all typos: she would happily spend as much time on my laptop as I do. And of course, many thanks go to Elizabeth Crites.

# CONTENTS

★ Parts of this thesis have or will appear in other publications. In particular, Chapter 2 is joint work with M. Harrison and D. Brandman, and Chapter 4 is joint work with D. Brandman, M. Harrison, and L. Hochberg, among others.

# LIST OF TABLES

# LIST OF FIGURES

# CHAPTER 1

# AN OVERVIEW OF BAYESIAN FILTERING

> Beim Anblick eines Wasserfalls meinen wir in den zahllosen
> Biegungen, Schlängelungen, Brechungen der Wellen Freiheit des
> Willens und Belieben zu sehen; aber alles ist notwendig, jede
> Bewegung mathematisch auszurechnen... wenn in einem Augenblick
> das Rad der Welt still stände und ein allwissender, rechnender
> Verstand da wäre, um diese Pause zu benützen, so könnte er bis in die
> fernsten Zeiten die Zukunft jedes Wesens weitererzählen und jede
> Spur bezeichnen, auf der jenes Rad noch rollen wird.

F. W. Nietzsche, *Human, All too human*

## 1.1   Preface

This chapter is primarily my work, but was definitely inspired by the survey of Chen [Che13]. Books have been written on this topic alone, e.g. Wiener [Wie49], Jazwinski [Jaz70], Anderson and Moore [AM79], Strobach [Str90], and Särkkä [Sär13], with applications including the Apollo program [Hal66; BL70; GA10], aircraft guidance [SWL70], GPS navigation [HLC01], weather forecasting [BMH17], and—of course—neural filtering (covered later), so I tried here to provide a digestible, salient overview.

---

At the sight of a waterfall we may opine that in the countless curves, spirations and dashes of the waves we behold freedom of the will and of the impulses. But everything is compulsory, everything can be mathematically calculated... If, on a sudden, the entire movement of the world stopped short, and an all knowing and reasoning Intelligence were there to take advantage of this pause, He could foretell the future of every being to the remotest ages and indicate the path that would be taken in the world's further course.

## 1.2 Introduction

Consider a state space model for $Z_{1:T} := Z_1, \ldots, Z_T$ (latent states) and $X_{1:T} := X_1, \ldots, X_T$ (observations) represented as a Bayesian network:

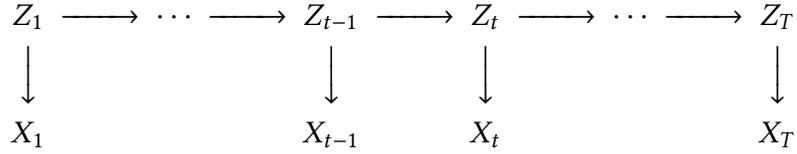

The conditional density of $Z_t$ given $X_{1:t}$ can be expressed recursively using the Chapman–Kolmogorov equation and Bayes' rule [Che03]

$$p(z_t|x_{1:t}) \propto p(x_t|z_t) \int p(z_t|z_{t-1}) \, p(z_{t-1}|x_{1:t-1}) \, dz_{t-1} \tag{1.1}$$

where the proportionality constant involves an integral over $z_t$. To be more explicit, we can re-write eq. (1.1) as follows:

$$p(z_t|x_{1:t-1}) = \int p(z_t|z_{t-1})p(z_{t-1}|x_{1:t-1}) \, dz_{t-1}, \tag{1.2a}$$

$$p(z_t|x_{1:t}) = \frac{p(x_t|z_t)p(z_t|x_{1:t-1})}{\int p(x_t|z_t)p(z_t|x_{1:t-1}) \, dz_t}. \tag{1.2b}$$

### 1.2.1 Methodology Taxonomy

Modeling these conditional probabilities and solving or approximating the integrals in eq. (1.2) constitutes Bayesian filtering. We taxonomize filtering methods according to how the integral in eq. (1.1) is computed. This mirrors closely the ways that Bayesians perform inference in general. To filter, one may:

1. *Use a model with an exact solution*, such as the Kalman filter [Kal60; KB61], or the model specifications of Beneš [Ben81] or Daum [Dau84; Dau86]. These models entail no approximation and integration is done in closed form.

2. *Employ a variational method that replaces the current model with a closely-related tractable one.* For example, the extended Kalman filter and the statistically-linearized filter fit a generic model to a linear model that then integrates exactly [Gel74; Sär13]. One can also approximate an arbitrary distribution as the sum of Gaussians and then handle each Gaussian component analytically [AS72]. Alternatively, integration can be done via a Laplace transform [Koy+10]. These similar methods have many names in the literature, including the Gaussian assumed density filter, Series expansion-based filters, Fourier–Hermite Kalman filter [SS12]. The model is approximated, but then integration can be done exactly.

3. *Integrate using a quadrature rule.* In this category we include sigma-point filters such as the Unscented Kalman Filter [JU97; WM00; Mer04] and also Quadrature Kalman filters [Ito00; IX00] and Cubature Kalman filters [AHE07; AH09]. Under these models, integrals are approximated based on function evaluations at deterministic points.

4. *Integrate with Monte Carlo.* Such approaches are called Sequential Monte Carlo or particle filtering [HM54; GSS93]. These methods apply to all classes of models, but tend to be the most expensive and suffer the curse of dimensionality [DH03]. Integration is done with a Monte Carlo approximation; the models do not need to be approximated.

## 1.3   Exact Filtering with the Kalman Filter (KF)

The Kalman filter specifies a linear, Gaussian relationship between states and observations to yield an analytic solution that can be efficiently computed. Here we derive the classic Kalman updates [Kal60; KB61].

### 1.3.1   Model

Let $\eta_d(z; \mu, \Sigma)$ denote the $d$-dimensional multivariate Gaussian distribution with mean vector $\mu \in \mathbb{R}^{d \times 1}$ and covariance matrix $\Sigma \in \mathbb{S}_d$ evaluated at $z \in \mathbb{R}^{d \times 1}$, where $\mathbb{S}_d$ denotes the set of $d \times d$ positive definite (symmetric) matrices. Assume that the latent states are a stationary, Gaussian, vector autoregressive model of order one; namely, for $A \in \mathbb{R}^{d \times d}$ and $S, \Gamma \in \mathbb{S}_d$,

$$p(z_0) = \eta_d(z_0; 0, S), \tag{1.3a}$$

$$p(z_t | z_{t-1}) = \eta_d(z_t; A z_{t-1}, \Gamma), \tag{1.3b}$$

For observations in $\mathcal{X} = \mathbb{R}^{n \times 1}$ and for fixed $H \in \mathbb{R}^{n \times d}$, $b \in \mathbb{R}^{n \times 1}$, and $\Lambda \in \mathbb{S}_n$, we have

$$p(x_t | z_t) = \eta_n(x_t; Hz_t + b, \Lambda). \tag{1.4}$$

### 1.3.2 Inference

Under the above model, we see that the posterior at each step will be Gaussian so we may adopt the ansatz:

$$p(z_t | x_{1:t}) = \eta_d(z_t; v_t, \Phi_t). \tag{1.5}$$

We solve for $v_t$ and $\Phi_t$ recursively using eq. (1.1):

$$p(z_t | x_{1:t}) \propto \eta_n(x_t; Hz_t + b, \Lambda) \int \eta_d(z_t; Az_{t-1}, \Gamma) \eta_d(z_{t-1}; v_{t-1}, \Phi_{t-1}) \, dz_{t-1} \tag{1.6}$$

$$\propto \eta_n(x_t; Hz_t + b, \Lambda) \, \eta_d(z_t; Av_{t-1}, A\Phi_{t-1}A^\top + \Gamma). \tag{1.7}$$

Setting

$$\hat{v}_{t-1} = Av_{t-1}, \tag{1.8}$$

$$\hat{\Phi}_{t-1} = A\Phi_{t-1}A^\top + \Gamma, \tag{1.9}$$

we have

$$p(z_t | x_{1:t}) \propto \eta_n(x_t; Hz_t + b, \Lambda) \, \eta_d(z_t; \hat{v}_{t-1}, \hat{\Phi}_{t-1}) \tag{1.10}$$

$$\propto e^{-(x_t - Hz_t - b)^\top \Lambda^{-1}(x_t - Hz_t - b)/2} e^{-(z_t - \hat{v}_{t-1})^\top \hat{\Phi}_{t-1}^{-1}(z_t - \hat{v}_{t-1})/2} \tag{1.11}$$

$$\propto e^{-z_t^\top H^\top \Lambda^{-1} Hz_t/2 + z_t^\top H^\top \Lambda^{-1}(x_t - b) - z_t^\top \hat{\Phi}_{t-1}^{-1} z_t/2 + z_t^\top \hat{\Phi}_{t-1}^{-1} \hat{v}_{t-1}} \tag{1.12}$$

$$\propto e^{-z_t^\top (H^\top \Lambda^{-1} H + \hat{\Phi}_{t-1}^{-1}) z_t/2 + z_t^\top (H^\top \Lambda^{-1}(x_t - b) + \hat{\Phi}_{t-1}^{-1} \hat{v}_{t-1})} \tag{1.13}$$

$$\eta_d\big(z_t; \Phi_t(H^\top \Lambda^{-1}(x_t - b) + \hat{\Phi}_{t-1}^{-1} \hat{v}_{t-1}), \Phi_t\big) \tag{1.14}$$

where

$$\Phi_t = (H^\top \Lambda^{-1} H + \hat{\Phi}_{t-1}^{-1})^{-1} \tag{1.15}$$

$$= \hat{\Phi}_{t-1} - \hat{\Phi}_{t-1} H^\top (H\hat{\Phi}_{t-1}H^\top + \Lambda)^{-1} H\hat{\Phi}_{t-1} \tag{1.16}$$

$$= (I_d - \hat{\Phi}_{t-1}H^\top (H\hat{\Phi}_{t-1}H^\top + \Lambda)^{-1}H)\hat{\Phi}_{t-1} \tag{1.17}$$

due to the Woodbury matrix identity, where $I_d$ is the $d$-dimensional identity matrix. Many textbook derivations define the Kalman gain

$$K_t := \hat{\Phi}_{t-1} H^\intercal (H \hat{\Phi}_{t-1} H^\intercal + \Lambda)^{-1} \tag{1.18}$$

so that

$$\Phi_t = (I_d - K_t H) \hat{\Phi}_{t-1} \tag{1.19}$$

and

$$v_t = \hat{v}_{t-1} + K_t(x_t - b - H\hat{v}_{t-1}). \tag{1.20}$$

These are the traditional Kalman updates [Kal60]. Kalman's original paper does not assume Gaussian dynamics; however under the Gaussian modeling assumptions, this filter yields exact solutions to eq. (1.1).

### 1.3.3 Remark

Note that the Kalman model implies

$$p(z_t | x_{1:t-1}) = \eta_d(z_t; \hat{v}_{t-1}, \hat{\Phi}_{t-1}) \tag{1.21}$$

so that

$$p(x_t | x_{1:t-1}) = \eta_n(x_t; H\hat{v}_{t-1} + b, H\hat{\Phi}_{t-1}H^\intercal + \Lambda). \tag{1.22}$$

Let $\bar{X}_t, \bar{Z}_t$ be distributed as $X_t, Z_t$ conditioned on $X_{1:t-1}$, respectively. Then

$$\mathbb{V}[\bar{X}_t] = H\hat{\Phi}_{t-1}H^\intercal + \Lambda, \tag{1.23}$$

$$\text{Cov}[\bar{Z}_t, \bar{X}_t] = \hat{\Phi}_{t-1}H^\intercal, \tag{1.24}$$

so we can re-write eq. (1.18), eq. (1.19), and eq. (1.20) as

$$K_t = \text{Cov}[\bar{Z}_t, \bar{X}_t](\mathbb{V}[\bar{X}_t])^{-1}, \tag{1.25}$$

$$\Phi_t = \mathbb{V}[\bar{Z}_t] - K_t \mathbb{V}[\bar{X}_t]K_t^\intercal, \tag{1.26}$$

$$v_t = \mathbb{E}[\bar{Z}_t] + K_t(x_t - \mathbb{E}[\bar{X}_t]). \tag{1.27}$$

This will form the basis for the Gaussian assumed density filter.

### 1.3.4 Related Work

Beneš [Ben81] and Daum [Dau84; Dau86; Dau05] extended the families of models under which eq. (1.1) may be solved exactly. In the case that the state space is finite, the grid-based method also provides an exact solution [Seg76; Mar79; Ell94; EY94; Aru+02; KP16]. The underlying idea is that when there are only a finite number of states, a particle filter with a particle for each state makes eq. (1.66) an exact representation for the posterior density, and such a representation can be updated exactly [Aru+02].

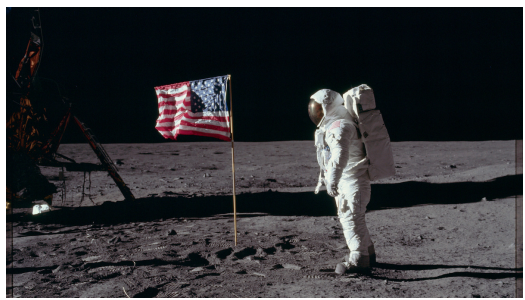

Figure 1.1 – The Apollo Lunar Module used a variant of the Kalman Filter to land Neil Armstrong on the moon [Hal66; Hoa69; BL70]. Image credit: NASA.

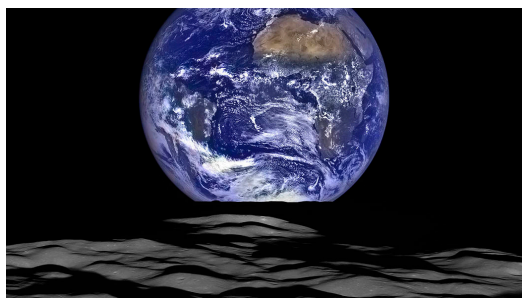

Figure 1.2 – GPS receivers use the Extended Kalman filter to model and mitigate satellite clock offset and atmospheric delays [AB95; HLC01]. Image credit: NASA.

## 1.4 Model Approximation with the Extended Kalman Filter

This approach expands the model from Section 1.3.1 and performs inference by finding the closest tractable model and using it instead.

### 1.4.1 Model

We extend our model now to allow the relationship between the latent states and observations to be nonlinear:

$$p(x_t|z_t) = \eta_n(x_t; h(z_t), \Lambda) \tag{1.28}$$

where $h : \mathbb{R}^d \rightarrow \mathbb{R}^n$ is a differentiable function. We use the same state process as in Section 1.3.1, namely

$$p(z_0) = \eta_d(z_0; 0, S), \tag{1.29a}$$

$$p(z_t | z_{t-1}) = \eta_d(z_t; A z_{t-1}, \Gamma). \tag{1.29b}$$

Many of the original derivations and references include a nonlinear, Gaussian state update the state model as well. The way that inference is adapted to allow for nonlinearity is identical for both the measurement and state models, so we discuss only the measurement model here.

### 1.4.2   Inference

We may approximate the solution to eq. (1.1) in the same form as eq. (1.5) by linearizing the function $h : \mathbb{R}^d \rightarrow \mathbb{R}^n$ around $\hat{v}_{t-1}$ (from Equation 1.8):

$$h(z_t) \approx h(\hat{v}_{t-1}) + \tilde{H}(z_t - \hat{v}_{t-1}) \tag{1.30}$$

where $\tilde{H} \in \mathbb{R}^{n \times d}$ is given component-wise for $1 \leq i \leq n$ and $1 \leq j \leq d$ as

$$\tilde{H}_{ij} = \frac{\partial}{\partial z_j} h_i(z) \big|_{z = \hat{v}_{t-1}}. \tag{1.31}$$

With this Taylor series approximation, we then take

$$p(x_t | z_t) = \eta_n(x_t; h(\hat{v}_{t-1}) + \tilde{H}(z_t - \hat{v}_{t-1}), \Lambda) \tag{1.32}$$

$$= \eta_n(x_t; \tilde{H} z_t + h(\hat{v}_{t-1}) - \tilde{H} \hat{v}_{t-1}, \Lambda) \tag{1.33}$$

$$= \eta_n(\tilde{x}_t; \tilde{H} z_t + \tilde{b}, \Lambda) \tag{1.34}$$

where $\tilde{b} = h(\hat{v}_{t-1}) - \tilde{H} \hat{v}_{t-1}$. This problem is now identical to that of the original Kalman filter, where $H, b$ have been replaced by $\tilde{H}, \tilde{b}$, respectively. Thus, the updated equations

(Equations 1.18, 1.19, and 1.20 for the KF) become

$$K_t = \hat{\Phi}_{t-1}\tilde{H}^\intercal(\tilde{H}\hat{\Phi}_{t-1}\tilde{H}^\intercal + \Lambda)^{-1}, \tag{1.35}$$

$$\Phi_t = (I_d - K_t\tilde{H})\hat{\Phi}_{t-1}, \tag{1.36}$$

$$v_t = \hat{v}_{t-1} + K_t(x_t - h(\hat{v}_{t-1})). \tag{1.37}$$

### 1.4.3 Related Work

Instead of a first order Taylor series approximation, it is also possible to use statistical linearization within the EKF framework [Gel74; Sär13]. The resulting filter is aptly named the statistically linearized filter. With $Z \sim \mathcal{N}(0, S)$, parameters for the linear approximation are chosen to minimize the MSE

$$\hat{b}, \hat{A} := \underset{b,A}{\arg\min}\{\mathbb{E}[(h(Z) - (\hat{b} + \hat{A}Z))^\intercal(h(Z) - (\hat{b} + \hat{A}Z))]\} \tag{1.38}$$

yielding

$$\hat{b} = \mathbb{E}[h(Z)] \tag{1.39}$$

$$\hat{A} = \mathbb{E}[h(Z)Z^\intercal]S^{-1} \tag{1.40}$$

The approximation

$$h(x) \approx \hat{b} + \hat{A}x \tag{1.41}$$

is then used in place of eq. (1.30).

It is also possible to use a second-order expansion [AWB68; GH12]. Alternatively, one can use a Fourier-Hermite series representation in the EKF framework [SS12].

### 1.4.4 The Iterative EKF (IEKF)

This approach iteratively updates the center point of the Taylor series expansion used in the EKF to obtain a better linearization [FB66; WTA69]. In place of eq. (1.35), eq. (1.36), and eq. (1.37), the IEKF updates are initialized by $v_t^0 = \hat{v}_{t-1}$ and $\Phi_t^0 = \hat{\Phi}_{t-1}$ and then proceed

as follows:

$$H^{i+1} = h'(v_t^i) \tag{1.42}$$

$$K_t^{i+1} = \hat{\Phi}_{t-1}(H^{i+1})^\intercal (H^{i+1}\hat{\Phi}_{t-1}(H^{i+1})^\intercal + \Lambda)^{-1}, \tag{1.43}$$

$$\Phi_t^{i+1} = (I_d - K_t^{i+1}H^{i+1})\hat{\Phi}_{t-1}, \tag{1.44}$$

$$v_t^{i+1} = \hat{v}_{t-1} + K_t^{i+1}(x_t - h(v_t^i) - H^{i+1}(\hat{v}_{t-1} - v_t^i)). \tag{1.45}$$

Bell and Cathey [BC93] showed that this algorithm is equivalent to iterative Gauss–Newton updates to place $v_t$ at the MAP for $Z_t$. Finding the MAP can be done with other iterative methods as well, e.g. Levenberg–Marquardt or progressive correction [Fat+12]. Iterative techniques have been applied to other filters [MNK14]. The iterative EKF itself was extended to the Backward-Smoothing Kalman Filter [Psi05; Psi13].

## 1.5  Model Approximation with Laplace-based Methods

The Laplace approximation performs integration by replacing an integrand $f$ with a Gaussian centered at the maximum of $f$ matching the curvature of $f$ at that point [KTK90; But07]. Such a method can be used generally for for Bayesian inference [Koy+10; QML15].

### 1.5.1  Laplace Approximation

The Laplace approximation can be used to approximate the integrals in Equation 1.1. At step $t$,

$$p(z_t|x_{1:t}) \propto p(x_t|z_t) \int p(z_t|z_{t-1})\, p(z_{t-1}|x_{1:t-1})\, dz_{t-1}$$

$$= p(x_t|z_t)\, p(z_t|x_{1:t-1}) =: r(z_t)$$

where $p(z_t|x_{1:t-1}) = \eta(z_t; v, \Phi)$ so that

$$g(z_t) := \log(r(z_t)) = -\frac{1}{2}(z_t - v)^\intercal \Phi^{-1}(z_t - v) - \frac{1}{2}\log\det(2\pi\Phi) + \log(p(x_t|z_t))$$

where $r(z_t) = e^{g(z_t)}$. We find $z_t^* = \arg\max_z g(z)$ and form the Laplace approximation

$$r(z_t) \approx e^{g(z_t^*) + \nabla g(z_t^*)(z - z_t^*) + \frac{1}{2}(z - z_t^*)^\top Hg(z_t^*)(z - z_t^*)}$$

where the gradient $\nabla g(z_t^*) = 0$ because $z_t^*$ is an extremal point and the Hessian $Hg(z_t^*)$ is negative definite because $z_t^*$ is a maximum. Thus we have the approximation:

$$p(z_t|x_{1:t}) \approx \eta(z_t; z_t^*, -(Hg(z_t^*))^{-1})$$

## 1.6  Model Approximation with the Gaussian Assumed Density Filter

Under the same model as the EKF (see Section 1.4.1), we can instead perform inference by taking the information projection of the posterior onto the space of Gaussian distributions [Min01a].

### 1.6.1  Inference

Under the assumption that

$$p(z_t|x_{1:t}) = \eta_d(z_t; \nu_t, \Phi_t) \tag{1.46}$$

for all $t$, the Chapman–Kolmogorov recursion eq. (1.1) becomes

$$p(z_t|x_{1:t}) \propto \eta_n(x_t; h(z_t), \Lambda)\, \eta_d(z_t; A\nu_{t-1}, A\Phi_{t-1}A^\top + \Gamma) \tag{1.47}$$

$$\propto \eta_n(x_t; h(z_t), \Lambda)\, \eta_d(z_t; \hat{\nu}_{t-1}, \hat{\Phi}_{t-1}) \tag{1.48}$$

as before. The information projection then finds $\nu_t, \Phi_t$ that minimize the following KL divergence [CT06; Mur12]:

$$\nu_t, \Phi_t = \arg\min_{a,b}\{D_{KL}(p(z_t|x_{1:t})||\eta_d(z_t; a, b)\} \tag{1.49}$$

With the following calculations:

$$\mu_x = \int h(z_t) \, \eta_d(z_t; \hat{v}_{t-1}, \hat{\Phi}_{t-1}) \, dz_t, \tag{1.50}$$

$$P_{xx} = \int (h(z_t) - \mu_x)(h(z_t) - \mu_x)^{\intercal} \, \eta_d(z_t; \hat{v}_{t-1}, \hat{\Phi}_{t-1}) \, dz_t + \Lambda, \tag{1.51}$$

$$P_{zx} = \int (z_t - \hat{v}_{t-1})(h(z_t) - \mu_x)^{\intercal} \, \eta_d(z_t; \hat{v}_{t-1}, \hat{\Phi}_{t-1}) \, dz_t, \tag{1.52}$$

this problem has the solution:

$$K = P_{zx} P_{xx}^{-1}, \tag{1.53}$$

$$\Phi_t = \hat{\Phi}_{t-1} - K P_{xx} K^{\intercal}, \tag{1.54}$$

$$v_t = \hat{v}_{t-1} + K(x_t - \bar{x}). \tag{1.55}$$

Compare eq. (1.53), eq. (1.54), eq. (1.55) to the analogues for the Kalman filter, eq. (1.25), eq. (1.26), eq. (1.27), respectively.

### 1.6.2    Related Work

Expectation Propagation extends Assumed Density Filtering with iterative refinement of estimates [Min01a; Min01b]. It iterates over the entire history of observations at every time step, and so may not be practical in an online filtering setting.

## 1.7    Integral Approximation to the Gaussian ADF Model Approximation

Under the same model as the EKF (see Section 1.4.1), we apply the variational method from the Gaussian assumed density filter (see Section 1.6) to obtain the integral equations eq. (1.50), eq. (1.51), and eq. (1.50). Various quadrature methods have been developed to approximate these integrals as the weighted average over integrand evaluations at a finite set of deterministic points. Such approaches don't require differentiating the function $h$ in eq. (1.28) and typically require a smaller number of evaluation points than Monte Carlo-based methods.

### 1.7.1 Unscented and Sigma Point Kalman Filters (UKF, SPKF)

The UKF propagates $2d + 1$ weighted points through $h$ to estimate the integrals eq. (1.50), eq. (1.51), and eq. (1.50), in a method known as the unscented transform [JU97]. Let $\hat{v}_{t-1}, \hat{\Phi}_{t-1}$ be as in Equations 1.8, 1.9. We introduce parameters $\alpha > 0, \beta \in \mathbb{R}$ [WM00] and consider the set of sigma vectors $\zeta_0, \ldots, \zeta_{2d} \in \mathbb{R}^d$ given by

$$\zeta_0 = \hat{v}_{t-1}, \tag{1.56}$$

$$\zeta_i = \hat{v}_{t-1} + \left(\sqrt{\alpha^2 d\hat{\Phi}}\right)_i, \qquad i = 1, \ldots, d \tag{1.57}$$

$$\zeta_i = \hat{v}_{t-1} - \left(\sqrt{\alpha^2 d\hat{\Phi}}\right)_i, \qquad i = d + 1, \ldots, 2d \tag{1.58}$$

where $\left(\sqrt{\alpha^2 d\hat{\Phi}}\right)_i$ denotes the $i$th row of the matrix square root. We set weights $w_0^{(m)} = 1 - 1/\alpha^2$, $w_0^{(c)} = 2 - 1/\alpha^2 - \alpha^2 + \beta$, and $w_0^{(m)} = w_0^{(c)} = 1/(2\alpha^2 d)$. We let

$$\mu_x = \sum_{i=0}^{2d} w_i^{(m)} h(\zeta_i), \tag{1.59}$$

$$P_{xx} = \sum_{i=0}^{2d} w_i^{(c)} (h(\zeta_i) - \bar{x})(h(\zeta_i) - \bar{x})^\top, \tag{1.60}$$

$$P_{zx} = \sum_{i=0}^{2d} w_i^{(c)} (\zeta_i - \hat{v}_{t-1})(h(\zeta_i) - \bar{x})^\top. \tag{1.61}$$

The framework of eq. (1.53), eq. (1.54), and eq. (1.55) is then used with the above values. Wan and Merwe [WM00] suggest default parameters $\alpha = 0.001$ and $\beta = 2$. Stirling's interpolation formula, a central difference scheme to approximate second derivatives, can also be used [the Central Difference Kalman filter (CDKF) of IX00; NPR00]. Merwe [Mer04] referred to these methods collectively as sigma-point Kalman filters.

### 1.7.2 Quadrature-type Kalman Filters (QKF, CKF)

The integrals eq. (1.50), eq. (1.51), and eq. (1.50) can also be approximated with the Gauss–Hermite quadrature rule [NS82]:

$$\int \eta(z; 0, 1) f(z) \approx \sum_{i=1}^{m} w_i f(\mu + \sqrt{2}\sigma z_i). \tag{1.62}$$

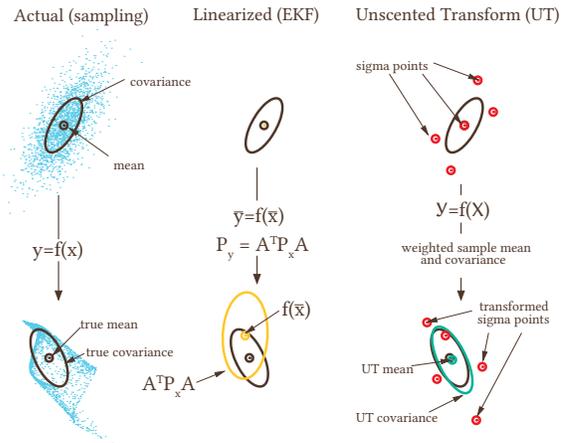

Figure 1.3 – This figure compares the evolution of the true density through a particle filter (left) to the EKF linear approximation (center) and the UKF sigma-point method (right). Image credit: Wan and Merwe [WM00] © 2000 IEEE.

Such an approximation is exact when $f$ is a polynomial of degree less than $m$, where the weights $w_i$ and knots $z_i$ are given in [Gol73]. A simple change of variables can be used to generalize eq. (1.62) to nonstandard normal distributions. Ito and Xiong [IX00] extend this rule to multivariate quadrature:

$$\int \eta_d(z; 0, S) f(z) \approx \sum_{i_1=1}^{m} \cdots \sum_{i_d=1}^{m} w_{i_1} \cdots w_{i_d} f(q_{i_1}, \ldots, q_{i_d}) \tag{1.63}$$

for specified weights $w_{i_j}$ and knots $q_{i_j}$. Such a rule is exact for all polynomials of multi-degree up to $(2m-1, \ldots, 2m-1)$. Using the above quadrature rule for integration in this filtering model yields the Quadrature Kalman filter of Challa, Bar-Shalom, and Krishnamurthy [CBK00] and Ito and Xiong [IX00]. The spherical-radial cubature rule implements a cubature scheme to the same effect [the Cubature–Quadrature Kalman filter of AHE07; AH09].

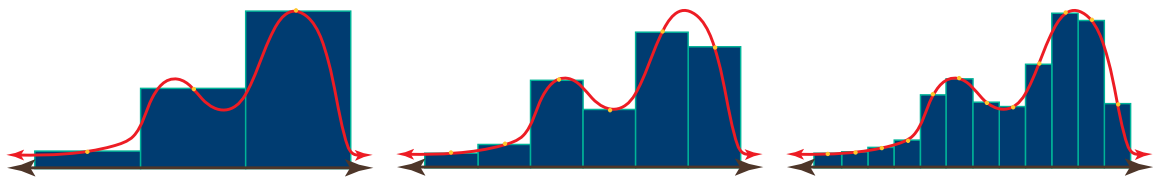

Figure 1.4 – The Midpoint Quadrature Rule approximates an integral by partitioning the domain of integration and approximating the integral over each partition $[a, b]$ as $\int_a^b f(x) \, dx \approx (b-a) \cdot f(\frac{a+b}{2})$. Such an approximation improves as the partition becomes finer.

### 1.7.3   Fourier–Hermite Kalman Filter (FHKF)

The integrals eq. (1.50), eq. (1.51), and eq. (1.50) can also be approximated as a Fourier–Hermite expansion [SS12].

### 1.7.4   Related Work

It is possible to propagate the first two moments for the Gaussian posterior using particle filtering [KD03a] or even a Gaussian process [Ko+07; KF09].

## 1.8   The Particle Filter (PF)

Metropolis and Ulam [MU49] developed the Monte Carlo method for numerical integration. The idea is to replace integration with summation, i.e. for $f \in L^1(dp)$:

$$\mathbb{E}[f(Z)] = \int f(z)\, p(z)\, dz \approx \frac{1}{N} \sum_{i=1}^{N} f(Z_i) \tag{1.64}$$

where $Z, Z_1, \ldots, Z_N \sim^{\text{i.i.d.}} p$. The Strong Law of Large Numbers implies that the approximation error tends to zero a.s. as $N \to \infty$. If it is easier to draw samples from some p.d.f. $q$ where $q \ll p$ ($q$ is absolutely continuous with respect to $p$), then we obtain importance sampling:

$$\mathbb{E}[f(Z)] = \int f(z)\, \frac{p(z)}{q(z)}\, q(z)\, dz \approx \frac{1}{N} \sum_{i=1}^{N} f(Z_i) \frac{p(Z_i)}{q(Z_i)}. \tag{1.65}$$

By optimizing over possible sampling distributions, it is also possible to use this method to reduce variance for the approximation. Applying the importance sampling approximation in eq. (1.65) to the Chapman–Kolmogorov recursion in eq. (1.2) yields Sequential importance sampling (SIS) [HM69; Han70; Kit96; Mor96; DGA00; CMR05; CGM07].

We outline the Sequential Importance Resampling (SIR) method pioneered by [GSS93]. We first represent $p(z_{t-1}|x_{1:t-1})$ as a sum of weighted particles

$$p(z_{t-1}|x_{1:t-1}) \approx \sum_{\ell=1}^{L} w_{t-1}^{(\ell)} \delta_{z_{t-1}^{(\ell)}}(z_{t-1}). \tag{1.66}$$

At each step in the recursion, we resample $z_{t-1}^{(\ell)}$ according to the weights $w_{t-1}^{(\ell)}$ to obtain equally-weighted samples $\tilde{z}_{t-1}^{(\ell)}$. This resampling prevents particle collapse (an issue where the vast majority of particles are assigned negligible weight so that the number of effective particles becomes very small) and gives us a modified form of eq. (1.1):

$$p(z_t|x_{1:t}) \propto \sum_{\ell=1}^{L} p(x_t|z_t^{(\ell)}) \int p(z_t|y) \delta_{z_{t-1}^{(\ell)}}(y) \, dy. \tag{1.67}$$

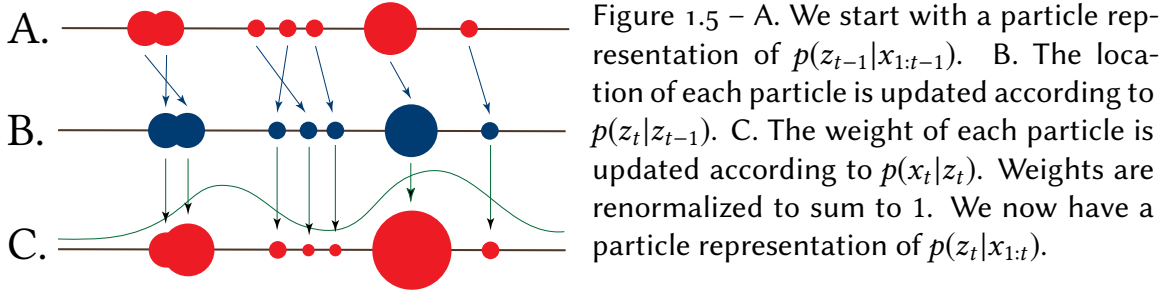

Figure 1.5 – A. We start with a particle representation of $p(z_{t-1}|x_{1:t-1})$. B. The location of each particle is updated according to $p(z_t|z_{t-1})$. C. The weight of each particle is updated according to $p(x_t|z_t)$. Weights are renormalized to sum to 1. We now have a particle representation of $p(z_t|x_{1:t})$.

(The resampling step is what distinguishes SIR from SIS. ) We then sample $z_t^{(\ell)}$ from

$$Z_t^{(\ell)} \sim Z_t|\{Z_{t-1} = \tilde{z}_{t-1}^{(\ell)}\} \tag{1.68}$$

so that

$$p(z_t|x_{1:t-1}) \approx \sum_{\ell=1}^{L} \delta_{z_t^{(\ell)}}(z_t)$$

and then update the weights

$$w_t^{(\ell)} \propto p(x_t|z_t^{(\ell)}). \tag{1.69}$$

Weights are normalized to sum to 1 and this yields our next representation in the form of eq. (1.66).

### 1.8.1  Related Work

Alternate sampling strategies [see, e.g., Che03; Liu08] can be used to improve filter performance, including: acceptance-rejection sampling [HM69], stratified sampling [DC05], hybrid MC [CF01], and quasi-MC [GC15].

There are also ensemble versions of the Kalman filter that are used to propagate the covariance matrix in high state-dimensions including the ensemble Kalman filter [enKF:

Figure 1.6 – A. We start with a particle representation of $p(z_{t-1}|x_{1:t-1})$. B. Particles are selected with replacement according to their relative weights. This gives us a new set of particles that are now equally weighted. C. The location of each particle is updated according to $p(z_t|z_{t-1})$. D. The weight of each particle is updated according to $p(x_t|z_t)$. Weights are renormalized to sum to 1, yielding a particle representation of $p(z_t|x_{1:t})$.

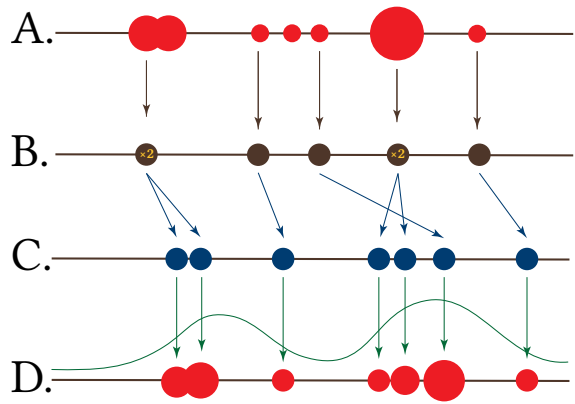

Eve94] and ensemble transform Kalman filter [ETKF: BEM01; Maj+02], along with versions that produce local approximations for covariance and can be parallelized [Ott+02; Ott+04; HKS07]. These filters seem well-suited to climate modeling.

Recent innovations include: the auxiliary particle filter that introduces an auxiliary variable to increase sampling efficiency and robustness to outliers [PS99], introducing block sampling techniques to particle filters [DBS06], resample–move algorithms that increase particle diversity without changing the estimated distribution [GB01], and MCMC moves within particle filters [ADH10].

Many of the algorithms mentioned in previous sections can be reformulated with a particle filter-based integral approximation: the unscented particle filter [Mer+01], the sigma-point particle filter [Mer04], the Gaussian mixture sigma-point particle filter [MW03], and a Laplace method-inspired particle filter [QML16].

## 1.9   Filtering Innovations

We describe here some meta-methods used to improve filter performance than have proven successful across multiple filter types.

### 1.9.1   Square-Root Transform

Various decompositions of the covariance estimate have been used to ensure positive-definiteness and obtain better matrix conditioning in filter recursions. The idea is to store and update the matrix square root of the covariance estimate instead of the covariance estimate itself, thereby working with a matrix that automatically yields a positive-definite covariance estimate and that possesses a condition number that is the square root of the original [AM79]. Potter [Pot63] pioneered the approach with the Cholesky decomposition and scalar-valued observations; Bellantoni and Dodge [BD67] extended the algorithm to multidimensional observations.

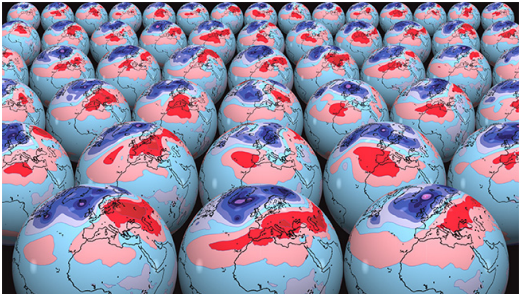

Figure 1.7 – Particles in a particle filter can represent a wide range of things, from global atmospheric conditions to phylogenetic trees [BSJ12; WBD15; DDM18]. Image credit: European Centre for Medium-Range Weather Forecasts.

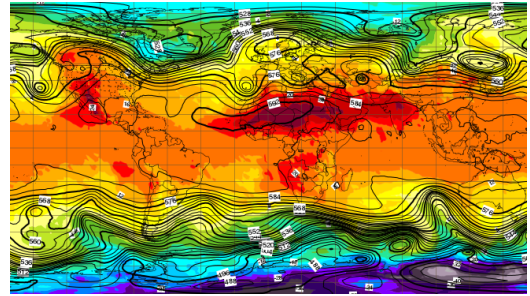

Figure 1.8 – Weather forecasts assimilate data with ensemble versions of the Kalman filter [Eve94; BEM01; Ott+04; HKS07; BMH17]. Image credit: European Centre for Medium-Range Weather Forecasts.

Note that eq. (1.17) is equivalent to:

$$(\Phi_t)^{-1} = (\hat{\Phi}_{t-1})^{-1} + H^\intercal \Lambda^{-1} H. \tag{1.70}$$

With the following Cholesky decompositions

$$\hat{\Phi}_{t-1} = PP^\intercal \tag{1.71}$$

$$\Lambda = LL^\intercal \tag{1.72}$$

it becomes

$$(\Phi_t)^{-1} = (PP^\intercal)^{-1} + H^\intercal (L^{-1})^\intercal L^{-1} H \tag{1.73}$$

$$= (P^{-1})^\intercal (I_d + BB^\intercal) P^{-1} \tag{1.74}$$

where $B = (L^{-1}HP)^\intercal$. Taking inverses then yields:

$$\Phi_t = P(I_d + BB^\intercal)^{-1}P^\intercal \tag{1.75}$$

and the Kalman gain is given by

$$K_t = PB(I_d + B^\intercal B)^{-1}L^{-1} \tag{1.76}$$

Subsequent work generalized it further [And68; Sch70; KBS71], including the use of other decompositions such as QR [SK93; SK94], Householder [DM69; Car73], and U-D [Bie75; Tho76]. Square-root versions have also been developed for the UKF and CDKF [MW01].

### 1.9.2   Rao–Blackwellization

The Rao–Blackwell theorem states that, given an unbiased estimator $\hat{\theta}$ and a sufficient statistic $T$ for some random variable $\theta$, the estimator $\tilde{\theta} := \mathbb{E}[\hat{\theta}|T]$ is unbiased and [CB01]

$$\mathbb{V}_\theta[\tilde{\theta}] \leq \mathbb{V}_\theta[\hat{\theta}],$$

i.e. conditioning an estimator with a sufficient statistic reduces variance. This notion can be applied to Bayesian particle filtering by finding a decomposition of the latent state model into a part that can be integrated exactly (with a Kalman filter) and a remaining (now smaller) part to which particle filtering is applied. This is the idea underlying Rao–Blackwellized particle filtering [DGA00; Dou+00].

### 1.9.3   Gaussian Sum and Mixture Models

To extend the class of models under consideration, it is common to reformulate a Gaussian filter to handle a mixture of Gaussians. The filter then propagates each Gaussian component separately, allowing for a richer representation of the posterior distribution [SA71; AS72; TPH99; CL00; Ter+11]. This approach has even be combined with other methods, such as the Gaussian sum particle filter [KD03b] and the Gaussian-sum Quadrature Kalman Filter [AHE07].

### 1.9.4   Dual Filtering

It is sometimes desirable to infer states and update model parameters simultaneously. Dual (Extended) Kalman filtering refers to a method to accomplish this [NW97; WN97; WMN00].

# FILTERING WITH A DISCRIMINATIVE MODEL: THE DISCRIMINATIVE KALMAN FILTER (DKF)

> „Gerettet ist das edle Glied
> Der Geisterwelt vom Bösen,
> Wer immer strebend sich bemüht,
> Den können wir erlösen.“
>
> J. W. von Goethe, *Faust*

## 2.1   Preface

This chapter presents what Prof. Harrison and I currently believe about the DKF. D. Brandman had many insights during the development process and was instrumental in implementing the DKF for human neural decoding.

## 2.2   Introduction

Consider a state space model for $Z_{1:T} := Z_1, \ldots, Z_T$ (latent states) and $X_{1:T} := X_1, \ldots, X_T$ (observations) represented as a Bayesian network:

$$
\begin{array}{ccccccccc}
Z_1 & \longrightarrow & \cdots & \longrightarrow & Z_{t-1} & \longrightarrow & Z_t & \longrightarrow & \cdots & \longrightarrow & Z_T \\
\downarrow & & & & \downarrow & & \downarrow & & & & \downarrow \\
X_1 & & & & X_{t-1} & & X_t & & & & X_T
\end{array}
\tag{2.1}
$$

---

"This worthy member of the spirit world is rescued from the devil: for him whose striving never ceases we can provide redemption."

The conditional density of $Z_t$ given $X_{1:t}$ can be expressed recursively using the Chapman–Kolmogorov equation and Bayes' rule [see Che03, for further details]

$$p(z_t|x_{1:t-1}) = \int p(z_t|z_{t-1})p(z_{t-1}|x_{1:t-1}) \, dz_{t-1}, \tag{2.2a}$$

$$p(z_t|x_{1:t}) = \frac{p(x_t|z_t)p(z_t|x_{1:t-1})}{\int p(x_t|z_t)p(z_t|x_{1:t-1}) \, dz_t}, \tag{2.2b}$$

where $p(z_0|x_{1:0}) = p(z_0)$ and where the conditional densities $p(z_t|z_{t-1})$ and $p(x_t|z_t)$ are either specified *a priori* or learned from training data prior to filtering. Computing or approximating eq. (2.2) is often called *Bayesian filtering*. Bayesian filtering arises in a large number of applications, including global positioning systems, target tracking, computer vision, digital communications, and brain computer interfaces [Che03; BH12; BCH17].

Exact solutions to eq. (2.2) are only available in special cases, such as the Kalman filter [Kal60; KB61]. The Kalman filter models the conditional densities $p(z_t|z_{t-1})$ and $p(x_t|z_t)$ as linear and Gaussian:

$$p(z_t|z_{t-1}) = \eta_d(z_t; Az_{t-1}, \Gamma), \tag{2.3}$$

$$p(x_t|z_t) = \eta_m(x_t; Hz_t, \Lambda), \tag{2.4}$$

so that the posterior distribution $p(z_t|x_{1:t})$ is also Gaussian and quickly computable. Beneš [Ben81] and Daum [Dau84; Dau86] broadened the class of models for which for which the integrals in eq. (2.2) are analytically tractable, but many model specifications still fall outside this class. When the latent state space is finite, the integrals in eq. (2.2) become sums that can be calculated exactly using a grid-based filter [Ell94; Aru+02].

For more general models, variational techniques have been developed that find a closely-related tractable model and use it to approximate the integrals in eq. (2.2). For example, the extended Kalman filter (EKF) and the statistically-linearized filter fit a generic model to a linear model that then integrates exactly [Gel74; Sär13]. Laplace and saddle-point approximations fit a Gaussian to an integrand by matching the local curvature at the maximum [But07; Koy+10; QML15]. It is also possible to use Taylor series expansions, Fourier–Hermite series, or splines [SS12]. One issue with these approaches is that the approximating tractable models must be calculated online. The EKF requires a derivative to be evaluated and tends to be quite fast, but methods such as the iterated EKF [FB66;

WTA69] and the Laplace transform entail solving an optimization problem in order to compute each filter update [BC93].

Alternatively, a form for the posterior can be specified, and at each recursive update, the posterior can be approximated by a density closest to the required form. This is known as the Assumed Density Filter [Kus67; Min01b]. For example, when the specified form is Gaussian and the approximating density is chosen to minimize the KL-divergence between the true and approximate densities, the resulting filter estimates the posterior as a Gaussian having the same mean and variance as the calculated posterior [Sär13]. For general models, we note that this still entails integration to calculate those first two moments. Expectation Propagation extends Assumed Density Filtering with iterative refinement of estimates, but iterating over the entire history of observations is typically not practical in an online setting [Min01a; Min01b].

Solving the integrals in eq. (2.2) can be done using quadrature rules. A primary example here are sigma-point filters including the unscented Kalman filter [JU97; WM00; Mer04], Quadrature Kalman filter [Ito00; IX00] and Cubature Kalman filter [AHE07; AH09]. Under these models, integrals are approximated based on function evaluations at deterministic points.

The integrals can also be solved using Monte Carlo integration [MU49]. Such approaches are called sequential Monte Carlo or particle filtering and include Sequential Importance Sampling and Sequential Importance Resampling [HM69; Han70; GSS93; Kit96; Mor96; DGA00; CMR05; CGM07]. These methods apply to all classes of models but tend to be the most expensive to compute online and suffer from the curse of dimensionality [DH03]. Alternate sampling strategies [see, e.g., Che03; Liu08] can be used to improve filter performance, including: acceptance-rejection sampling [HM69], stratified sampling [DC05], hybrid MC [CF01], and quasi-MC [GC15]. There are also ensemble versions of the Kalman filter that are used to propagate the covariance matrix in high dimensions including the ensemble Kalman filter [enKF: Eve94] and ensemble transform Kalman filter [ETKF: BEM01; Maj+02], along with versions that produce local approximations for covariance and can be parallelized [Ott+02; Ott+04; HKS07].

In this paper we introduce another type of Gaussian approximation for eq. (2.2), called the *Discriminative Kalman Filter (DKF)*, that retains much of the computational simplicity of the Kalman filter, but that can perform well in several situations where existing approaches either fail or are too computationally demanding. In particular, calculating an update step for the DKF entails only function evaluation and matrix algebra: neither optimization nor integration is required while the filter is online.

The DKF retains the linear, Gaussian model for the state dynamics $p(z_t|z_{t-1})$, but uses a discriminative formulation $p(z_t|x_t)$ for incorporating new observations instead of the generative specification $p(x_t|z_t)$. Approximating $p(z_t|x_t)$ as Gaussian leads to a new filtering algorithm that can perform well even on models where $p(x_t|z_t)$ is highly nonlinear and/or non-Gaussian. The model is Gaussian, but in a fundamentally different way, and it is allowed to be nonlinear, i.e.,

$$p(z_t|x_t) = \eta_d(z_t; f(x_t), Q(x_t)) \tag{2.5}$$

where $f : \mathbb{R}^m \to \mathbb{R}^d$ and $Q : \mathbb{R}^m \to \mathbb{S}_d$, using $\mathbb{S}_d$ to denote the set of $d \times d$ covariance matrices. There are several advantages to this approach:

- There is an exact, closed form solution for $p(z_t|x_{1:t})$ using eq. (2.2) and the component densities specified in eq. (2.3) and eq. (2.5). This is true regardless of the functional form of the nonlinearities in $f(\cdot)$ and $Q(\cdot)$. See section 2.4. Note that a Gaussian Assumed Density Filter (ADF) under general specifications still entails integration for its update steps [Ito00; IX00].

- The Gaussian assumption in eq. (2.5), which relates to the states, is often much more natural than the one in eq. (2.4), which relates to the observations. This is particularly true when $m \gg d$. Under mild regularity assumptions, the Bernstein-von Mises Theorem states that $p(z_t|x_t)$ in equation eq. (2.5) is asymptotically normal (in total variation distance) as the dimensionality of $x_t$ increases. The observations themselves are not required to be conditionally Gaussian or even continuously-valued. For instance, in neural decoding, the observations are often counts of neural spiking events (action potentials), which might be restricted to small integers, or even binary-valued.

- The DKF subsumes the Kalman filter as a special case by restricting $f$ to be linear and $Q$ to be constant.

The DKF requires knowledge of the conditional mean and covariance of the latent state $Z_t$ given the observations $X_t$. In some cases this information can be derived or approximated from a known model. If the observation model is unknown and must be learned from supervised training data prior to filtering (as is the case for our motivating BCI application) then it can be advantageous to learn these conditional means and covariances directly using off-the-shelf nonlinear and/or nonparametric regression tools, thereby avoiding some of the challenges of nonparametric density estimation.

$$Z_{t-1}|X_{1:t-1} \xrightarrow{\eta_d(z_t;Az_{t-1},\Gamma)} Z_t \qquad\qquad Z_{t-1}|X_{1:t-1} \xrightarrow{\eta_d(z_t;Az_{t-1},\Gamma)} Z_t$$
$$\downarrow \eta_n(x_t;Hz_t,\Lambda) \qquad\qquad\qquad\qquad\qquad \downarrow \approx \frac{\eta_d(z_t;f(x_t),Q(x_t))}{\eta_d(z_t;0,S)}$$
$$X_t \qquad\qquad\qquad\qquad\qquad\qquad\qquad\qquad\qquad X_t$$

Figure 2.1 – On the left, we have the Kalman filter that takes both the measurement and state models to be linear, Gaussian. On the right, we have the DKF that approximates $p(x_t|z_t)$ using Bayes rule and a nonlinear Gaussian approximation for $p(z_t|x_t)$. For both of these filters, if $Z_{t-1}|X_{1:t-1}$ is normally distributed, then $Z_t|X_{1:t}$ can also be computed as normal with a closed-form recursion.

## 2.3 Motivating the DKF

Our motivating application for the development of the DKF is neural decoding for closed-loop brain computer interfaces (BCIs). BCIs use neural information recorded from the brain for the voluntary control of external devices [Wol+02; HD06; BCH17]. Intracortical BCI systems (iBCIs) have been shown to provide paralyzed users the ability to control computer cursors [Pan+15; Jar+15], robotic arms [Hoc+12], and functional electrical stimulation systems [Bou+16; Aji+17] with their thoughts. State-of-the-art decoding approaches have been based on the Kalman filter [Pan+17; Jar+15; Gil+15], which learns a linear model between neural features (observed) and motor intention (latent). Motor intentions are inferred from training data as vectors from the instantaneous cursor position to the target position $Z_t$ [Bra+18b].

The DKF is a natural choice for closed-loop neural decoding using iBCIs placed in the primary motor cortex for a few reasons. First, evidence suggests that neurons may have very complex behavior. Neurons in the motor cortex have been shown to encode direction of movement [GKS88], velocity [Sch94], acceleration [Pan+04], muscle activation [Lem08;

Poh+07], proprioception [BM14], visual information related to the task [RD14] and preparatory activity [Chu+12]. In particular, iBCI-related recordings are highly complex and their relationship with user intention may be highly non-linear [Var+15]. Moving away from the linear constraints of the Kalman filter could potentially capture more of the inherent complexity of the signals, resulting in higher end-effector control for the user.

Second, evidence suggests that the quality of control is directly related to the rate at which the decoding systems perform real-time decoding. Modern iBCI sytems update velocity estimates on the order of 20ms [Jar+15] or even 1ms [Pan+15]. Thus any filtering technique must be computationally feasible to implement for real-time use.

Third, over the past decades, new technologies have allowed neuroscientists to record simultaneously from increasingly large numbers of neurons. The dimensionality of observed brain signals has been growing exponentially [SK11]. By contrast, the dimensionality of the underlying device being controlled remains small, generally not exceeding ten dimensions [Wod+15; Var+10].

We have previously reported how three people with spinal cord injuries could use the DKF with Gaussian process regression to rapidly gain closed-loop neural control [Bra+18b]. Here, we present data with a person with amyotrophic lateral sclerosis (participant T9) using the DKF with Gaussian process regression (See Section 2.9).

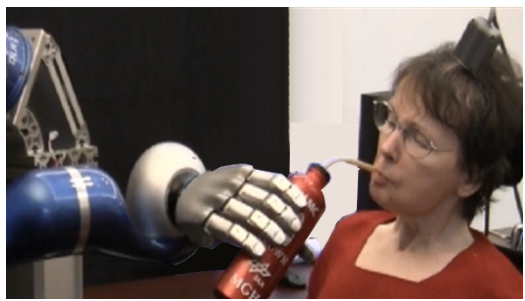

Figure 2.2 – Cathy Hutchinson used the BrainGate system to control a robotic arm with mental imagery alone. Here she picks up a bottle of water and drinks from it [Ven12]. Image credit: Brain-Gate.

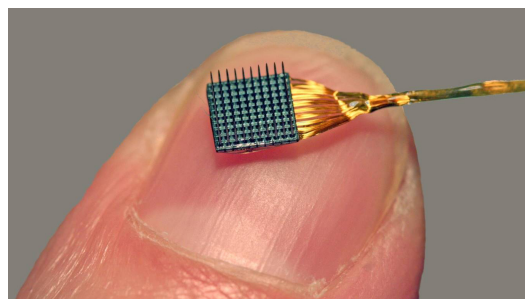

Figure 2.3 – The BrainGate project uses the Utah Array (Blackrock Microsystems, Salt Lake City, UT) to collect raw neural signals. Advances in technology have allowed for increasingly more detailed measurements. Image credit: BrainGate.

## 2.4   Filter Derivation

We derive the DKF under a simplified model for the latent states and discuss generalizations later. Let $\eta_d(z; \mu, \Sigma)$ denote the $d$-dimensional multivariate Gaussian distribution with mean vector $\mu \in \mathbb{R}^{d \times 1}$ and covariance matrix $\Sigma \in \mathbb{S}_d$ evaluated at $z \in \mathbb{R}^{d \times 1}$, where $\mathbb{S}_d$ denotes the set of $d \times d$ positive definite (symmetric) matrices. Assume that the latent states are a stationary, Gaussian, vector autoregressive model of order 1, namely, for $A \in \mathbb{R}^{d \times d}$ and $S, \Gamma \in \mathbb{S}_d$,

$$p(z_0) = \eta_d(z_0; 0, S), \tag{2.6a}$$

$$p(z_t|z_{t-1}) = \eta_d(z_t; Az_{t-1}, \Gamma), \tag{2.6b}$$

for $t = 1, 2, \ldots$, where $S = ASA^\top + \Gamma$, so that the process is stationary. Eq. (2.6) is the model for the latent states that underlies the stationary Kalman filter.

The observations take values in an abstract space $\mathcal{X}$. The observation model $p(x_t|z_t)$ is assumed to not vary with $t$, so that the joint $(Z_t, X_t)$ process is stationary, but it is otherwise arbitrary. It can be non-Gaussian, multimodal, discrete, etc. For instance, in neural decoding, the observations are often vectors of counts of neural spiking events (binned action potentials), which might be restricted to small integers, or even binary-valued.

The DKF is based on a Gaussian approximation for $p(z_t|x_t)$, namely,

$$p(z_t|x_t) \approx \eta_d(z_t; f(x_t), Q(x_t)), \tag{2.7}$$

where $f : \mathcal{X} \to \mathbb{R}^d$ and $Q : \mathcal{X} \to \mathbb{S}_d$. Note that eq. (2.7) is not an approximation of the observation model, but rather of the conditional density of the latent state given the observations at a single time step. When the dimensionality of the observation space ($\mathcal{X}$) is large relative to the dimensionality of the state space ($\mathbb{R}^d$), the Bernstein–von Mises theorem states that there exists $f$ and $Q$ such that this approximation will be accurate, requiring only mild regularity conditions on the observation model $p(x_t|z_t)$ [see Section 2.6.1 and Vaa98]. In this paper, we take $f$ and $Q$ to be the conditional mean and covariance of $Z_t$ given $X_t$, namely,

$$f(x) = \mathbb{E}(Z_t|X_t = x), \quad Q(x) = \mathbb{V}(Z_t|X_t = x), \tag{2.8}$$

where $\mathbb{E}$ and $\mathbb{V}$ denote expected value and variance/covariance, respectively.

To make use of eq. (2.7) for approximating eq. (2.2), we first rewrite eq. (2.2b) in terms of $p(z_t|x_t)$, namely

$$p(z_t|x_{1:t}) = \frac{p(z_t|x_t)p(z_t|x_{1:t-1})/p(z_t)}{\int p(z_t|x_t)p(z_t|x_{1:t-1})/p(z_t) \, dz_t}. \tag{2.9}$$

We then substitute the latent state model in eq. (2.6) and the DKF approximation in eq. (2.5) into the filtering equations in Eqs. (2.2a) and (2.9), and absorb terms not depending on $z_t$ into a normalizing constant $\kappa$ to obtain

$$p(z_t|x_{1:t}) \approx \kappa(x_{1:t})\frac{\eta_d(z_t; f(x_t), Q(x_t))}{\eta_d(z_t; 0, S)} \cdot \int \eta_d(z_t; Az_{t-1}, \Gamma)p(z_{t-1}|x_{1:t-1}) \, dz_{t-1}. \tag{2.10}$$

If $p(z_{t-1}|x_{1:t-1})$ is approximately Gaussian, which it is for the base case of $t = 1$ from eq. (2.6a), then all of the terms on the right side of eq. (2.10) are approximately Gaussian. If these approximations are exact, we find that the right side of eq. (2.10) is again Gaussian, giving a Gaussian approximation for $p(z_t|x_{1:t})$.

Let

$$p(z_t|x_{1:t}) \approx \eta_d(z_t; \mu_t(x_{1:t}), \Sigma_t(x_{1:t})), \tag{2.11}$$

be the Gaussian approximation of $p(z_t|x_{1:t})$ obtained from successively applying the approximation in eq. (2.10). Defining $\mu_0 = 0$ and $\Sigma_0 = S$, we can sequentially compute $\mu_t = \mu_t(x_{1:t}) \in \mathbb{R}^{d \times 1}$ and $\Sigma_t = \Sigma_t(x_{1:t}) \in \mathbb{S}_d$ via

$$\boxed{\begin{aligned} M_{t-1} &= A\Sigma_{t-1}A^{\intercal} + \Gamma, \\ \Sigma_t &= (Q(x_t)^{-1} + M_{t-1}^{-1} - S^{-1})^{-1}, \\ \mu_t &= \Sigma_t(Q(x_t)^{-1}f(x_t) + M_{t-1}^{-1}A\mu_{t-1}). \end{aligned}} \tag{2.12}$$

The function $Q$ needs to be defined so that $\Sigma_t$ exists and is a proper covariance matrix. A sufficient condition that is easy to enforce in practice[†] is $Q(\cdot)^{-1} - S^{-1} \in \mathbb{S}_d$.

The DKF is encapsulated in eq. (2.12). The explicit computations in eq. (2.12) are simple and at least as fast as the Kalman filter. The power of the DKF, along with potential computational difficulties, comes from evaluating $f$ and $Q$. If $f$ is linear and $Q$ is constant, then the DKF is equivalent to the Kalman filter. More general $f$ and $Q$ allow the filter to depend nonlinearly on the observations, improving performance in many cases. If $f$ and $Q$ can be quickly evaluated and the dimension $d$ of $Z_t$ is not too large, then the DKF is fast enough for use in real-time applications, such as the BCI deocoding example below.

---

[†] For our experiments below, if $Q(x_t)^{-1} - S^{-1} \notin \mathbb{S}_d$ for some $x_t$, we set $S^{-1} = 0$ in eq. (2.12), i.e., we use the robust DKF algorithm for that time step (Section 2.6.2). This occurs only rarely in practice, as expected from the Bernstein–von Mises Theorem (see Section 2.6.1). Even without this result, the Law of Total Variance implies that $\mathbb{E}(S - Q(X_t)) = \mathbb{V}(Z_t) - \mathbb{E}(\mathbb{V}(Z_t|X_t)) = \mathbb{V}(\mathbb{E}(Z_t|X_t)) \in \mathbb{S}_d$.

## 2.5 Learning

The parameters in the DKF are $A$, $\Gamma$, $Q(\cdot)$, and $f(\cdot)$. ($S$ is specified from $A$ and $\Gamma$ using the stationarity assumption.) For many real-life problems, these parameters are unknown to us and must be learned from supervised training examples $\{(Z_i', X_i')\}_{i=1}^m$ assumed to be sampled from the underlying Bayesian network in eq. (2.1). When we learn DKF parameters, either from a known generating model or from supervised samples, we will denote the learned parameters $\hat{A}$, $\hat{\Gamma}$, $\hat{Q}(\cdot)$, and $\hat{f}(\cdot)$, respectively.

The model parameters are learned from training data and then fixed and evaluated on testing data. $A$ and $\Gamma$ are the parameters of a well-specified statistical model given by Equations 2.6a–2.3. In the experiments below we learn them from $(Z_{t-1}, Z_t)$ pairs using only Equation 2.3, which reduces to multiple linear regression and is a common approach when learning the parameters of a Kalman filter from fully observed training data [see, for example, Wu+02].

The parameters $f$ and $Q$ are more unusual, since they are not uniquely defined by the model, but are introduced via a Gaussian approximation in Equation 2.5. In cases where the model is known, it may be possible to directly calculate $f$ and $Q$ using eq. (2.8). When the model is not known, $f$ and $Q$ can be learned from a supervised training set. One possibility is to first learn $p(z_t|x_t)$, either directly or by learning the observation model $p(x_t|z_t)$ and using Bayes' rule, and then derive suitable functions $\hat{f}$ and $\hat{Q}$. An alternative approach, and the one we focus on here, is to learn $f$ and $Q$ directly from training data by assuming that Equation 2.5 holds exactly, so that $\hat{f}$ and $\hat{Q}$ are the conditional mean and covariance of $Z_t$ given $X_t$, namely,

$$\hat{f}(x) = \mathbb{E}(Z_t|X_t = x), \qquad \hat{Q}(x) = \mathbb{V}(Z_t|X_t = x), \qquad (2.13)$$

where $\mathbb{E}$ and $\mathbb{V}$ denote expected value and variance/covariance, respectively. The Bernstein–von Mises Theorem holds with this choice of $\hat{f}$ and $\hat{Q}$ under some additional mild regularity conditions [Vaa98]. Using Equation 2.13, we learn $f$ and $Q$ from $(Z_t, X_t)$ pairs ignoring the overall temporal structure of the data, which reduces to a standard nonlinear regression problem with normal errors and heteroscedastic variance. The conditional mean $\hat{f}$ can be learned using any number of off-the-shelf regression tools and then, for instance, $\hat{Q}$ can be learned from the residuals, ideally, using a held-out portion of the training data. We think that the ability to easily incorporate off-the-shelf discriminative learning tools into a closed-form filtering equation is one of the most exciting and useful aspects of this approach.

In the experiments below, we compare three different nonlinear methods for learning $f$: Nadaraya-Watson (NW) kernel regression, neural network (NN) regression, and Gaussian process (GP) regression. In all cases, we start with a training set $\{(Z_i', X_i')\}_{i=1}^{m}$ and form an estimate for $\mathbb{E}[Z|X = x]$. While we have found that these methods work well with the DKF framework, one could readily use any arbitrary regression model with normal errors. Depending on the problem domain and the how the observations are generated, one might also consider random forest models, k-nearest neighbors, support vector regression, or even a simple linear regression model using some nonlinear transform of the observed data.

### 2.5.1 Nadaraya-Watson Kernel Regression

The well-known Nadaraya-Watson kernel regression estimator [Nad64; Wat64]

$$\hat{f}(x) = \frac{\sum_{i=1}^{m} Z_i' \kappa_X(x, X_i')}{\sum_{i=1}^{m} \kappa_X(x, X_i')} \tag{2.14}$$

where $\kappa_X(\cdot, \cdot)$ is taken to be a Gaussian kernel. Our implementations of Nadaraya-Watson in the DKF used $\hat{f}$ as described in eq. (2.14). Bandwidth selection was performed by minimizing leave-one-out MSE on the training set.

### 2.5.2 Neural Network Regression

We can learn $f : \mathbb{R}^n \to \mathbb{R}^d$ as a neural network (NN). With mean squared error (MSE) as an objective function, we optimize parameters over the training set. Typically, optimization continues until performance stops improving on a validation subset (to prevent overfitting), but instead we use Bayesian regularization to ensure network generalizability [Mac92; FH97].

We implemented all feedforward neural networks with Matlab's Neural Network Toolbox R2017b. Our implementation consisted of a single hidden layer of tansig neurons trained via Levenberg-Marquardt optimization [Lev44; Mar63; HM94] with Bayesian regularization.

### 2.5.3 Gaussian Process Regression

Gaussian process (GP) regression [RW06] is another popular method for nonlinear regression. The idea is to put a prior distribution on the function $f$ and approximate $f$ with its posterior mean given training data. We will first briefly describe the case $d = 1$. We form an $m \times n$-dimensional matrix $X'$ by concatenating the $1 \times n$-dimensional vectors $X_i'$ and a $m \times d$-dimensional matrix $Z'$ by concatening the vectors $Z_i'$. We assume that $p(z_i'|x_i', f) = \eta(z_i'; f(x_i'), \sigma^2)$, where $f$ is sampled from a mean-zero GP with covariance kernel $K(\cdot, \cdot)$. Under this model,

$$\hat{f}(x) = \mathbb{E}(f(x)|Z', X') = K(x, X')(K(X', X') + \sigma^2 I_m)^{-1} Z', \qquad (2.15)$$

where $K(x, X')$ denotes the $1 \times m$ vector with $i$th entry $K(x, X_i')$, where $K(X', X')$ denotes the $m \times m$ matrix with $ij$th entry $K(X_i', X_j')$, where $Z'$ is a column vector, and where $I_m$ is the $m \times m$ identity matrix. The noise variance $\sigma^2$ and any parameters controlling the kernel shape are hyperparameters. For our examples, we used the radial basis function kernel with two parameters: length scale and maximum covariance. These hyperparameters were selected via maximum likelihood. For $d > 1$, we repeated this process for each dimension to separately learn the coordinates of $f$.

All GP training was performed using the publicly available GPML package [RN10].

### 2.5.4 Learning $Q(\cdot)$

In all cases, we learned the function $Q$ as follows. We consider the random variable $R_t = Z_t - f(X_t)$ and want to learn

$$\hat{Q}(x) = \mathbb{E}(R_t R_t^\mathsf{T} | X_t = x).$$

Written in this way, we see that $\hat{Q}(x)$ is a conditional expectation and can in principle be learned with regression from $(\mathrm{vec}(R_t R_t^\mathsf{T}), X_t)$ pairs, where the vectorization of a matrix $\mathrm{vec}(M)$ simply concatenates the columns into a vector. Since $R_t$ is not observable (because $f$ is unknown), we instead approximate $R_t$ using the learned $f$. Given a training set $\{(Z_i'', X_i'')\}_{i=1}^m$ distinct from the one used to learn the function $f$, we define the residuals $\hat{R}_i = Z_i'' - \hat{f}(X_i'')$, and then learn $Q$ using NW regression on the $\hat{R}_i$'s of the training data. Because the NW estimator is a positive linear combination of valid covariance matrices, it will return a valid covariance matrix $Q(x)$ for any $x$. To avoid overfitting, particularly when $f$ is learned with NW regression, it is helpful to learn $f$ and $Q$ on distinct subsets of the training data. For the macaque example, the training data was randomly partitioned at a ratio of $70\% - 30\%$ into subsets: the first was used to learn $f$ and the second was used to learn $Q$.

## 2.6 Approximation Accuracy

### 2.6.1 Bernstein–von Mises theorem

Let the observation space be $X = \mathbb{R}^n$. As $n$ grows, the Bernstein–von Mises (BvM) theorem guarantees under mild assumptions that the conditional distribution of $Z_t | X_t$ is asymptotically normal in total variation distance and concentrates at $Z_t$ [Vaa98]. This asymptotic normality result is the main rationale for our key approximation expressed in eq. (2.5). The BvM theorem is usually stated in the context of Bayesian estimation. To apply it in our context, we equate $Z_t$ with the parameter and $X_t$ with the data, so that $p(z_t | x_t)$ becomes the posterior distribution of the parameter. Then we consider the situation where the dimension $n$ of $x_t$ grows, meaning that we are observing growing amounts of data associated with the parameter $Z_t$.

One concern is that eq. (2.9) will amplify approximation errors. Along these lines, we prove the following result that holds whenever the BvM theorem is applicable for eq. (2.5):

**Theorem.** *Under mild assumptions, the total variation between the DKF approximation for $p(z_t|x_{1:t})$ and the true distribution converges in probability to zero as $n \to \infty$.*

This result is stated formally and proven in the appendix. We interpret the theorem to mean that under most conditions, as the dimensionality of the observations increases, the approximation error of the DKF tends to zero.

The proof is elementary, but involves several subtleties that arise because of the $p(z_t)$ term in the denominator of eq. (2.9). This term can amplify approximation errors in the tails of $p(z_t|x_t)$, which are not uniformly controlled by the asymptotic normality results in the BvM theorem. To remedy this, our proof also uses the concentration results in the BvM theorem to control pathological behaviors in the tails. As an intermediate step, we prove that the theorem above still holds when the $p(z_t)$ term is omitted from the denominator of eq. (2.9).

### 2.6.2   Robust DKF

Omitting the $p(z_t)$ from the denominator of eq. (2.9) is also helpful for making the DKF robust to violations of the modeling assumptions and to errors introduced when $f$ and $Q$ are learned from training data. Repeating the original derivation, but without this denominator term gives the following filtering algorithm, that we call the *robust DKF*. Defining $\mu_1(x_1) = f(x_1)$ and $\Sigma_1(x_1) = Q(x_1)$, we sequentially compute $\mu_t$ and $\Sigma_t$ for $t \geq 2$ via

$$\begin{aligned}
M_{t-1} &= A\Sigma_{t-1}A^\top + \Gamma, \\
\Sigma_t &= (Q(x_t)^{-1} + M_{t-1}^{-1})^{-1}, \\
\mu_t &= \Sigma_t(Q(x_t)^{-1}f(x_t) + M_{t-1}^{-1}A\mu_{t-1}).
\end{aligned} \tag{2.16}$$

Justification for the robust DKF comes from our theoretical results above showing that the robust DKF accurately approximates the true $p(z_t|x_{1:t})$ in total variation distance as $n$ increases. We routinely find that the robust DKF outperforms the DKF on real-data examples, but not on simulated examples that closely match the DKF assumptions.

## 2.7 More on the Function $Q(\cdot)$

For random variables $Z \sim \mathcal{N}_d(0, S)$ and $X \in \mathbb{R}^m$, suppose there exist functions $g : \mathbb{R}^m \to \mathbb{R}^d$ and $R : \mathbb{R}^m \to \mathbb{S}_d$ such that

$$p(x|z) = \eta_d(g(x); z, R(x)),$$

where $\eta_d(y; \mu, \Sigma)$ denotes the $d$-dimensional Gaussian density with mean $\mu \in \mathbb{R}^d$ and covariance $\Sigma \in \mathbb{S}_d$ evaluated at $y \in \mathbb{R}^d$. *The issue is that there may not exist many interesting examples of such functions that yield a proper conditional density for $X|Z = z$ for all $z$.* Then

$$
\begin{aligned}
p(x, z) &= p(x|z)\, p(z) \\
&= \eta_d(z; g(x), R(x))\, \eta_d(z; 0, S) \\
&= \eta_d(g(x); 0, R(x) + S)\, \eta_d(z; \tilde{f}(x), \tilde{Q}(x)),
\end{aligned}
$$

where $\tilde{Q}(x) = (R(x)^{-1} + S^{-1})^{-1}$ and $\tilde{f}(x) = \tilde{Q}(x)R(x)^{-1}g(x)$. It follows that

$$
\begin{aligned}
p(x) &= \int p(x, z)\, dz \\
&= \eta_d(g(x); 0, R(x) + S) \int \eta_d(z; \tilde{f}(x), \tilde{Q}(x))\, dz \\
&= \eta_d(g(x); 0, R(x) + S)
\end{aligned}
$$

so we have

$$
\begin{aligned}
p(z|x) &= \frac{p(x, z)}{p(z)} \\
&= \eta_d(z; \tilde{f}(x), \tilde{Q}(x)).
\end{aligned}
$$

If our functions $\tilde{f}, \tilde{Q}$ arrive to us in this way, we can see why it is important to stipulate that $(\tilde{Q}(x)^{-1} - S^{-1})^{-1} \in \mathbb{S}_d$ or equivalently that $\tilde{Q}(x)^{-1} - S^{-1} \in \mathbb{S}_d$.

## 2.8 Examples

In this section, we compare filter performance on both artificial models and on real neural data. Corresponding MATLAB code (and Python code for the LSTM comparison) is freely

available online at:

`github.com/burkh4rt/Discriminative-Kalman-Filter`

For timing comparisons, the code was run on a Mid-2012 MacBook Pro laptop with a 2.9 GHz Intel Core i7 processor using MATLAB v. 2017b and Python v. 3.6.4.

### 2.8.1  The Kalman filter: a special case of the DKF

The stationary Kalman filter observation model is

$$p(x_t|z_t) = \eta_n(x_t; Hz_t, \Lambda)$$

for observations in $\mathcal{X} = \mathbb{R}^{n \times 1}$ and for fixed $H \in \mathbb{R}^{n \times d}$ and $\Lambda \in \mathbb{S}_n$. Defining $f$ and $Q$ via eq. (2.13) gives

$$f(x) = (S^{-1} + H^\intercal \Lambda^{-1} H)^{-1} H^\intercal \Lambda^{-1} x$$

and

$$Q(x) \equiv (S^{-1} + H^\intercal \Lambda^{-1} H)^{-1}.$$

It is straightforward to verify that the DKF in eq. (2.12) is exactly the well-known Kalman filter recursion. Hence, the DKF computes the exact posterior $p(z_t|x_{1:t})$ in this special case.

### 2.8.2  Kalman observation mixtures

In this section, we consider a family of models for which the Kalman, EKF, and UKF can perform arbitrarily poorly. Consider the observation model

$$p(x_t|z_t) = \sum_{\ell=1}^L \pi_\ell \eta_n(x_t; H_\ell z_t, \Lambda_\ell)$$

for a probability vector $\pi = \pi_{1:L}$. This a probabilistic mixture of Kalman observation models, one type of switching state space model [see SS91; GH00] At each time step, one of $L$ possible Kalman observation models is randomly and independently selected according to $\pi$ and then used to generate the observation. Defining $f$ and $Q$ via eq. (2.13) gives

$$f(x) = \frac{\sum_{\ell=1}^L w_\ell(x) V_\ell x}{\sum_\ell w_\ell(x)}$$

and

$$Q(x) = \frac{\sum_{\ell=1}^{L} w_\ell(x)(D_\ell + V_\ell x (V_\ell x)^\intercal)}{\sum_{\ell=1}^{L} w_\ell(x)} - f(x) f(x)^\intercal,$$

where $D_\ell = (H_\ell^\intercal \Lambda_\ell^{-1} H_\ell + S^{-1})^{-1}$, $V_\ell = D_\ell H_\ell^\intercal \Lambda_\ell^{-1}$, $G_\ell = H_\ell S H_\ell^\intercal + \Lambda_\ell$, and $w_\ell(x) = \pi_\ell \eta_n(x; 0, G_\ell)$.

Define $\bar{H} = \sum_{\ell=L} \pi_\ell H_\ell$ so that

$$\mathbb{E}(X_t | Z_t) = \bar{H} Z_t. \tag{2.17}$$

An interesting special case of this model is when $\bar{H} = 0$, so that the mean of $X_t$ given $Z_t$ does not depend on $Z_t$. Information about the states is only found in higher-order moments of the observations. Algorithms that are designed around $\mathbb{E}(X_t | Z_t)$, such as the Kalman filter, EKF, and UKF, are not useful when $\bar{H} = 0$, illustrating the important difference between a Gaussian approximation for the observation model and the DKF approximation in eq. (2.5).

Figure 2.4 shows how the DKF compares to a high-accuracy particle filter (PF) in a simple instance of this model with $d = 10$ hidden dimensionality, $L = 2$ categories, $\pi = (0.5, 0.5)$, $\Lambda_1 = I_n$, $\Lambda_2 = I_n/8$, and $H_2 = -H_1$, so that $\bar{H} = 0$. (The entries of $H_1$ were generated as independent normals.) $S$ is the identity matrix and $A = 0.91I - 0.1$ has off-diagonal elements, so that the coordinates of $Z_t$ are dependent. $\Gamma$ was chosen so that $S = ASA^\intercal + \Gamma$. The test set was 10000 time steps long.

We believe that the PF is a good approximation to the true posterior, which cannot be exactly computed (to our knowledge). As $n$ increases, the RMSE (root mean square error) performance of the DKF approaches that of the PF, but the DKF is many orders of magnitude faster (see Figure 2.5) . A PF that is constrained to run as fast as the DKF has worse performance.

### 2.8.3 Independent Bernoulli mixtures

Here we describe a model where observations live in $\{0, 1\}^n$. First, consider the case of a single hidden dimension $d = 1$. Let $-\infty = c_0 < c_1 < \cdots < c_n = \infty$ and define the discrete observation model

$$p(x_t | z_t) = \sum_{\ell=1}^{L} \pi_\ell \prod_{i=1}^{n} g_{\ell i}(z_t)^{x_{ti}} (1 - g_{\ell i}(z_t))^{1 - x_{ti}}$$

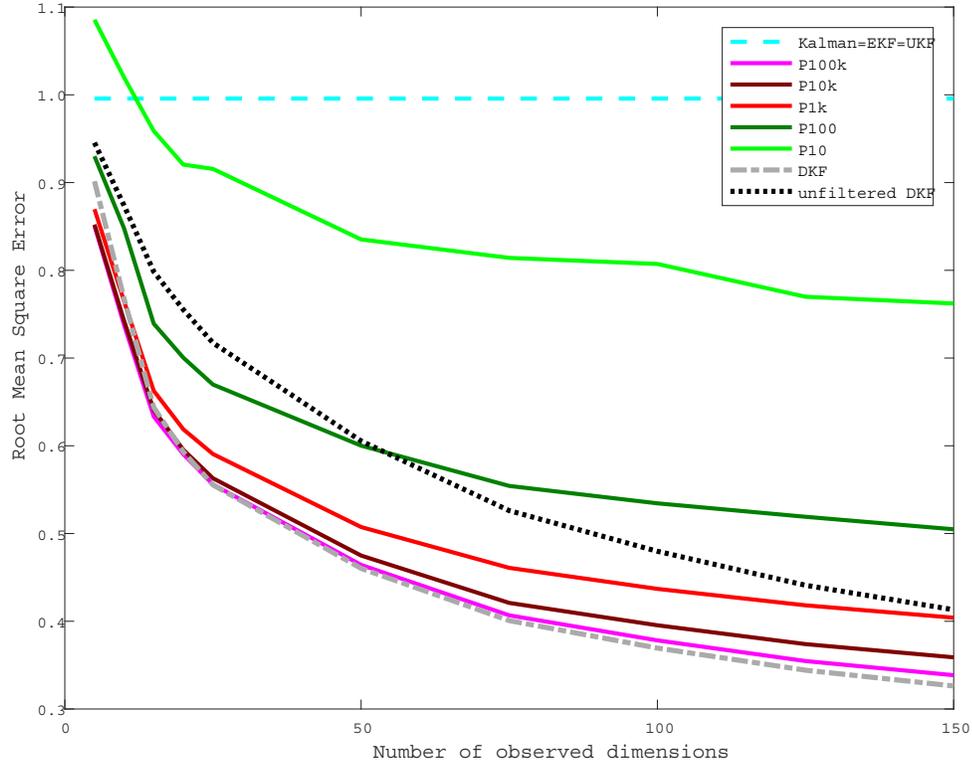

Figure 2.4 – We plot filtering performance (RMSE) on model 2.8.2 as more dimensions are revealed.

where

$$g_{\ell i}(z) = \alpha_{\ell i} + \beta_{\ell i}\mathbb{1}_{\{z \geq c_{i-1}\}},$$

with $\alpha_{\ell i} \geq 0$ and $0 \leq \alpha_{\ell i} + \beta_{\ell i} \leq 1$ for each $1 \leq \ell \leq L$ and $1 \leq i \leq n$. This is a probabilistic mixture of independent Bernoulli observation models. At each time step $t$, one of $L$ possible models is randomly selected and then used to generate $X_t$. If the $\ell$th model is selected, then the observations are generated as independent Bernoulli's with $P(X_{ti} = 1|Z_t) = g_{\ell i}(Z_t)$. In this case, if $\beta_{\ell i} > 0$, then $X_{ti}$ will be one more frequently when $Z_t \geq c_{i-1}$, and vice-versa for $\beta_{\ell i} < 0$. Define

$$\gamma_{i,\ell}(x) = \pi_\ell \left( \prod_{j \geq i}(\alpha_{\ell j} + \beta_{\ell j})^{x_j}(1 - \alpha_{\ell j} - \beta_{\ell j})^{1-x_j} \right) \left( \prod_{j < i}(\alpha_{\ell j})^{x_j}(1 - \alpha_{\ell j})^{1-x_j} \right)$$

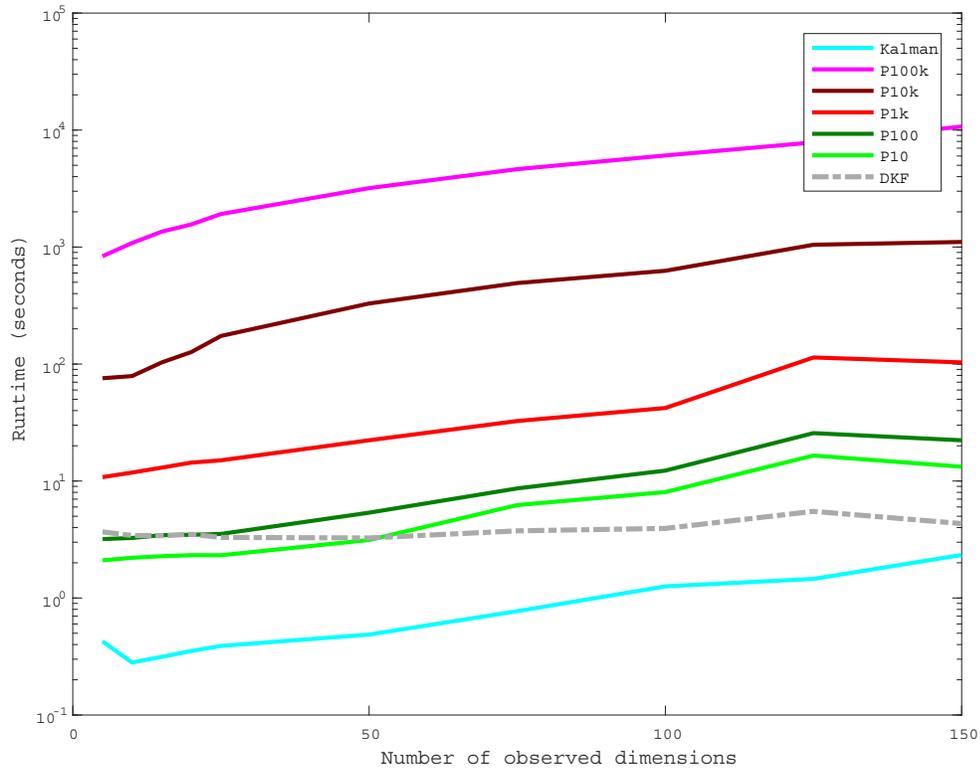

Figure 2.5 – Time (in seconds) to calculate all 10000 predictions on model 2.8.2 as more dimensions are revealed

so that

$$p(x_t|z_t) = \sum_{\ell=1}^{L} \gamma_{i,\ell}(x_t)$$

for $z_t$ in $C_i = [c_{i-1}, c_i)$. As the $C_i$ partition $\mathbb{R}$, for any measurable function $M : \mathbb{R} \to \mathbb{R}$ we can rewrite the following integral as a sum

$$\int M(z_t)p(x_t|z_t)p(z_t)dz_t = \sum_{i=1}^{n} \int_{C_i} M(z_t)p(x_t|z_t)p(z_t)dz_t$$
$$= \sum_{i=1}^{n} \left( \int_{C_i} M(z_t)p(z_t)dz_t \right) \left( \sum_{\ell=1}^{L} \gamma_{i,\ell}(x_t) \right).$$

The functions $f$ and $Q$ from eq. (2.13) can then be expressed as

$$f(x) = \frac{\sum_{i=1}^{n} \left( \int_{C_i} z p(z)dz \right) \left( \sum_{\ell=1}^{L} \gamma_{i,\ell}(x) \right)}{\sum_{i=1}^{n} \left( \int_{C_i} p(z)dz \right) \left( \sum_{\ell=1}^{L} \gamma_{i,\ell}(x) \right)}$$

and

$$Q(x) = \frac{\sum_{i=1}^{n} \left( \int_{C_i} zz^\intercal p(z) dz \right) \left( \sum_{\ell=1}^{L} \gamma_{i,\ell}(x) \right)}{\sum_{i=1}^{n} \left( \int_{C_i} p(z) dz \right) \left( \sum_{\ell=1}^{L} \gamma_{i,\ell}(x) \right)} - f(x)(f(x))^\intercal.$$

Note that the above integrals can be re-written in terms of the normal cdf and fully pre-computed before filtering starts.

Define $\bar{g}_i = \sum_\ell g_{\ell i}$, so that

$$\mathbb{E}(X_{ti} | Z_t = z) = \mathsf{P}(X_{ti} = 1 | Z_t = z) = \bar{g}_i(z).$$

An interesting special case of this model is when $\bar{g}_i$ is constant for each $i$, so that the individual components of $X_t$ carry no information about $Z_t$. The vector $X_t$ is needed for predicting $Z_t$. As in the previous section, filtering methods like the Kalman filter, EKF, and UKF are not useful when $\bar{g}_i$ is constant.

We can add observed dimensions to the model by refining the partition $c_0 < c_1 < \cdots < c_n$. This yields a model of the same form where the additional dimensions in $X_t$ provide more information about $Z_t$.

Figure 2.6 shows how the DKF compares to a high-accuracy particle filter in a simple instance of this model with $d = 3$, where each dimension of $Z_t$ was generated independently under an independent choice of model $\ell$ and the $X_t$ values for each dimension were concatenated. The single-dimensional model took $L = 2$, $\pi = (0.5, 0.5)$, $\alpha_{1i} = 0.001$, $\beta_{1i} = 0.998$, $\alpha_{2i} = 1 - \alpha_{1i}$, $\beta_{2i} = -\beta_{1i}$, so that $g_{2i}(z) = 1 - g_{1i}(z)$ and $\bar{g}_i(z) \equiv 0.5$, for all $i$. The partition $-\infty = c_0 < c_1 < \cdots < c_n = \infty$ was chosen so that $Z_t$ was equally likely to fall into any of the intervals, i.e. to make $\mathsf{P}(c_{i-1} \leq Z_t < c_i) = 1/n$ for all $i$. $S$ is the identity matrix and $A = 0.3I + 0.2$ has off-diagonal elements, so that the coordinates of $Z_t$ are dependent. $\Gamma$ was chosen so that $S = ASA^\intercal + \Gamma$.

This example emphasizes that the DKF can be used with highly non-Gaussian observation models, because its Gaussian approximation is in the state space, not the observation space. Again, PF's that run faster than the DKF perform significantly worse as the dimensionality of observations grows (see Figure 2.7).

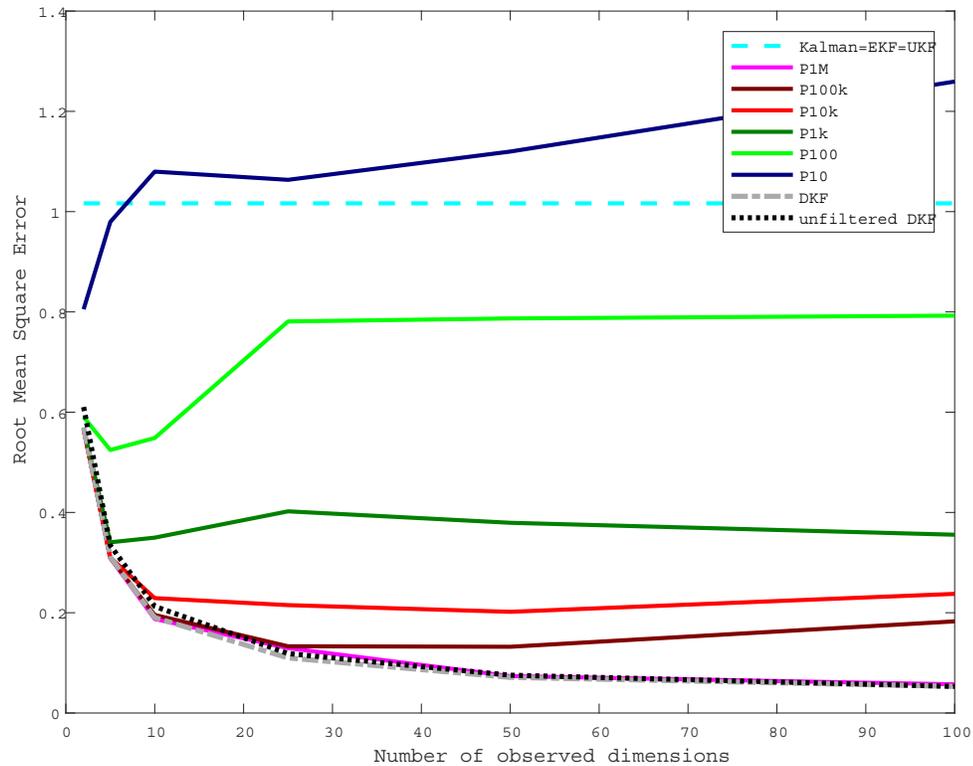

Figure 2.6 – We plot filtering performance (RMSE) on model 2.8.3 as more dimensions are revealed.

### 2.8.4 Unknown observation model: Macaque reaching task data

Flint et al. [Fli+12] implanted a rhesus monkey with a 96-channel microelectrode array (Blackrock Microsystems LLC) over the arm area of its primary motor cortex (M1). The monkey was trained to move a manipulandum to acquire illuminated targets for a juice reward. While performing this task, the monkey's neural spikes were recorded with a 128-channel acquisition system (Cerebus, Blackrock Microsystems LLC). The signal was sampled at 30 kHz, high-pass filtered at 300 Hz, and then manually thresholded and sorted into spikes offline. Walker and Kording [WK13] continue to make this data publicly available as part of the Database for Reaching Experiments and Models (DREAM). We took the data from Flint et al. [Fli+12] and took spike counts over 100ms bins. The first 10 PCA components of neural data became the observed variable.

Filtering results can be found in Table 2.1. We normalize RMSE by dividing it by the root mean square of the observation vector, so that predicting identically zero would yield a normalized RMSE of 1. We also present mean absolute angular error. Because cursor speed is often adjustable in BCIs, this may provide a more informative measure of performance.

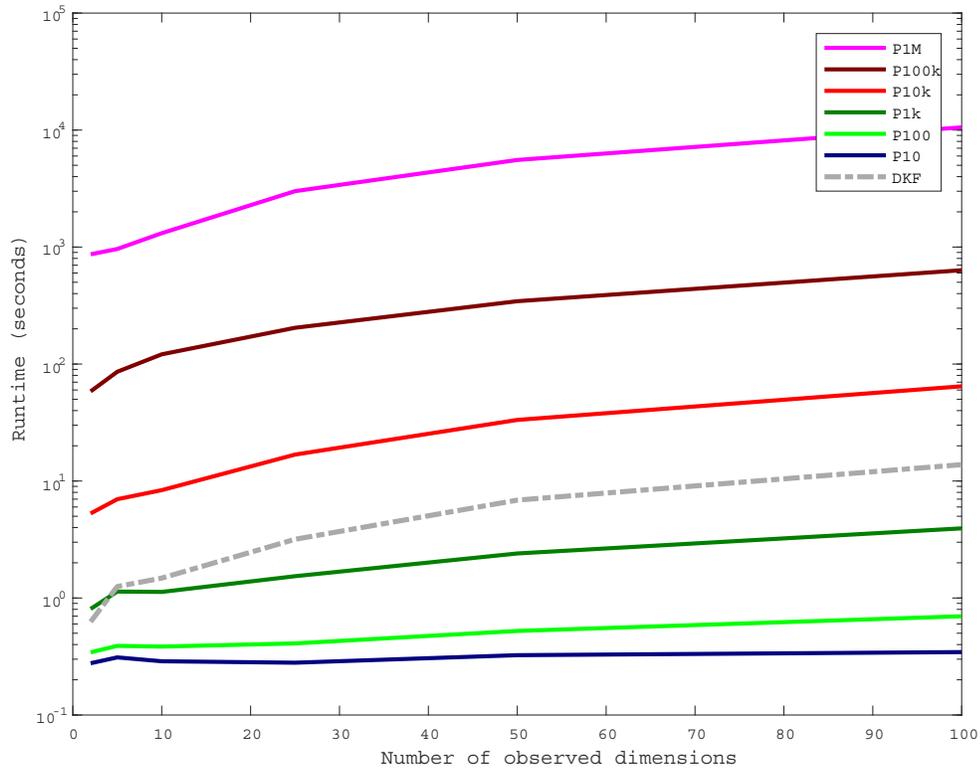

Figure 2.7 – Time (in seconds) to calculate all 1000 predictions on model 2.8.3 as more dimensions are revealed

### 2.8.5 Comparison with a Long Short Term Memory (LSTM) neural network

An LSTM is a stateful recurrent neural network designed to overcome error backflow problems [HS97]. Such recurrent neural networks have previously been shown to outperform state-of-the-art Kalman-based filters on this primate neural decoding task and so provide a good point of comparison [Sus+12; Sus+16]. While there are many variants on the LSTM architecture, none seem to universally improve on the basic design [JZS15; Gre+16].

All LSTM trials were conducted with TensorFlow r1.5 [Aba+15] in a Python 3.6.4 environment. The LSTM cell used in these trials was built from scratch in TensorFlow following [GSC00]. Dropout was used to prevent overfitting [Sri+14], but it was only applied to feedforward connections, not recurrent connections [Pha+14; ZSV14]. The recurrent states and outputs at each intermediate timestep were batch-normalized to accommodate internal covariate shift [IS15]. Model parameters were initialized via a Xavier-type method [GB10] designed to stabilize variance from layer to layer. Optimization was then performed with Adadelta [Zei12], an algorithm designed to improve upon Adagrad [DHS11] with the explicit goals of decreasing sensitivity to hyperparameters and permitting the learning rate to sometimes increase.

One advantage to the DKF approach is that good model selection requires minimal expert intervention. The LSTM and variants require manually selecting a neural network architecture. This is often done by experts through trial and error. Automating this process remains an area of active research requiring extensive computational resources [ZL17; Rea+17].

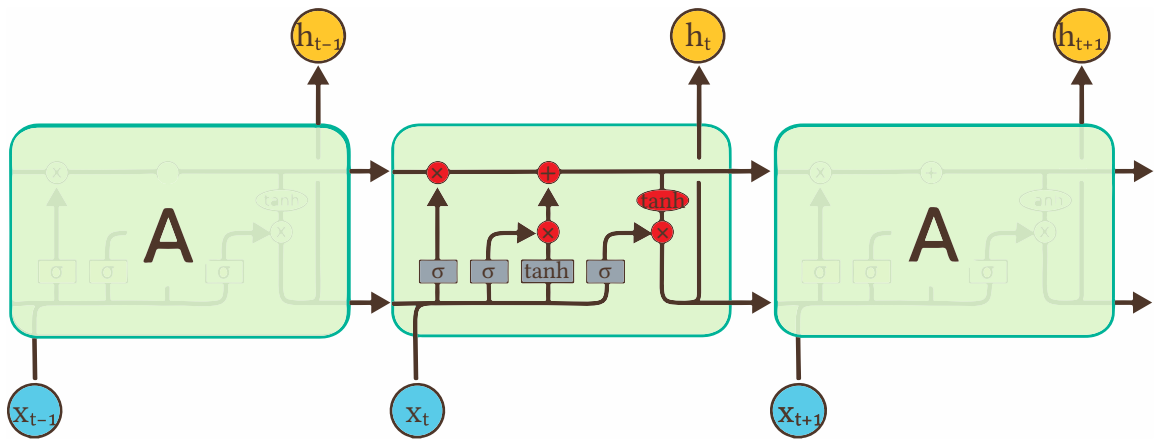

Figure 2.8 – This diagram gives a schematic representation of an LSTM cell. Re-used with permission: Olah [Ola15].

Table 2.1 – % Change in Mean Absolute Angular Error (Radians) Relative to Kalman

|  | Trial 1 | Trial 2 | Trial 3 | Trial 4 | Trial 5 | Trial 6 | Avg |
|---|---|---|---|---|---|---|---|
| Kalman | 0.889 | 0.955 | 1.025 | 0.933 | 0.964 | 0.926 | 0.949 |
| DKF-NW | -15% | -1% | -20% | -17% | -25% | -28% | -18% |
| DKF-NN | -7% | -2% | -17% | -16% | -21% | -23% | -14% |
| DKF-GP | -11% | 7% | -22% | -16% | -24% | -25% | -15% |
| UKF | 0% | 3% | -3% | -3% | -8% | -6% | -3% |
| EKF | 4% | 3% | -2% | -4% | -8% | -7% | -2% |
| LSTM | -2% | -2% | -12% | -6% | -10% | -8% | -7% |
| Unfiltered NW | -9% | 1% | -22% | -17% | -20% | -26% | -16% |
| Unfiltered NN | -3% | -0% | -17% | -14% | -17% | -17% | -11% |
| Unfiltered GP | -5% | 10% | -21% | -14% | -18% | -20% | -12% |

## 2.9 Closed-loop decoding in a person with paralysis

### 2.9.1 Participant

The participant in this study was T9, a 52 year-old right-handed male with paralysis from late stage amyotrophic lateral sclerosis (ALSFRS-score = 7). T9 underwent surgical placement of two 96-channel intracortical silicon microelectrode arrays [MNN97] (1.5-mm electrode length, Blackrock Microsystems, Salt Lake City, UT) in the primary motor cortex as previously described [Kim+08; Sim+11]. Data was used from trial (post-implant) days 292 and 293.

### 2.9.2 Signal acquisition

Raw neural signals for each channel (electrode) were sampled at 30kHz using the Neuro-Port System (Blackrock Microsystems, Salt Lake City, UT). Further signal processing and neural decoding were performed using the xPC target real-time operating system (Mathworks, Natick, MA). Raw signals were downsampled to 15kHz for decoding, and de-noised by subtracting an instantaneous common average reference [Gil+15; Jar+15] using 40 of the 96 channels on each array with the lowest root-mean-square value (selected based on their baseline activity during a one minute reference block run at the start of each session). The de-noised signal was band-pass filtered between 250 Hz and 5000 Hz using an 8th order non-causal Butterworth filter [Mas+15]. Spike events were triggered by crossing a threshold set at 3.5x the root-mean-square amplitude of each channel, as determined by data from the reference block. The neural features used was the the total power in the band-pass filtered signal [Jar+15; Bra+18b]. Neural features were binned in 20ms non-overlapping increments for decoding. We used the top 40 features ranked by signal-to-noise-ratio [Mal+15].

### 2.9.3 Decoder Calibration

Decoder calibration was performed using the standard Radial-8 task [Sim+11; Gil+15] using custom built software running Matlab (Natick, MA). An LCD monitor was placed 55-60 cm at a comfortable angle and orientation to T9. Targets (size = 2.4 cm, visual angle = 2.5°) were presented sequentially in a pseudo-random order, alternating between one of eight radially distributed targets and a center target (radial target distance from center = 12.1 cm, visual angle = 12.6°). Successful target acquisition required the user to place the cursor (size = 1.5cm, visual angle = 1.6°) within the target's diameter for 300ms, before a pre-determined timeout of 15 seconds. Target timeouts resulted in the cursor moving directly to the intended target, with immediate presentation of the next target.

Calibration began with two minute of open-loop presentation of a cursor; that is, the cursor moved automatically to pseudorandomly presented targets in a straight path. During this time, T9 was instructed to "imagine" or "attempt" to move the computer cursor as if he had control of it. After two minutes, initial coefficients were computed for the Gaussian process decoder. Next, T9 acquired targets for three minutes with 80% of the component of the decoded vector perpendicular to the vector between the cursor and the target [Jar+13; Vel+08]. Coefficients were then recomputed with all of the available data. The Radial-8 task was repeated two more times with the attenuated components at 50% and 20%, for a total of 11 minutes of calibration data collected. We collected a total of 3000 datapoints randomly subsampled from the 11 minutes of collected data, using all 192 neural features (96 features per array, two arrays).

### 2.9.4 Performance measurement

We quantified the performance of the DKF decoder with the mFitts1 task [Gil+15; Sim+11]. A single target was presented on the screen in a pseudorandom location, with one of three pseudorandomly fixed diameters (size = 1.6cm, 3.5cm, and 5.6cm, visual angles $1.7°$, $3.7°$, and $5.8°$). Targets were acquired by having the cursor contact the target for 500ms milliseconds, within a timeout of 10 seconds.

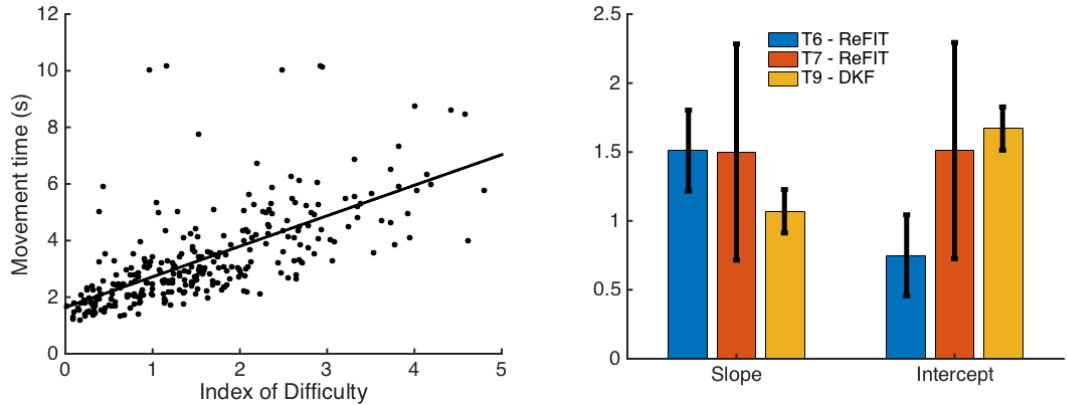

Figure 2.9 – Fitts plots comparing the DKF to Kalman ReFit

For the mFitts1 task, the Index of Difficulty for each trial was calculated as follows:

$$ID = \log_2 \left[ \frac{D}{W} + 1 \right] \tag{2.18}$$

where D is the distance from the cursor's start position to the goal, and W is the sum of the target's diameter and cursor's radius. Hence, $\frac{D}{W}$ reflects a measure of difficulty for acquiring targets.

### 2.9.5 Results

T9 acquired 98% of targets presented over two research sessions (N = 299) with the mFitts1 task. The Fitts regression parameters were comparable to the previously described performance using the ReFIT decoder [Gil+15] (Fig. 2.9, slope = $1.08 \pm 0.06p < 1.2 \times 10^{-30}$, intercept = $1.6 \pm 1.3, p < 2.2 \times 10^{-41}$).

## 2.10  Run Time

As code runtime can vary highly by choice of language and implementation details, we discuss the theoretical time-limiting steps involved for the different filtering methods and describe how these costs grow with model parameters and size of the training set.

### 2.10.1 Training Requirements

Training the DKF entails learning $f(\cdot) \in \mathbb{R}^d$ and $Q(\cdot) \in \mathbb{R}^{d \times d}$ from Equation 2.5. We consider how training costs grow with the number of training points $m$. If $f$ is learned with NW, bandwidth can be chosen using rule-of-thumb (free) or with leave-one-out cross validation scaling as $O(m^2)$. If $f$ is learned as a NN, training costs depend on the training algorithm chosen. Traditional optimizers include:s2: stochastic gradient descent, scaling with $O(m)$; scaled conjugate gradient, with $O(m^2)$; Levenberg–Marquardt, with $O(m^3)$ [Cas+10]. More recently, Hessian-free approaches have been developed to train NN's on larger data sets [Sch15]. Training costs also grow with $d$, depending on choice of architecture. If $f$ are learned as a GP, training costs scale as $O(m^3)$. Sparse approximations to GP's can reduce training requirements to $O(m \cdot N_S^2)$ where $N_S$ is the size of the sparse GP [QR05].

Fitting a linear model for the KF requires least squares regression, which tends to be fast. Training an EKF or UKF model entails learning the function $h$ and the noise covariance parameter $\Lambda$ in the measurement model $p(x_t|z_t) = \eta_m(h(z_t), \Lambda)$. If done with a NN, costs depend on the training algorithm chosen (see above). LSTM optimization uses many of the same methods that work for feedforward NN's [Sch15].

### 2.10.2 Prediction Requirements

An iteration of the DKF requires computing $f(\cdot) \in \mathbb{R}^d$ and $Q(\cdot) \in \mathbb{R}^{d \times d}$ along with a few inversions and multiplications in $\mathrm{Mat}(d \times d)$, where $d$ tends to be small. If $Q(\cdot)$ is calculated on held-out training data and then fixed, the posterior covariance quickly converges to a limiting value and can itself be fixed, further reducing computational costs.

The EKF and UKF both tend to be relatively fast, but they specify a very specific form for $p(x_t|z_t)$ and can perform rather poorly when that form is a poor approximation to the true model (for example, they are completely ineffective on the models described in sections 2.8.2 and 2.8.3).

Transitioning to more general nonlinear filtering often entails nonparametric methods with costs that scale with $m$. NW regression also scales $O(m)$ for evaluating $\hat{f}$. NN's scale $O(1)$ for evaluation. GP's scale $O(m)$ for $\hat{f}$ and their sparse approximations scale $O(N_S)$ where $N_S$ is the size of the sparse GP [QR05]. Among the non-probabilistic methods, evaluating a neural network scales $O(1)$ with training size, including the LSTM.

## 2.11 Discussion

The DKF is a novel approximation scheme that should be a helpful addition to the filtering toolbox. It provides a fast, analytic filtering approximation for models with linear, Gaussian dynamics, but nonlinear, non-Gaussian observations. The approximations underlying the DKF tend to improve as the dimensionality of the observation space increases relative to the dimensionality of the state space. Of existing filtering approximations, the DKF seems most closely related to Laplace approximations, or saddle-point approximations, but there are important differences. Laplace approximations are based on the local curvature of the posterior, whereas the DKF is a global approximation. Laplace approximations also involve an online optimization step that can be computationally demanding. The approximations underlying the DKF are conceptually distinct from those underlying the well-known EKF and UKF, and the methods are useful in different situations.

One potential drawback of the DKF is that it requires the conditional mean and variance of states ($Z_t$) given observations ($X_t$), which are often difficult to compute from the standard generative formulation of a state-space model. If, however, the model must be learned from supervised training data prior to filtering, then off-the-shelf nonlinear and/or nonparametric regression tools can be used to learn the conditional mean and variance directly, avoiding the more complicated task of learning the complete observation model $p(x_t|z_t)$. The benefits of this simplification are amplified in situations where the observation-dimensionality is much higher than the state-dimensionality. Using the DKF in this way appears to be novel within the large literature on learning state space models. Most approaches either learn a fully generative model and invert it for filtering (this includes the of use discriminative methods for training filters derived from generative models [Abb+05; HF09]), or learn a fully discriminative model that directly predicts states from the sequence of observations. The DKF allows a generative model for the state dynamics to be combined in principled way with a discriminative model for predicting the states from the observations at individual time steps. We think that the ability to easily incorporate off-the-shelf discriminative learning tools into a closed-form filtering equation is one of the most exciting and useful aspects of this methodology.

Another drawback of the DKF is its restriction to linear, Gaussian state dynamics. However, it is possible to use the discriminative measurement update approximation

$$p(x_t|z_t) = p(x_t)\frac{p(z_t|x_t)}{p(z_t)} \approx \kappa(x_t)\frac{\eta_d(z_t; f(x_t), Q(x_t))}{\eta_d(z_t; 0, S)} \tag{2.19}$$

in conjunction with the EKF or UKF method for propagating state dynamics. In the case that $p(z_t|z_{t-1})$ is nonlinear, it is worth noting that the denominator $\eta_d(z_t; 0, S)$ will no longer precisely correspond to $p(z_t)$ but will also be an approximation. If the Gaussian approximations for $p(z_t|x_t)$ and $p(z_t)$ are learned separately, some care may need to be taken to ensure the resulting approximation to $p(x_t|z_t)$ remains a good one. In Section 2.12, we illustrate an example where using a discriminative observation update with EKF/UKF state updates yields a much better filter than the standard EKF/UKF. Analogously, within a particle filter, particle weights can be updated using the discriminative approximation in eq. (2.19). This approach may be useful in situations where the observation model must be learned from data. In future work, we plan to explore this and other approaches that might allow a DKF-style approximation to be incorporated into more general filtering models.

The DKF approximation assumes a Gaussian posterior and is unlikely to work well in problems where it is important to maintain the full shape of a multimodal posterior. In situations with unknown models, however, there may be benefits to combining more accurate methods, like particle filtering, with alternatively-specified filtering equations, as in eq. (2.9), in order to create general purpose filters that are both more convenient to learn from data and more convenient to use in filtering applications. The DKF is a first step in this direction.

## 2.12   Example with Nonlinear State Dynamics

Consider the state model

$$p(z_t|z_{t-1}) = \eta_d(A(\sin(z_{t-1}) + z_{t-1}), \Gamma) \tag{2.20}$$

for $A \in \mathbb{R}^{d \times d}, \Gamma \in \mathbb{S}_d$ and observation model

$$p(x_t|z_t) = \eta_m(h(z_t) + z_t/3, \Lambda) \tag{2.21}$$

where $\Lambda \in \mathbb{S}_m$ and $h : \mathbb{R}^d \to \mathbb{R}^m$ concatenates the component-wise floor function $\lfloor z_t - a_j \rfloor$ over a set of $m/d$ elements $a_j \in \mathbb{R}^d$.

The stationary distribution of $Z_t$ is not Gaussian (and cannot be expressed analytically) so we learn $\hat{f}$ by generating ten thousand samples from the joint distribution of $(X_t, Z_t)$ and learning a NN. We compute the covariance of the residuals on heldout data and use this fixed value for $\hat{Q}$. Finally, $S$ can be approximated well using a Taylor series expansion for sine and the recursion eq. (2.20) or from samples. This gives us the necessary ingredients for our discriminative approximation

$$p(x_t|z_t) = p(x_t)\frac{p(z_t|x_t)}{p(z_t)} \approx \kappa(x_t)\frac{\eta_d(z_t; f(x_t), Q(x_t))}{\eta_d(z_t; 0, S)} \tag{2.22}$$

For this example, we used EKF/UKF approach to move from a Gaussian approximation of $p(z_{t-1}|x_{1:t-1})$ to a Gaussian approximation of $p(z_t|x_{1:t-1})$ and then the DKF approximation in eq. (2.22) to move from a Gaussian approximation of $p(z_t|x_{1:t-1})$ to a Gaussian approximation of $p(z_t|x_{1:t})$. Additionally, we implemented a SIR particle filter that used the true state dynamics and the DKF approximation eq. (2.22) for particle re-weighting. Results over five independent trials are summarized in Table 2.2.

Table 2.2 – Normalized RMSE for different filtering approaches to Model 2.12

|  | Trial 1 | Trial 2 | Trial 3 | Trial 4 | Trial 5 | Average |
|---|---|---|---|---|---|---|
| Particle Filter | 0.327 | 0.318 | 0.331 | 0.335 | 0.335 | 0.329 |
| PF with DKF re-weighting | 0.328 | 0.319 | 0.334 | 0.336 | 0.341 | 0.332 |
| EKF | 0.414 | 0.392 | 0.414 | 0.424 | 0.430 | 0.415 |
| EKF-state, DKF-observation | 0.328 | 0.319 | 0.333 | 0.336 | 0.340 | 0.331 |
| UKF | 0.476 | 0.471 | 0.444 | 0.480 | 0.491 | 0.472 |
| UKF-state, DKF-observation | 0.327 | 0.318 | 0.332 | 0.336 | 0.339 | 0.331 |
| Unfiltered | 0.352 | 0.341 | 0.355 | 0.362 | 0.355 | 0.353 |

The EKF and UKF filters can be implemented exactly for this model, but because $h'(z) = 0$ for almost every $z$, both filters perform quite poorly. This example serves to illustrate how a discriminatively-learned observation model can be successfully incorporated within standard filtering frameworks.

# CHAPTER 3

# DKF CONSISTENCY

> « Les choses ne peuvent être autrement : car, tout étant fait pour une
> fin, tout est nécessairement pour la meilleure fin. Remarquez bien que
> les nez ont été faits pour porter des lunettes, aussi avons–nous des
> lunettes. Les jambes sont visiblement instituées pour être chaussées,
> et nous avons des chausses. Les pierres ont été formées pour être
> taillées, et pour en faire des châteaux, aussi monseigneur a un très
> beau château ; le plus grand baron de la province doit être le mieux
> logé ; et, les cochons étant faits pour être mangés, nous mangeons du
> porc toute l'année : par conséquent, ceux qui ont avancé que tout est
> bien ont dit une sottise ; il fallait dire que tout est au mieux. »

F.-M. "Voltaire" Arouet, *Candide*

In this chapter, we prove the main result concerning the DKF's performance as the
number of observed dimensions tends to infinity.

## 3.1   The Bernstein–von Mises Theorem

For a random variable $\Theta \sim \text{Uniform}([0, 1])$, suppose we were to draw an infinite sequence
of random variables

$$X_1, X_2 \ldots \sim^{\text{i.i.d.}} \text{Bernoulli}(\Theta).$$

"Things cannot be other than as they are: for, since everything is made to serve an end, everything is
necessarily for the best of ends. Observe how noses were formed to support spectacles, therefore we have
spectacles. Legs are clearly devised for the wearing of breeches, therefore we wear breeches. Stones were
formed to be hewn and made into castles, hence his lordship's beautiful castle, for the greatest baron in
the province must perforce be the best housed; and since pigs were made to be eaten, we eat pork all year
round; consequently, those who have argued that is all well have been talking nonsense: they should have
said that all is for the best."

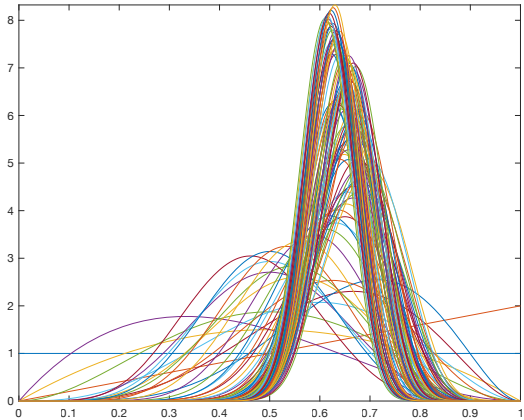

Figure 3.1 – I sampled $X_1, \ldots, X_{100} \sim^{\text{i.i.d.}}$ Bernoulli(0.6) and plotted the conditional densities $p(\theta | X_1 = x_1, \ldots, X_n = x_n)$ for $n = 1, \ldots, 100$ on the left. The bell shape is well-established by $n = 10$.

Then:

$$p(\theta | x_1, \ldots, x_n) \propto \theta^{n\bar{x}}(1 - \theta)^{n(1-\bar{x})} \mathbb{1}_{[0,1]}(\theta)$$

where $\bar{x} = \frac{1}{n} \sum_{i=1}^{n} x_i$ so that $\Theta | X_1, \ldots, X_n \sim \text{Beta}(n\bar{x} + 1, n(1 - \bar{x}) + 1)$. This distribution has mode $\bar{x}$ and variance bounded by $1/(n + 2) \to 0$. The distribution of the prior on $\Theta$ is flat on $[0, 1]$ and the distribution of $\Theta | X_1$ is a line on $[0, 1]$, but quite quickly $\Theta | X_1, \ldots, X_n$ looks bell-shaped, and vaguely Gaussian. This is not a coincidence. Loosely speaking, for any continuous random variable $\Theta$, if you draw a sequence $X_1, X_2 \ldots$ of random variables that each provide a bit more information about $\Theta$, then $\Theta | X_1, \ldots, X_n$ becomes Gaussian in the total variation metric.

Strictly speaking, we quote the following theorem from Vaart [Vaa98]:

*Let the experiment $(P_\theta : \theta \in \Theta)$ be differentiable in quadratic mean at $\theta_0$ with nonsingular Fisher information matrix $I_{\theta_0}$, and suppose that for every $\epsilon > 0$ there exists a sequence of tests $\phi_n$ such that:*

$$P_{\theta_0}^n \phi_n \to 0, \qquad \sup_{\|\theta - \theta_0\| \geq \epsilon} P_\theta^n(1 - \phi_n) \to 0$$

*Furthermore, let the prior measure be absolutely continuous in a neighborhood of $\theta_0$ with a continuous positive density at $\theta_0$. Then the corresponding posterior distributions satisfy*

$$\left\| P_{\sqrt{n}(\hat{\Theta}_n - \theta_0) | X_1, \ldots, X_n} - \mathcal{N}(\Delta_{n,\theta_0}, I_{\theta_0}^{-1}) \right\| \xrightarrow{P_{\theta_0}^n} 0$$

where $\Delta_{n,\theta_0} = \frac{1}{\sqrt{n}} \sum_{i=1}^n I_{\theta_0}^{-1} \dot{\ell}_{\theta_0}(X_i)$, $\dot{\ell}_{\theta_0}$ is the score function of the model, and the norm is that of total variation. A test is defined as "a measurable function of the observations taking values in the interval $[0,1]$." Total variation is invariant to location and scale changes, so it follows that

$$\left\| P_{\Theta|X_1,\ldots,X_n} - \mathcal{N}(\hat{\theta}_n, \tfrac{1}{n} I_\theta^{-1}) \right\| \xrightarrow{P_\theta^n} 0.$$

While this result can be used to justify statements along the lines of "the prior does not matter in Bayesian inference as the amount of data becomes large," we use this result instead to argue that, under relatively general conditions, $Z_t|X_t$ will become Gaussian as $\dim(X_t) \to \infty$. What follows now is justification for the argument that, if the conditions for the Bernstein–von Mises Theorem are satisfied, then the posterior computed by the DKF will become close to the true posterior in the total variation metric as $\dim(X_t) \to \infty$.

## 3.2  Proof of Theorem

Our main technical result is Theorem 1. After stating the theorem we translate it into the setting of the paper. Densities are with respect to Lebesgue measure over $\mathbb{R}^d$. $\|\cdot\|$ and $\|\cdot\|_\infty$ denote the $L_1$ and $L_\infty$ norms, respectively, $\to$ denotes weak convergence of probability measures (equivalent, for instance, to convergence of the expected values of bounded continuous functions), and $\delta_c$ denotes the unit point mass at $c \in \mathbb{R}^d$. Define the Markov transition density $\tau(y,z) = \eta_d(z; Ay, \Gamma)$, and let $\tau h$ denote the function

$$(\tau h)(z) = \int \tau(y,z) h(y) dy$$

for an arbitrary, integrable $h$. Define $p(z) = \eta_d(z; 0, S)$, where $S$ satisfies $S = ASA^\top + \Gamma$.

*Theorem 1.* Fix pdfs $s_n$ and $u_n$ ($n \geq 1$) so that the pdfs

$$p_n = \frac{u_n \tau s_n / p}{\|u_n \tau s_n / p\|} \tag{3.1}$$

are well-defined for each $n$. Suppose that for some $b \in \mathbb{R}^d$ and some probability measure $P$ over $\mathbb{R}^d$

A1.  $s_n \to P$ as $n \to \infty$;

A2. there exists a sequence of Gaussian pdfs $(s_n')$ such that $\|s_n - s_n'\| \to 0$ as $n \to \infty$;

A3. $u_n \to \delta_b$ as $n \to \infty$;

A4. there exists a sequence of Gaussian pdfs $(u_n')$ such that $\|u_n - u_n'\| \to 0$ as $n \to \infty$;

A5. $p_n \to \delta_b$ as $n \to \infty$;

Then

C1. $s_n' \to P$ as $n \to \infty$;

C2. $u_n' \to \delta_b$ as $n \to \infty$;

C3. the pdf

$$p_n' = \frac{u_n' \tau s_n' / p}{\|u_n' \tau s_n' / p\|}$$

is well defined and Gaussian for $n$ sufficiently large;

C4. $p_n' \to \delta_b$ as $n \to \infty$;

C5. $\|p_n - p_n'\| \to 0$ as $n \to \infty$.

*Remark* 2. We are not content to show the existence of a sequence of Gaussian pdfs $(p_n')$ that satisfy C4–C5. Rather, we are trying to show that the specific $p_n'$ defined in C3 satisfies C4–C5 regardless of the choice of $u_n'$ and $s_n'$.

*Remark* 3. An inspection of the proof shows that the pdf $r_n' = p_n' p / \|p_n' p\| = u_n' \tau s_n' / \|u_n' \tau s_n'\|$ is well-defined and Gaussian with $r_n' \to \delta_b$ and $\|p_n - r_n'\| \to 0$.

*Remark* 4. Suppose the pdfs $s_n, s_n', u_n, u_n'$ $(n \geq 1)$, the constant $b$, and the probability measure $P$ are themselves random, defined on a common probability space, so that $p_n$ is well-defined with probability one, and suppose that the limits in A1–A5 hold in probability. Then the probability that $p_n'$ is a well-defined, Gaussian pdf converges to one, and the limits in C1–C5 hold in probability.

For the setting of the paper, first fix $t \geq 1$ and note that $p$ is the common pdf of each $Z_t$ and $\tau$ is the common conditional pdf of $Z_t$ given $Z_{t-1}$. The limit of interest is for increasing dimension $(n)$ of a single observation. To formalize this, we let each $X_t$ be infinite dimensional and consider observing only the first $n$ dimensions, denoted $X_t^{1:n} \in$ $\mathbb{R}^n$. Similarly, $X_{1:t}^{1:n} = (X_1^{1:n}, \ldots, X_t^{1:n})$. We will abuse notation and use $\mathsf{P}(Z_t = \cdot | W)$ to denote the conditional pdf of $Z_t$ given another random variable $W$. These conditional pdfs (formally defined via disintegrations) will exist under very mild regularity assumptions [CP97]. Note that we are in the setting of Remark 4, where the randomness comes from $X_{1:t}, Z_{1:t}$. With this in mind, define

$$u_n(\cdot) = u_n(\cdot; X_t^{1:n}) = \mathsf{P}(Z_t = \cdot | X_t^{1:n})$$
$$u_n'(\cdot) = u_n'(\cdot; X_t^{1:n}) = \eta_d(\cdot; f_n(X_t^{1:n}), Q_n(X_t^{1:n}))$$
$$s_n(\cdot) = s_n(\cdot; X_{1:t-1}^{1:n}) = \mathsf{P}(Z_{t-1} = \cdot | X_{1:t-1}^{1:n}) \quad (t > 1)$$
$$s_n'(\cdot) = s_n'(\cdot; X_{1:t-1}^{1:n}) = \eta_d(\cdot; \mu_{t-1,n}(X_{1:t-1}^{1:n}), \Sigma_{t-1,n}(X_{1:t-1}^{1:n})) \quad (t > 1)$$
$$p_n(\cdot) = p_n(\cdot; X_{1:t}^{1:n}) = \mathsf{P}(Z_t = \cdot | X_{1:t}^{1:n})$$
$$p_n'(\cdot) = p_n'(\cdot; X_{1:t}^{1:n}) = \eta_d(\cdot; \mu_{t,n}(X_{1:t}^{1:n}), \Sigma_{t,n}(X_{1:t}^{1:n}))$$
$$b = Z_t$$
$$P(\cdot) = P(\cdot; Z_{t-1}) = \delta_{Z_{t-1}} \quad (t > 1),$$

and define $s_n \equiv s_n' \equiv P \equiv p$ when $t = 0$. The pdf $u_n'$ is our Gaussian approximation of the conditional pdf of $Z_t$ for a given $X_t^{1:n}$. We have added the subscript $n$ to $f$ and $Q$ from the main text to emphasize the dependence on the dimensionality of the observations. The pdfs $s_n'$ and $p_n'$ are our Gaussian approximations of $Z_{t-1}$ and $Z_t$ given $X_{1:t-1}^{1:n}$ and $X_{1:t}^{1:n}$, respectively. Again, we added the subscript $n$ to $\mu_t$ and $\Sigma_t$ from the text. Note that Equation 3.1 above is simply a condensed version of Equation (6) in the main text, and, for the same reason, the $p_n'$ defined in C3 is the same $p_n'$ defined above.

The Bernstein–von Mises (BvM) Theorem gives conditions for the existence of functions $f_n$ and $Q_n$ so that A3–A4 hold in probability. We refer the reader to Vaart [Vaa98] for details. Very loosely speaking, the BvM Theorem requires $Z_t$ to be completely determined in the limit of increasing amounts of data, but not completely determined after observing only a finite amount of data. The simplest case is when $X_t^{1:n}$ are conditionally iid given $Z_t$ and distinct values of $Z_t$ give rise to distinct conditional distributions for $X_t^{1:n}$, but the result holds in much more general settings. A separate application of the BvM Theorem gives A5 (in probability). In applying the BvM Theorem to obtain A5, we also obtain the existence of a sequence of (random) Gaussian pdfs $(p_n'')$ such that $\|p_n - p_n''\| \to 0$ (in probability), but we do not make use of this result, and, as explained in Remark 2, we care about the specific sequence $(p'_n)$ defined in C3.

As long as the BvM Theorem is applicable, the only remaining thing to show is A1–A2 (in probability). For the case $t = 1$, we have $s_n \equiv s'_n \equiv P \equiv p$, so A1–A2 are trivially true and the theorem holds. For any case $t > 1$, we note that $s_n$ and $s'_n$ are simply $p_n$ and $p'_n$, respectively, for the case $t - 1$. So the conclusions C4–C5 in the case $t - 1$ become the assumptions A1–A2 for the subsequent case $t$. The theorem then holds for all $t \geq 1$ by induction. The key conclusion is C5, which says that our Gaussian filter approximation $p'_n$ will be close in total variation distance to the true Bayesian filter distribution $p_n$ with high probability when $n$ is large.

*Proof of Theorem 1.* C1 follows immediately from A1 and A2. C2 follows immediately from A3 and A4. C3 and C4 are proved in Lemma 5 below. To show C5 we first bound

$$\|p_n - p'_n\| \leq \underbrace{\left\|p_n - \frac{p_n p}{p(b)}\right\|}_{A_n} + \underbrace{\left\|\frac{p_n p}{p(b)} - \frac{p_n p}{\|p_n p\|}\right\|}_{B_n}$$
$$+ \underbrace{\left\|\frac{p_n p}{\|p_n p\|} - \frac{p'_n p}{\|p'_n p\|}\right\|}_{C_n} + \underbrace{\left\|\frac{p'_n p}{\|p'_n p\|} - \frac{p'_n p}{p(b)}\right\|}_{B'_n} + \underbrace{\left\|\frac{p'_n p}{p(b)} - p'_n\right\|}_{A'_n}.$$

Since $p_n \to \delta_b$ and $p(z)$ is bounded and continuous,

$$A_n = \int p_n \left|1 - \frac{p}{p(b)}\right| = \mathbb{E}_{Z_n \sim p_n} \left|1 - \frac{p(Z_n)}{p(b)}\right| \to \left|1 - \frac{p(b)}{p(b)}\right| = 0$$

and

$$B_n = \int \frac{p_n p}{\|p_n p\|} \left|\frac{\|p_n p\|}{p(b)} - 1\right| = \left|\frac{\|p_n p\|}{p(b)} - 1\right| = \left|\frac{\mathbb{E}_{Z_n \sim p_n} |p(Z_n)|}{p(b)} - 1\right| \to \left|\frac{p(b)}{p(b)} - 1\right| = 0.$$

Similarly, since $p'_n \to \delta_b$,

$$A'_n = \int p'_n \left|1 - \frac{p}{p(b)}\right| = \mathbb{E}_{Z_n \sim p'_n} \left|1 - \frac{p(Z_n)}{p(b)}\right| \to \left|1 - \frac{p(b)}{p(b)}\right| = 0$$

and

$$B_n' = \int \frac{p_n'p}{\|p_n'p\|} \left| \frac{\|p_n'p\|}{p(b)} - 1 \right| = \left| \frac{\|p_n'p\|}{p(b)} - 1 \right| = \left| \frac{\mathbb{E}_{Z_n \sim p_n'} |p(Z_n)|}{p(b)} - 1 \right| \to \left| \frac{p(b)}{p(b)} - 1 \right| = 0.$$

All that remains is to show that $C_n \to 0$.

We first observe that

$$\frac{p_n p}{\|p_n p\|} = \frac{u_n \tau s_n}{\|u_n \tau s_n\|} \qquad \text{and} \qquad \frac{p_n'p}{\|p_n'p\|} = \frac{u_n' \tau s_n'}{\|u_n' \tau s_n'\|}.$$

Define

$$\alpha = \mathbb{E}_{(Y,Z) \sim P \times \delta_b} \, \eta_d(Z; AY, \Gamma) = \mathbb{E}_{Y \sim P} \, \eta_d(b; AY, \Gamma) \in (0, \infty).$$

Since $s_n \to P$, $u_n \to \delta_b$, and $(z,y) \mapsto \tau(y,z) = \eta_d(z; Ay, \Gamma)$ is bounded and continuous, we have

$$\|u_n \tau s_n\| = \iint \eta_d(z; Ay, \Gamma) s_n(y) u_n(z) dy \, dz = \mathbb{E}_{(Y_n, Z_n) \sim s_n \times u_n} \, \eta_d(Z_n; AY_n, \Gamma) \to \alpha.$$

Similarly, since $s_n' \to P$ and $u_n' \to \delta_b$,

$$\|u_n' \tau s_n'\| = \iint \eta_d(z; Ay, \Gamma) s_n'(y) u_n'(z) dy \, dz = \mathbb{E}_{(Y_n, Z_n) \sim s_n' \times u_n'} \, \eta_d(Z_n; AY_n, \Gamma) \to \alpha.$$

Defining $\beta = \eta_d(0; 0, \Gamma) \in (0, \infty)$, gives

$$\|\tau h\|_\infty \le \sup_z |(\tau h)(z)| \le \sup_{z,y} \eta_d(z; Ay, \Gamma) \int |h(t)| dt \le \eta_d(0; 0, \Gamma) \|h\| = \beta \|h\|$$

for any integrable $h$. With these facts in mind we obtain

$$C_n = \left\| \frac{u_n \tau s_n}{\|u_n \tau s_n\|} - \frac{u_n' \tau s_n'}{\|u_n' \tau s_n'\|} \right\| \leq \left\| \frac{u_n \tau s_n}{\|u_n \tau s_n\|} - \frac{u_n' \tau s_n}{\|u_n \tau s_n\|} \right\| + \left\| \frac{u_n' \tau s_n}{\|u_n \tau s_n\|} - \frac{u_n' \tau s_n'}{\|u_n' \tau s_n'\|} \right\|$$

$$\leq \frac{\|\tau s_n\|_\infty}{\|u_n \tau s_n\|} \|u_n - u_n'\| + \left\| \frac{\tau s_n}{\|u_n \tau s_n\|} - \frac{\tau s_n'}{\|u_n' \tau s_n'\|} \right\|_\infty \|u_n'\|$$

$$\leq \frac{\beta}{\|u_n \tau s_n\|} \|u_n - u_n'\| + \left\| \frac{\tau s_n}{\|u_n \tau s_n\|} - \frac{\tau s_n}{\|u_n' \tau s_n'\|} \right\|_\infty + \left\| \frac{\tau s_n}{\|u_n' \tau s_n'\|} - \frac{\tau s_n'}{\|u_n' \tau s_n'\|} \right\|_\infty$$

$$\leq \frac{\beta}{\|u_n \tau s_n\|} \|u_n - u_n'\| + \frac{\|\tau s_n\|_\infty}{\|u_n \tau s_n\|} \left| 1 - \frac{\|u_n \tau s_n\|}{\|u_n' \tau s_n'\|} \right| + \frac{\|\tau s_n - \tau s_n'\|_\infty}{\|u_n' \tau s_n'\|}$$

$$\leq \underbrace{\frac{\beta}{\|u_n \tau s_n\|}}_{\to \beta/\alpha} \underbrace{\|u_n - u_n'\|}_{\to 0} + \underbrace{\frac{\beta}{\|u_n \tau s_n\|}}_{\to \beta/\alpha} \underbrace{\left| 1 - \frac{\|u_n \tau s_n\|}{\|u_n' \tau s_n'\|} \right|}_{\to |1 - \alpha/\alpha| = 0} + \underbrace{\frac{\beta}{\|u_n' \tau s_n'\|}}_{\to \beta/\alpha} \underbrace{\|s_n - s_n'\|}_{\to 0}$$

Since $\alpha > 0$, we see that $C_n \to 0$ and the proof of the theorem is complete.

Remark 4 follows from standard arguments by making use of the equivalence between convergence in probability and the existence of a strongly convergent subsequence within each subsequence. The theorem can be applied to each strongly convergent subsequence. ⛿

*Lemma* 5 (DKF equation). If $s_n'(z) = \eta_d(z; a_n, V_n)$ and $u_n'(z) = \eta_d(z; b_n, U_n)$, then defining

$$p_n' = \frac{u_n' \tau s_n'/p}{\|u_n' \tau s_n'/p\|},$$

gives

$$p_n'(z) = \eta_d(z; c_n, T_n),$$

where $G_n = A V_n A^\top + \Gamma$, $T_n = (U_n^{-1} + G_n^{-1} - S^{-1})^{-1}$, and $c_n = T_n(U_n^{-1} b_n + G_n^{-1} A a_n)$, as long as $T_n$ is well-defined and positive definite. Furthermore, if $s_n' \to P$ and $u_n' \to \delta_b$, then $p_n'$ is eventually well-defined and $p_n' \to \delta_b$.

*Proof.* See above for the definition of $\tau$, $p$, $A$, $\Gamma$, $S$. Assuming $u_n' \tau s_n'/p$ is integrable, we have

$$p_n'(z) \propto \frac{\eta_d(z; b_n, U_n)}{\eta_d(z; 0, S)} \int \eta_d(z; Ay, \Gamma) \, \eta_d(y; a_n, V_n) \, dy.$$

Since

$$\int \eta_d(z; Ay, \Gamma)\, \eta_d(y; a_n, V_n)\, dy = \eta_d(z; Aa_n, AV_nA^\top + \Gamma) = \eta_d(z; Aa_n, G_n)$$

and

$$
\begin{aligned}
\frac{\eta_d(z; b_n, U_n)}{\eta_d(z; 0, S)} &\propto \frac{\exp(-\frac{1}{2}(z - b_n)^\top U_n^{-1}(z - b_n))}{\exp(-\frac{1}{2}z^\top S^{-1}z)} \\
&\propto \exp\left(-\frac{1}{2}(z^\top(U_n^{-1} - S^{-1})z - 2z^\top U_n^{-1}b_n)\right) \\
&\propto \exp\left(-\frac{1}{2}(z - b_n')^\top (U_n')^{-1}(z - b_n')\right) \\
&\propto \eta_d(z, b_n', U_n')
\end{aligned}
$$

for $U_n' = (U_n^{-1} - S^{-1})^{-1}$ and $b_n' = U_n' U_n^{-1} b_n$, we have

$$
\begin{aligned}
p_n'(z) &\propto \eta_d(z; b_n', U_n')\, \eta_d(z; Aa_n, G_n) \\
&\propto \eta_d(z; T_n((U_n')^{-1}b_n' + G_n^{-1}Aa_n), T_n) \\
&= \eta_d(z; c_n, T_n).
\end{aligned}
$$

As the normal density integrates to 1, the proportionality constant drops out.

Now, suppose additionally that $s_n' \to P$ and $u_n' \to \delta_b$. It is well known that this implies $a_n \to a$, $V_n \to V$, $b_n \to b$, and $U_n \to 0_{d \times d}$, where $a$ and $V$ are the mean vector and covariance matrix, respectively, of the distribution $P$, which must itself be Gaussian, although possibly degenerate. Thus, $G_n \to G = AVA^\top + \Gamma$, which is invertible, since $\Gamma$ is positive definite, and so $G_n^{-1} \to G^{-1}$.

The Woodbury matrix identity gives

$$T_n = (U_n^{-1} + G_n^{-1} - S^{-1})^{-1} = U_n - U_n((G_n^{-1} - S^{-1})^{-1} + U_n)^{-1}U_n.$$

Since $U_n \to 0_{d \times d}$ and $((G_n^{-1} - S^{-1})^{-1} + U_n)^{-1} \to G^{-1} - S^{-1}$, we see that $T_n \to 0_{d \times d}$.

To show $T_n$ is eventually well-defined and strictly positive definite, it suffices to show the same for

$$T_n^{-1} = U_n^{-1} + D_n$$

where we set $D_n = G_n^{-1} - S^{-1}$. For a symmetric matrix $M \in \mathbb{R}^{d \times d}$, let $\lambda_1(M) \geq \cdots \geq \lambda_d(M)$ denote its ordered eigenvalues. As a Corollary to Hoffman and Wielandt's result [see Cor. 6.3.8. in HJ13], it follows that

$$\max_j |\lambda_j(T_n^{-1}) - \lambda_j(U_n^{-1})| \leq \|D_n\|$$

where $\|D_n\| \to \|G^{-1} - S^{-1}\|$. Thus the difference between the $j$th ordered eigenvalues for $T_n^{-1}$ and $U_n^{-1}$ is bounded independently of $n$ for $1 \leq j \leq d$. Since $U_n$ is positive definite, it follows that

$$\lambda_d(U_n^{-1}) = 1/\lambda_1(U_n),$$

where $\lambda_1(U_n) = \|U_n\|_2$ is the spectral radius which is equivalent to the max norm. We see that $U_n \to 0_{d \times d}$ implies that the smallest eigenvalue for $U_n^{-1}$ becomes arbitrarily large. We conclude that all eigenvalues of $T_n^{-1}$ must eventually become positive, so that $T_n^{-1}$ becomes positive definite, hence also $T_n$.

For the means, we have

$$c_n = T_n U_n^{-1} b_n + T_n G_n^{-1} A a_n.$$

Because $T_n \to 0_{d \times d}$ and $G_n^{-1} \to G^{-1}$ we have $T_n G_n^{-1} A a_n \to \vec{0}$. Using the Woodbury identity for $T_n$,

$$T_n U_n^{-1} b_n = b_n - U_n((G_n^{-1} - S^{-1})^{-1} + U_n)^{-1} b_n,$$

where the eventual boundedness of $(G_n^{-1} - S^{-1})^{-1} + U_n)^{-1}$ implies

$$U_n((G_n^{-1} - S^{-1})^{-1} + U_n)^{-1} b_n \to \vec{0}.$$

As $b_n \to b$, we conclude $c_n \to b$. Hence, $p_n' \to \delta_b$.

# CHAPTER 4

# MAKING THE DKF ROBUST TO NONSTATIONARITIES

'Nihil est toto, quod perstet, in orbe.
cuncta fluunt, omnisque vagans formatur imago.'

F. "Ovid" Naso, *Metamorphoses*

## 4.1 Problem Description

To filter neural data for use in BCI's, we must model the relationship between the neural data and the desired responses. However, due to both physiological phenomena and engineering challenges, this relationship is constantly changing (nonstationary), sometimes over the course of mere hours [Per+13; Per+14]. To overcome this problem, two types of solutions have been proposed in the literature: (1) incorporating frequent filter retraining to regularly update the model for this relationship and (2) developing new models that that can be trained to ignore small changes in this relationship.

## 4.2 Approach I: Closed Loop Decoder Adaptation

Given a constant stream of new data, it is possible to retrain a filtering model at predetermined intervals. This method requires establishing ground truth for the desired responses. Any nonstationarity that develops must be present for some time (during which the decoder is presumably performing poorly) in order to catalog enough data so that it may be trained out at the next model refitting. This method does not require the user to predict

---

"In the whole of the world there is nothing that stays unchanged. All is in flux. Any shape that is formed is constantly shifting."

the nature of future nonstationarities and can adapt itself to drastic changes in the relationship between neural data and desired responses. Researchers have demonstrated BCI filter robustness with this meta-approach using a wide variety of specific methods: adapting a discriminative Bayesian filter [BCH17], refitting a Kalman filter [Gil+12; Dan+13], Bayesian updating for an unscented Kalman filter [Li+11], reweighting a naïve Bayes classifier [Bis+14], retraining a kernelized ARMA model [SLV09], and reinforcement learning [Mah+13; Poh+14], among others.

## 4.3 Approach II: Robust modeling

Training a robust model presents a relatively less-explored but very promising solution to the problem of nonstationarities. If changes in the relationship between neural data and desired response are predictable, then a filter can be designed to effectively ignore the expected changes. Such a filter can immediately adapt to small variability in this relationship and so does not require updating (or any feedback at all) in order to successfully decode in the presence of nonstationarities. Because linear models (like those used in the Kalman filter) have limited expressiveness and tend to underfit rich datasets [Sus+16], we propose instead two powerful models from machine learning that can learn an arbitrary functional relationship between neural data and the desired responses: stateful recurrent neural networks (RNN's) and Gaussian Processes (GP's).

Figure 4.1 – Data augmentation. A. We iterate over each (neural,intention)-pair in our training set. B. To protect against erratic behavior in the first dimension, we copy the (neural,intention)-pair, add noise to the first dimension, and add it to the training set with the same intention label. We repeat this process, adding different random noise each time. C. We repeat part (B) for each subsequent dimension in the neural space. This process can increase the size of the training set by orders of magnitude.

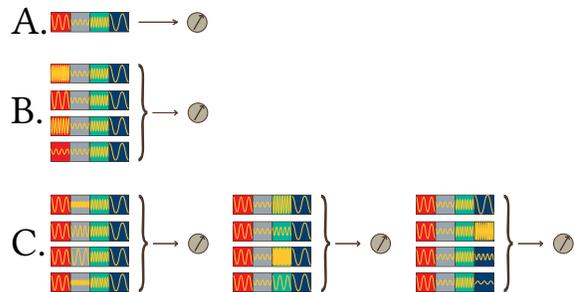

### 4.3.1 Stateful RNN's

Sussillo et al pioneered the training of fixed nonlinear BCI decoders for robustness in macaques [Sus+16]. To do so, they trained and successfully tested a multiplicative RNN-based neural filter. Sussillo et al achieved robustness by augmenting their training data with perturbations that mimicked the desired nonstationities against which they wished to train. For example, in order to train against dropping the $i$th neuron, they would add exemplars to their training set where firing from the $i$th neuron had been zeroed-out. This technique of augmenting a training set with noisy data is well-established for increasing generalization performance in neural networks [An96]. It requires generating and training over new artificial data for each individual nonstationarity they target. In particular, the exemplars to protect against dropping the $i$th neuron do not protect against dropping the $j$th neuron. Instead a new set of exemplars must be generated and trained on in order to achieve robustness for the $j$th neuron. It is easy to see how this technique often enlarges an already massive dataset by orders of magnitude, entailing a commensurate increase in training burden for the neural network.

### 4.3.2 GP's

In contrast to Sussillo, we achieve robustness through our choice of kernel, effectively altering the way we measure similarity between two firing patterns. Given a new vector $z_t$ of neural features, the Gaussian process model predicts the mean inferred velocity $f(z_t)$ as a linear combination (*cf.* [RW06] for an explanation):

$$f(z_t) = \sum_{i=1}^{n} \alpha_i K_\theta(\zeta_i, z_t)$$

where $\alpha := (K_\theta(\zeta, \zeta) + \theta_3^2 I)^{-1} \chi$. Thus the kernel $K_\theta$ evaluated on $\zeta_i$, $z_t$ directly determines the impact of the datapoint $(\zeta_i, \chi_i)$ on the prediction $f(z_t)$. We choose a kernel that ignores large differences between $\zeta_i$ and $z_t$ if they occur along only a relatively few number of dimensions. This makes our filter resilient to erratic firing patterns in an arbitrary single neuron (and presumably also in two or a few neurons, although we have not yet demonstrated this). In particular, we do not need to handle dropping neuron $i$ and dropping neuron $j$ separately. Altering our model to accommodate more or different nonstationarities would amount to a simple change in kernel and not result in increased training time.

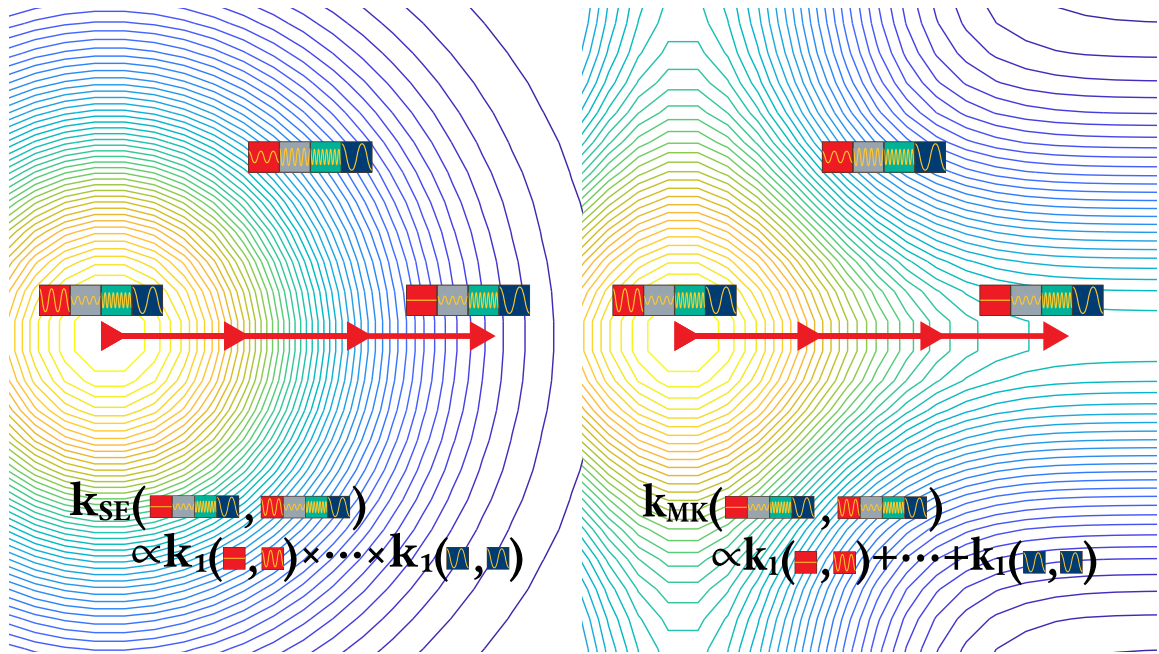

Figure 4.2 – On the left, the squared exponential kernel calculates the similarity between two vectors as the geometric mean of the similarities along each dimension. On the right, the multiple kernel calculates the similarity between two vectors as the arithmetic mean of the similarities along each dimension. The multiple kernel approach limits the impact a single dimension can have on the determination of how similar two vectors are.

## 4.4 Preface

The remainder of this chapter has been submitted to the Journal Neural Computation as an article entitled "Robust closed-loop control of a cursor in a person with tetraplegia using Gaussian process regression." David Brandman had the idea to explore robustness, conducted the online experiments with human volunteers, and wrote much of the paper. The kernel selection for the Gaussian process was worked out by myself and Prof. Harrison. We are grateful to our co-authors Prof. Hochberg, J. Kelemen, and B. Franco at BrainGate for making this investigation possible.

## 4.5    Abstract

Intracortical Brain computer interfaces can enable individuals with paralysis to control external devices. Decoding quality has been previously shown to degrade with signal nonstationarities, including changes to the neural tuning profiles, baseline shifts in firing rates, and non-physiological noise. While progress has been made towards providing long-term user control via decoder recalibration, relatively little work has been dedicated towards making the decoder itself more resilient to signal nonstationarities. Here, we describe how principled kernel selection with Gaussian process regression can be used within a Bayesian filtering framework to mitigate the effects of a commonly-encountered type of nonstationarity (large changes in a single neuron). Given a supervised training set of (neural,intention)-pairs, we use a Gaussian process regression with a specialized kernel (a multiple kernel) to estimate intention(t)|neural(t). This multiple kernel sums over each individual neural dimension, allowing the kernel to effectively ignore large differences in neural vectors that occur only in a single neuron. The summed kernel is used for real-time predictions of the posterior mean and variance of intention(t)|neural(t) using a Gaussian process regression framework. The Gaussian process predictions are then filtered using the discriminative Kalman filter to produce an estimate for intention(t)|neural(1:t). We refer to the multiple kernel approach within the DKF framework as the MK-DKF. We found that the MK-DKF decoder was more resilient to non-nonstationarities frequently encountered in-real world settings, yet provided similar performance to the currently used Kalman decoder. These results demonstrate a method by which neural decoding can be made more resistant to nonstationarities.

## 4.6    Introduction

Brain Computer Interfaces (BCIs) use neural information recorded from the brain for the voluntary control of external devices [Wol+02; Hoc+06; Sch+06; LN06; Fet07; Che+09; Car13]. At the heart of BCI systems is the decoder: the algorithm that maps neural information to generate a signal used to control external devices. Modern intracortical BCI decoders used by people with paralysis infer a relationship between neural features (e.g. neuronal firing rates) and the motor intentions from training data. Hence, high quality control of an external effector, such as a computer cursor, is predicated on appropriate selection of a decoding algorithm.

Decoder selection for intracortical BCI (iBCI) systems traditionally has been based on extensive study of cortical physiology. In what are now classic experiments, non-human primates (NHPs) were taught to move a planar manipulandum to one of eight different directions [Geo+82]. The firing rate as a direction of arm movement was parsimoniously modeled as a sinusoidal curve. For each neuron, the vector corresponding to the maximum firing rate (i.e. the phase offset of a cosine-function with a period of 360 degrees) is often referred to as the neuron's "preferred direction". The population vector algorithm scales the preferred directions of the recorded neurons by their recorded firing rate; the sum is the decoded vector of the intended direction of neural control [TTS02; Jar+08; Vel+08]. Given sufficient diversity in preferred directions, the problem reduces to linear regression: decoding involves learning the least-squares solution to the surface mapping firing rates to kinematic variables [KVB05]. Alternative decoding approaches include modeling the probability of observing a neural spike as a draw from from a time-varying poisson process [Tru+08; Bro+02; BTB14; Sha+17]; using support-vector regression [SLV08] or neural networks [Sus+16].

An ongoing area of research in iBCI systems is to ensure robust control for the user. Degradation in neural control is often attributed to nonstationarities in the recorded signals. With linear models, changes to the baseline firing rate, preferred direction, or depth of modulation result in degradation in decoding performance [Sus+15; Jar+15; Per+13]. The most common approach to addressing this mismatch involves recalibrating the decoder's parameters by incorporating more recent neural data. This has been described using batch-based updates during user-defined breaks [Jar+15; Bac+15; Gil+15], batch-based updates during ongoing use [Ors+14; SLV08], and continuous updating during ongoing use [Sha+14; SOC16; Dan+11]. Ongoing decoder recalibration traditionally requires information regarding the cursor's current location and a known target; alternatively, retrospective target inference has been described as a way to label neural data with the BCI users' intended movement directions based on selections during self-directed on-screen keyboard use [Jar+15].

While attempts to mitigate nonstationarities have largely focused on recalibration, few efforts have aimed to make the decoder inherently more resilient to nonstationarities.

To our knowledge, the most extensive study of examining decoder robustness investigated the use of deep neural networks trained from large amounts of offline data [Sus+16]. While effective for decoding, this method requires tremendous computational data and resources, and required the decoder to be specifically "trained" to handle nonstationarities using extensive regularization.

Here, we demonstrate a new decoding algorithm that is more resilient to signal nonstationarities than the currently used linear decoding models. Our approach builds on the previously well-established linear state-space dynamical model for neural decoding. Building upon prior work [Bra+18b], the key innovation is making a nonlinear decoder robust to noise through kernel selection and data sparsification. In this report, we have modified the kernel used for the Gaussian process decoder such that it is more resilient to nonstationarities. We refer to this new method as the MK-DKF method.

## 4.7    Mathematical Methods

We have previously described closed-loop decoding using Gaussian process regression (GP) in detail [Bra+18b]. Briefly: a collection of neural features, $\xi_i$, and inferred velocity vectors, $\zeta_i$, for $1 <= i <= n$, are collected during calibration. To perform closed-loop neural decoding at time step $t$, new neural features, $x_t$, are compared to $\xi_i$ according to a similarity metric (e.g. radial basis function). The $n$ similarities are then used to scale the corresponding $\zeta_i$ labels. The current unfiltered estimate, $f(x_t)$ for the expected value of $Z_t|X_t$ is given by the sum of the scaled $\zeta_i$ values. Filtering then produces an estimate $\mu_t$ for the mean of $Z_t|X_{1:t}$; this value $\mu_t$ is returned by the decoder.

Our previous approach to GP decoding used the entire high-dimensional neural dataset as the basis for computing the measure of similarity between $x_t$ and $\xi$ [Bra+18b]. In this report, we made two important changes to the decoder to increase its robustness to signal nonstationarities.

First, we adopted a kernel that calculated the similarity between two neural vectors as the arithmetic average over similarities in each neuron, as opposed to the product that was used by the more standard isotropic Gaussian kernel (see Section 4.7.2). This had the effect of limiting the impact any single neuron could have on the calculated similarity between two vectors of neural features. When a nonstationarity occurred in a feature, the decoder "disregarded" this feature without compromising decoding quality.

Second, we sparsified the data by averaging $(\xi, \zeta)$ pairs into octants. This dramatically decreased the computational load for real-time decoding. We found that the observed neural features had noise events with surprising frequency (see Results). Without data sparsification, the training features contained a large number of noisy features which were then used for decoding. Averaging across octants had the effect of mitigating the importance of these noisy features for decoding.

### 4.7.1 Description of decoding method

We model the latent state space model with hidden states $Z_1, \ldots, Z_T \in \mathbb{R}^d$ representing the intended cursor velocity, and the observed states $X_1, \ldots, X_T \in \mathbb{R}^m$ representing the neural features related through the following graphical model:

$$Z_1 \longrightarrow \cdots \longrightarrow Z_{t-1} \longrightarrow Z_t \longrightarrow \cdots \longrightarrow Z_T$$
$$\downarrow \qquad\qquad\qquad \downarrow \qquad\quad \downarrow \qquad\qquad\qquad \downarrow$$
$$X_1 \qquad\qquad\quad X_{t-1} \qquad X_t \qquad\qquad\quad X_T$$

In typical use, $d$ is 2 dimensional (e.g kinematic computer cursor control) while $m$ is 40 [Jar+13; Jar+15; Bac+15; Bra+18b]. Each dimension of $X$ corresponds to a neural feature. We are interested in the posterior distribution $p(z_t|x_{1:t})$ of the current hidden state given all observations up to present. Upon specifying the state model $p(z_t|z_{t-1})$ that relates how the hidden state changes over time and the measurement model $p(x_t|z_t)$ that relates the hidden and observed variables, the posterior can be found recursively using the Chapman–Kolmogorov equation

$$p(z_t|x_{1:t}) \propto p(x_t|z_t) \int p(z_t|z_{t-1})\, p(z_{t-1}|x_{1:t-1})\, dz_{t-1}, \tag{4.1}$$

where $\propto$ means proportional as a function of $z_t$. The standard Kalman filter is obtained when both the state and measurement models are specified as linear with Gaussian noise [Wu+05; Sim+11]. Here, we use a stationary, linear state model with Gaussian noise

$$p(z_0) = \eta_d(z_0; 0, S), \tag{4.2a}$$

$$p(z_t|z_{t-1}) = \eta_d(z_t; Az_{t-1}, \Gamma), \tag{4.2b}$$

where $A, S, \Gamma$ are $d \times d$, $S$ and $\Gamma$ are proper covariance matrices, $S = ASA^\mathsf{T} + \Gamma$, and $\eta_d(z; \mu, \Sigma)$ denotes the $d$-dimensional multivariate normal density with mean $\mu$ and covariance $\Sigma$ evaluated at a point $z$. We approximate the measurement model using Bayes' rule,

$$p(x_t | z_t) \propto \frac{p(z_t | x_t)}{p(z_t)} \approx \frac{\eta_d(z_t; f(x_t), Q)}{\eta_d(z_t; 0, S)}, \tag{4.3}$$

where $f : \mathbb{R}^m \rightarrow \mathbb{R}^d$ is a nonlinear function learned from training data and $Q$ is a $d \times d$ covariance matrix. The posterior is then given recursively by

$$p(z_t | x_{1:t}) \approx \eta_d(z_t; \mu_t, \Sigma_t), \tag{4.4}$$

where $\mu_1 = f(x_1)$, $\Sigma_1 = Q$, and for $t \geq 2$,

$$\begin{aligned} M_{t-1} &= A\Sigma_{t-1}A^\mathsf{T} + \Gamma, \\ \Sigma_t &= (Q^{-1} + M_{t-1}^{-1} - S^{-1})^{-1}, \\ \mu_t &= \Sigma_t(Q^{-1}f(x_t) + M_{t-1}^{-1}A\mu_{t-1}). \end{aligned} \tag{4.5}$$

In this way, we allow the relationship between $X_t$ and $Z_t$ to be nonlinear through the function $f$, while retaining fast, closed-form updates for the posterior. While $f$ can be learned from supervised training data using a number of off-the-shelf discriminative methods [Bur+16], in this paper we take $f$ to be the posterior mean from a Gaussian process regression, and set $Q$ as the covariance of the training dataset. We call the resulting filter the Discriminative Kalman Filter (DKF) [Bur+16].

### 4.7.2 Kernel Selection for Robustness

As part of decoder calibration, we collect a dataset consisting of neural features and intended velocities, which we refer to as $\{(\xi_i, \zeta_i)\}_{1 \leq i \leq n}$. These are assumed to be samples from the above graphical model and are used to train a Gaussian process regression for $Z_t | X_t$. The Gaussian process model takes asymmetric, positive-definite kernel $K_\theta(\cdot, \cdot)$ with hyperparameters $\theta$ and predicts the mean inferred velocity $f(x_t)$ as

$$f(x_t) = k_*^\mathsf{T}(K + \sigma_n^2 I_n)^{-1}\zeta, \tag{4.6}$$

where $K$ is the $n \times n$ matrix given component-wise by $K_{ij} = K_\theta(\xi_i, \xi_j)$, $\sigma_n^2$ is a noise parameter for the training data, $I_n$ is the $n$-dimensional identity matrix, and $\zeta$ is an $n \times 1$ vector with components $\zeta_i$. We can re-express eq. (4.6) as a linear combination (see [RW06] for details):

$$f(x_t) = \sum_{i=1}^{n} \alpha_i K_\theta(\xi_i, x_t), \tag{4.7}$$

where $\alpha = (K + \sigma_n^2 I_n)^{-1}\zeta$ so that $\alpha_i$ is a smoothed version of $\zeta_i$. This demonstrates how the kernel-determined similarity between $\xi_i$ and $x_t$ directly determines the impact of the training point $(\xi_i, \zeta_i)$ on the prediction $f(x_t)$.

In designing a kernel for robust decoding, we select a kernel that ignores large differences between $\xi_i$ and $x_t$ that occur along only a relatively few number of dimensions. This would potentially make the filter "resilient" to erratic firing patterns in an arbitrary single neuron. That is, a nonstationary shift in the mean firing rate of a single neuron would not result in degraded cursor control for the user.

We use a multiple kernel (MK) approach [GA11] and take

$$K_\theta(x, y) = \frac{1}{m} \sigma_f^2 \sum_{d=1}^{m} \eta_1(x^d - y^d; 0, \sigma_\ell^2), \tag{4.8}$$

where $\theta = (\sigma_f^2, \sigma_\ell^2)$ are hyperparameters and $x^d$ denotes the $d$-th dimension of $x$. The similarity between inputs $x$ and $y$ is given as the average over the similarities in each dimension, where all dimensions are equally informative.

To illustrate our choice of kernel, it is helpful to compare it against the more standard isotropic squared exponential kernel, where the sum in eq. (4.8) is replaced by a product, as follows

$$\tilde{K}_{\tilde{\theta}}(x, y) = \tilde{\sigma}_f^2 \prod_{d=1}^{m} \cdot \eta_1(x^d - y^d; 0, \tilde{\sigma}_\ell^2). \tag{4.9}$$

On identical inputs $x = y$, the $\tilde{K}$ and $K$ both return their maximum value of $\sigma_f^2$, indicating that $x$ and $y$ are similar. If, holding all other dimensions equal, the the absolute difference $|x^i - y^i|$ grows large (this would occur if readings from a single neuron became very noisy/unreliable), the standard kernel $\tilde{K}$ would become arbitrarily small while the multiple kernel $K$ would never fall below $\frac{m-1}{m}\sigma_f^2$. Thus, the multiple kernel continues to identify two neural vectors as close if they differ only along a single arbitrary dimension (Fig. 4.3 shows a visualization in two dimensions). Note that as *m* increases beyond two, this difference between the kernels becomes even more pronounced.

In contrast to data augmentation methods [An96; Sus+16], we do not need to handle dropping neuron *i* and dropping neuron *j* separately. Altering our model to accommodate more or different nonstationarities would amount to a simple change in kernel and not result in increased training time.

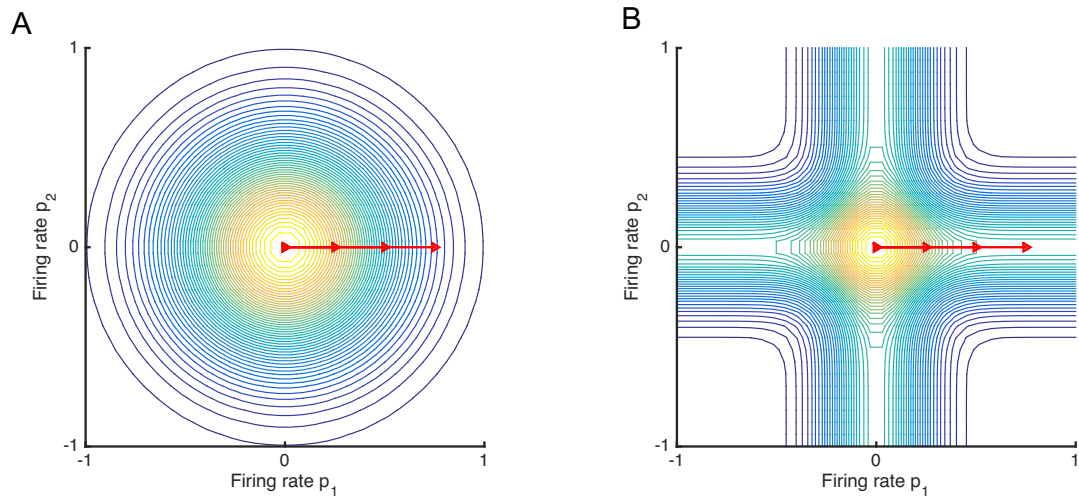

Figure 4.3 – Schematic demonstrating the effect of kernel selection on the measure of similarity for 2-dimensional neural features. Since kernel similarity between two points only depends on their coordinate-wise differences, we let $p_1 = (0, 0)$ be a point at the origin and consider the kernel-determined similarity between $p_1$ and a second point $p_2 = (x, y)$. For each plot, the color at $(x, y)$ represents the measure of similarity according to the selected kernel $K_{\hat{\theta}}(p_1, p_2)$. Traveling along the red line illustrates the effect of increasing the difference in measurements for a single neuron. For the RBF kernel (A), moving along the arrow results in the kernel becoming arbitrarily small. By contrast, the MK kernel (B) never falls below 1/2 of the value at the origin as it moves along the arrow. For 40 dimensions, the MK kernel would never fall below 39/40 of its maximal value. Hence, when the RBF kernel is used for closed-loop decoding, nonstationarities from a single neural feature would result in no similarity between the current neural feature and any of the training data. By contrast, the MK kernel will remain relatively unaffected by even a drastic change in a single neuron, and continue to effectively use the information from the remaining neurons.

### 4.7.3 Training Set Sparsification for Robustness

Training data was gathered during a standard radial center-out task during which the user attempted to move the cursor to one of eight equally-spaced targets arranged on a circle. We took the (neural,velocity)-pairs and averaged the neural data over each of the eight targets. The training set used for Gaussian process prediction consisted of these eight (average neural, velocity)-pairs. Besides making prediction much faster, we found that using this sparsified training set additionally increased decoder robustness (see Results, cf. [SG05]).

## 4.8 Experimental Methods

### 4.8.1 Permissions

The Institutional Review Boards of Brown University, Partners Health/Massachusetts General Hospital and the Providence VA Medical Center, as well as The US Food and Drug Administration granted permission for this study (Investigational Device Exemption). The participants for this study were enrolled in a pilot clinical trial of the BrainGate Neural Interface System[†].

### 4.8.2 Participant

At the time of the study, T10 was a 35 year-old man with C4 AIS-A spinal cord injury. He underwent surgical placement of two 96-channel intracortical silicon microelectrode arrays [MNN97] as previously described [Sim+11; Kim+08]. Electrodes were placed into the dominant precentral gyrus and dominant caudal middle frontal gyrus. Closed-loop recording data were used from trial (post-implant) days: 259, 265, 272, and 300.

### 4.8.3 Signal acquisition

Raw neural signals for each electrode were sampled at 30kHz using the NeuroPort System (Blackrock Microsystems, Salt Lake City, UT) and then processed using the xPC target real-time operating system (Mathworks, Natick, MA). Raw signals were downsampled to 15kHz for decoding, and then de-noised by subtracting an instantaneous common average reference [Jar+15; Gil+15] using 40 of the 96 channels on each array with the lowest root-mean-square value. The de-noised signals were band-pass filtered between 250 Hz and 5000 Hz using an 8th order non-causal Butterworth filter [Mas+15]. Spike events were triggered by crossing a threshold set at 3.5× the root-mean-square amplitude of each channel, as determined by data from a one-minute reference block at the start of each research session. The following neural features were extracted: (1) the rate of threshold crossings (not spike sorted) on each channel, and (2) the total power in the band-pass filtered signal [Jar+13; Jar+15; Bac+15; Bra+18b]. A total of M=40 features were selected. Neural features were binned in 20ms non-overlapping increments.

---

[†]ClinicalTrials.gov Identifier: NCT00912041. Caution: Investigational device. Limited by federal law to investigational use.

### 4.8.4   Decoder calibration

Task cueing was performed using custom built software running Matlab (Natick, MA). The participants used standard LCD monitors placed at 55-60 cm, at a comfortable angle and orientation. T10 engaged in the Radial-8 Task as previously described [Jar+13; Jar+15; Bac+15; Bra+18b]. Briefly, targets (size = 2.4 cm, visual angle = 2.5°) were presented sequentially in a pseudo-random order, alternating between one of eight radially distributed targets and a center target (radial target distance from center = 12.1 cm, visual angle = 12.6°). Successful target acquisition required the user to place the cursor (size = 1.5cm, visual angle = 1.6°) within the target's diameter for 300ms, before a predetermined timeout (5 seconds). Target timeouts resulted in the cursor moving directly to the intended target, with immediate presentation of the next target.

Each calibration block lasted 3 minutes. During calibration, decoder parameters were updated every 2-5 seconds as previously described [Bra+18b]. During the initial stages of calibration, we assisted cursor performance by attenuating the component of the decoded velocity perpendicular to the target [Jar+13; Vel+08]. This automated assistance was gradually decreased, until it was removed 100-130 seconds after the start of calibration. The coefficients for the MK-DKF decoder were computed using the calibration block used for the Kalman decoder.

### 4.8.5  Noise Injection Experiment

Once the decoder was calibrated, we sought to investigate the impact of nonstationarities to the MK-DKF and Kalman decoders. Our approach was to have T10 perform the Radial-8 Task, while randomly "injecting noise" to a single feature and also randomly selecting the decoder currently being used.

Each trial ended after either (1) the target was acquired by having the cursor hold within the target for 300ms, or (2) a 5 second timeout. At the start of every noise injection trial, the cursor was (1) re-centered over the previously presented target, and (2) the velocity was reset to zero (this ensured that any potential impact of cursor's behavior from the previous trial was removed). We performed block randomization of the six experimental conditions: combining one of two decoders (Kalman and MK-DKF) with one of three noise levels (no noise, 1 z-score, 5 z-scores). Both the researchers and T10 were blinded to which decoder/noise combination was currently being used. To simulate noise, we provided a z-score offset to the channel with the highest signal to noise ratio [Mal+15], based on the value computed from the calibration block. We standardized the 40 features and the noise-injected feature for both the MK-DKF and Kalman decoders. Experiments were performed in 4 minute blocks.

In order to ensure that T10 was blinded to the decoder and noise combination, we ensured that the kinematic "feel" of the decoders were similar. That is, we sought to match the mean speed, smoothing, and innovation terms for the two decoders, since these parameters are known to impact decoding quality [Wil+17]. For the Radial-8 noise-injection experiment, we matched kinematic parameters in two ways. First, we set the $A$ and $\Gamma$ of eq. (4.5) to match the Kalman values. Second, to ensure that both decoders moved at the same speed, we first computed the mean speed values for $K\zeta$ in the training block. Next, we computed the mean speed value of $f(x_t)$, and then linearly scaled $f(x_t)$ to match the mean $K\zeta$ value.

Hence, in performing head-to-head comparisons we opted to match the kinematics of the MK-DKF decoder to the Kalman. We note that it is *very likely* that we were negatively impacting the MK-DKF decoder performance by doing so, since the parameters used were likely sub-optimal compared to those that would have been computed.

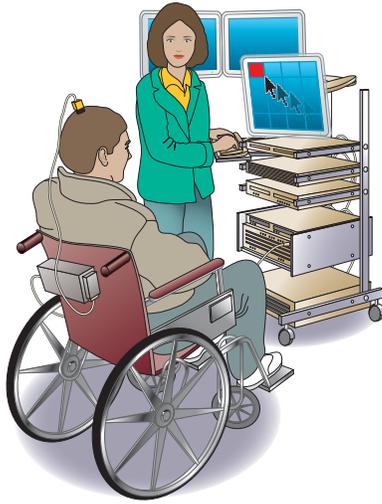

Figure 4.4 – All equipment necessary for decoding is based on a cart that can be stored in the participant's home.

### 4.8.6 Performance measurement

We quantified performance using a Grid Task after locking decoder parameters [Bra+18b; Pan+17; Nuy+15]. This task consisted of a grid of 36 square targets arranged in a square grid, where the length of one side of square grid was 24.2 cm (visual angle = 24.8°). One of 36 targets was presented at a time in a pseudo-random order. Targets were acquired when the cursor was within the area of the square for 1 second. Incorrect selections occurred if the cursor dwelled on a non-target square for an entire hold period. Each comparison block was 3 minutes in length.

We measured the achieved bit rate (BR), which measures the effective throughput of the system [Nuy+15]:

$$\text{BR} = \frac{\log_2(N-1) \ \max(S_c - S_i, 0)}{t}$$

where $N$ is the number of possible selections, $S_c$ and $S_i$ are the number of correctly and incorrectly selected targets, respectively, and $t$ is the elapsed time within the block.

### 4.8.7 Offline analysis

We retrospectively analyzed data collected from previous research sessions. We restricted our analysis to sessions where T10 moved a computer cursor using motor imagery. He acquired targets using the Radial-8 task, the Grid Task, or free typing tasks [Jar+15].

**Injecting noise for the MK-DKF and Kalman decoders**

To investigate the impact of noise on decoder performance, we performed offline simulations of both the Kalman and MK-DKF decoders. We computed the angular error between the predicted decoder value without filtering (i.e. the $Kz$ term and the $f(x_t)$ terms of the Kalman and MK decoders, respectively) and the label modeled as the vector from cursor to target [Sim+11; Bra+18b]. Data from a single research session were concatenated together. A decoder was trained using half of the data available for a session without replacement, and then used to predict the mean angular error for the other half of the dataset. Decoder predictions were bootstrapped 100 times.

**Offline assessment of noise**

As part of feature pre-processing for closed-loop decoding with the Kalman filter, we performed z-score normalization of the neural features. The incoming features are normalized using the mean and the standard deviation of the previous block's worth of data; this has the effect of increasing cursor control quality by attempting to address signal non-stationarities [Jar+15]. Our standard practice is to use a subset of the channels, selected according to a signal to noise ratio [Mal+15]. Our standard practice is to recalibrate the Kalman decoder at user-defined intervals. That is, we recompute the $K$ Kalman between blocks when the researcher is setting up the next experiment, or the user is taking a break [Jar+15].

For each session, we incrementally calibrated Kalman decoders in chronological order. We then computed the number of times a feature exceeded a z-score offset in the next block. For instance, to compute the number of noise events at 2 z-scores for Trial Day 295 Block 5, we computed the z-score mean and standard deviations based on data for Blocks 1-4, and then counted the number of 20ms blocks with deviations more than 2 z-scores away from the mean for each feature.

## 4.9 Results

### 4.9.1 Quantifying the effect of noise on closed-loop neural decoding

We investigated the impact of noise injection for both the Kalman and MK-DKF decoders by performing offline simulations of previously collected data. There were a total of 124 research sessions recorded from participant T10. We identified 97 sessions and a total of 48.2 hours of closed-loop neural control of a computer cursor, during which many variations of neural decoders had been explored. For each of the 97 sessions, we bootstrapped the data 100 times into non-overlapping training and testing sets (50/50 splits), and then used the training dataset to compute the coefficients for both the Kalman and MK-DKF decoders. We measured decoder performance using the predicted angular error between the simulated decoded direction and the known vector from cursor to target (see methods).

Our implementation of the Kalman decoder for closed-loop neural control [Jar+15; Bac+15; Bra+18b] uses a measure of signal-to-noise to sub-select 40 of the 384 features to be used in closed-loop decoding [Mal+15]. We added noise to the single feature with the highest signal to noise ratio in the testing dataset (Fig 2A). With the Kalman decoder, we found a nearly linear relationship between the amount of injected noise and the percent change in angular error ($R^2 = 0.994, p < 10^{-24}$). We then repeated this experiment using the same features for both calibration and noise injection with the MK-DKF decoder. We found that noise injection had only minimal changes to the MK-DKF performance despite large noise injection values.

Given the detrimental effect of z-score offsets on decoding performance, a straightforward solution would be to simply "saturate" the features used for decoding. That is, all values greater than a saturation value (e.g. 2 z-scores) would be set to the saturation value. We computed the change in angular error as a function of feature saturation threshold (Fig 2B). We found that the angular error decreased as saturation levels increased, reaching the base performance at 2 z-scores. These results suggested that features could be saturated at 2 z-scores without negatively impacting decoding performance.

Next, we quantified the frequency at which 2 z-score noise events occurred. Across all features, the 2 z-score deviations occurred 5.6% ±1.2 (std) of observed 20ms bins (Fig 2C). Importantly, the same features that had large noise events were those that had been highly informative and incorporated into the Kalman filter according to the feature's SNR [Jar+15; Bac+15; Bra+18b]. Since real-time neural decoding is commonly performed in 20ms bins, these results suggest that apparent noise events are observed roughly 3 times per second with our current clinical research-grade neural recording setup.

Taken together, these results suggest that: (1) the Kalman decoder is highly sensitive to z-score offsets, even arising from a single feature; (2) z-score offsets that degrade decoding performance for the Kalman occur approximately 3 times per second; (3) principled thresholding of features will alleviate some, but not all, of the effects of z-score offsets. These results also suggest that the MK-DKF is relatively insensitive to z-score offsets for single features.

### 4.9.2 Online analysis: Closed loop assessment of both the Kalman and MK-DKF decoders

We characterized the effect that noise events had on closed-loop neural decoding with T10 (Fig 3A, Supplementary Movie 1). At the start of the research session, we first calibrated both the Kalman and MK-DKF decoders, and then matched their kinematics coefficients and the subset of features used for decoding. Next, we performed a double-blinded randomization procedure where both the decoder and the amount of noise injected were randomly selected every two targets. Neither T10 nor the researchers were aware of the current decoder/noise combination. Noise was injected by offsetting the z-score of a single feature, standardized for both decoders (see Methods).

T10 was presented with a total of 596 targets in a center-out task over three research sessions (Trial Days 259, 265, and 272). For the Kalman decoder, there was a statistically significant dose dependent response between the amount of injected noise (no noise, 1 z-score offset, and 5 z-score offsets) and the fraction of targets acquired within a 5 second timeout ($\chi^2, p < 10^{-37}$). By contrast, there found no statistically significant difference between the three noise conditions with the MK-DKF decoder ($\chi^2, p = 0.81$).

We note that in this comparison, the fraction of targets acquired by the MK-DKF decoder was inferior to that of the Kalman decoder without injected noise. In order to have performed this comparison, we matched the kinematic coefficients of the MK-DKF to the Kalman decoder (see methods). This ensured that the "feel" of the decoders were indistinguishable, allowing us to perform the randomized experiment. However, in so doing, we were likely selecting sub-optimal kinematic coefficients for the decoder.

To quantify the performance of both decoders without injected noise and with optimal kinematic parameters, we calibrated the Kalman and MK-DKF decoders using their respective optimal kinematic coefficients. After decoder calibration, T10 acquired targets in the Grid Task, and the decoder being used was alternated every block (Fig 3B). We found there was no statistically significant difference in bit rate between the two decoders (Trial Days 272 and 300, Wilcoxon rank-sum test p = 0.48).

## 4.10    Discussion

A new class of neural decoder based on Gaussian process regression was more resilient to noise than the traditionally used linear decoding strategy used for closed-loop neural control. When z-score offsets were added to single channels in the Kalman filter, the decoding performance degraded; this was not seen with the MK-DKF decoding approach. After optimizing the parameterizations of both decoders, the communication bit-rate was not statistically different.

### 4.10.1    Addressing nonstationarities in neural data

Robust and reliable control with an intracortical brain computer interface is predicated on the properties of the decoding algorithm selected to map high-dimensional neural features to low-dimensional commands used to control external effectors. End-effector control degrades without recalibration of decoder parameters [Jar+15; Per+13]. To this end, multiple solutions have been proposed to recalibrate decoders based on closed-loop neural data during use, either when targets are known [Hoc+06; Kim+08; Hoc+12; Jar+13; Col+13; Wod+15; Gil+15; Ors+14; Sha+17; Dan+11; Car13] or retrospectively inferred [Jar+15]. Other approaches have investigated BCI decoder robustness using a wide variety of specific methods, including: adapting a discriminative Bayesian filter [Bra+18b], refitting a Kalman filter [Gil+12; Dan+13], Bayesian updating for an unscented Kalman filter [Li+11], reweighting a naïve Bayes classifier [Bis+14], retraining a kernelized ARMA model [SLV09], and reinforcement learning [Mah+13; Poh+14], among others.

Rather than adapting the coefficients of the decoder given new closed-loop data, the goal of robust model selection is to design the decoder to be more resilient to nonstationarities. One previously described decoder achieved robustness with a multiplicative recursive neural network and augmenting the training data with perturbations that mimicked the desired nonstationities against which they wished to train [Sus+16]. For example, in order to train against dropping the $i$th neuron, exemplars were added to the training dataset where the $i$th neuron had been zeroed-out. This technique of augmenting a training set with noisy data is well-established for increasing generalization performance in neural networks, and is commonly referred to as data augmentation [An96]. It requires generating and training over new artificial data for each individual targeted nonstationarity for each feature. Hence, exemplars generated to protect the decoder against dropping the $i$th feature do not protect against dropping the $j$th feature.

While effective, there are limitations in applying data augmentation for closed-loop BCI systems for human users. First, one of the goals of pursuing iBCI research for people is to develop devices that are intuitive and easy to use, with minimal technician oversight, and requiring calibration. It would not be possible to apply a deep neural network with bagging technique in the case where the user is using the system with limited available data, such as using the system for the first time [Bra+18b]. Second, the system requires significant computational resources. Bagging enlarges an already massive dataset by orders of magnitude, entailing a commensurate increase in training burden for the neural network. At least with today's available hardware and the requirement for local computation, the increase in computational resources would not be possible for portable iBCI systems to be used inside the homes of users.

By contrast, the MK-DKF decoder did not require explicit training to acquire robustness. The robust kernel design was able to distinguish between signal and noise within three minutes of calibration.

## 4.10.2   Growth directions for MK-DKF

Our implementation of the MK-DKF decoder provides an exciting foundation from which to explore decoder robustness. For instance, our approach naïvely provided a uniformly weighted linear addition of multiple kernels, thereby making the explicit assumption that each feature is equally important for decoding. One approach would be to incorporate techniques in kernel learning [GA11]. For instance, one could learn a convex sum of weights for the linear combination of kernels that "align" to a training kernel [CMR12]. Alternatively, one could be explore alternative distance metrics. For instance, rather than using Euclidean distances between features, one could apply a spike-train distance metric [VP97]. This metric can be adapted as a valid kernel embedding function and used for decoding neural data [Par+13; Bro+14; Li+14]. It has also been shown to perform better than Euclidean distances when visualizing complex neuronal datasets [Var+15].

## 4.11    Conclusion

In order for BCIs to succeed as an assistive technology, they will need to perform well over longer periods of time without user feedback or manual recalibration. Incorporating a robust model allows a filter to anticipate nonstationarities and seamlessly adapt to them. Here we present the first experimental evidence that a fixed, robust decoder can provide reliable and high quality neural cursor control in the presence of injected nonstationarities.

## 4.12    Acknowledgements

The authors would like to thank participant T10 and their family; B. Travers, and D. Rosler for administrative support; C. Grant for clinical assistance; and Prof. Arthur Gretton for discussions on kernel selection. This work was supported by the National Institutes of Health: National Institute on Deafness and Other Communication Disorders - NIDCD (R01DC009899), Rehabilitation Research and Development Service, Department of Veterans Affairs (B6453R and N9228C); National Science Foundation (DMS1309004), National Institute of Health (IDeA P20GM103645, R01MH102840); Massachusetts General Hospital (MGH) - Deane Institute for Integrated Research on Atrial Fibrillation and Stroke; Joseph Martin Prize for Basic Research; the Executive Committee on Research (ECOR) of Massachusetts General Hospital; Canadian Institute of Health Research (336092); Killam Trust Award Foundation; Brown Institute of Brain Science. The content of this paper is solely the responsibility of the authors and does not necessarily represent the official views of the National Institutes of Health, the Department of Veterans Affairs or the United States Government.

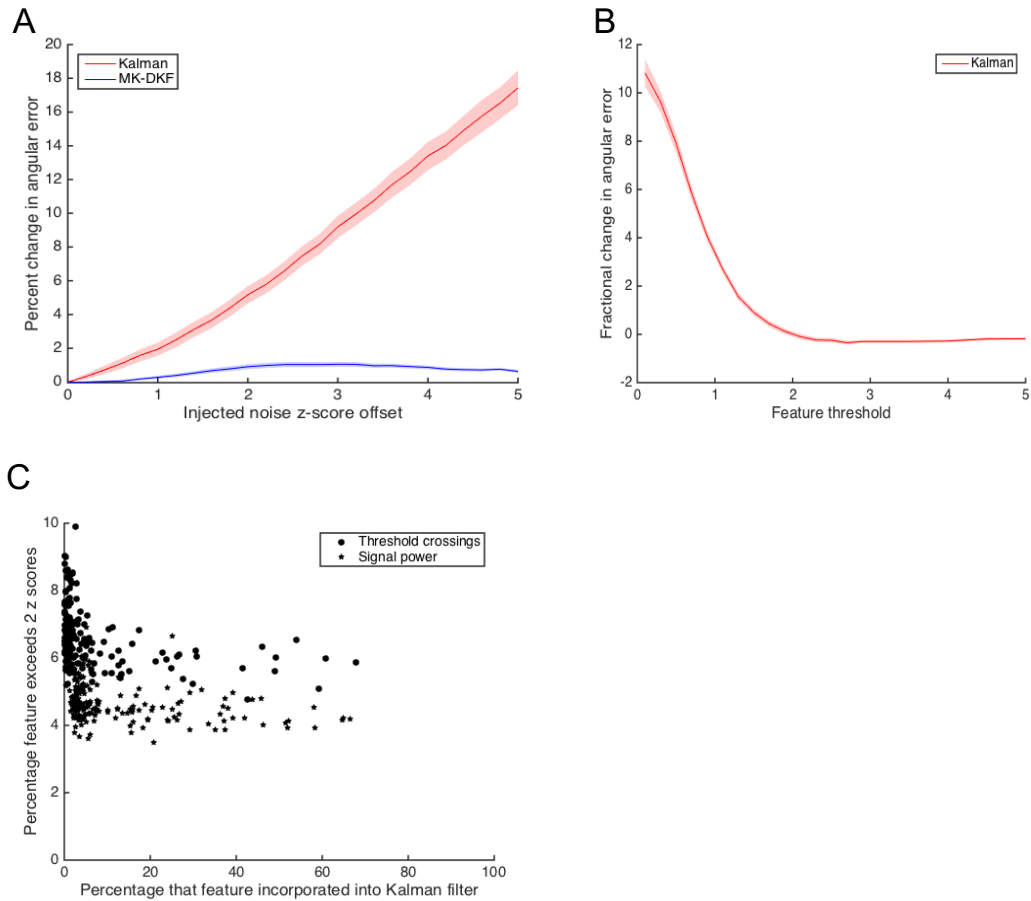

Figure 4.5 – A: Change in angular error as a function of z-score offset for both the Kalman and MK-DKF decoders. We identified 96 research sessions where T10 performed closed-loop neural control. For each session, we performed a 50/50 split, using the training data to compute the coefficients for the Kalman and MK-DKF decoders, and then predicting the angular error on the testing data. Next, we added a z-score offset to a single channel (standardized for each decoder). The shaded areas represent the standard error of measurement for each decoder. B: Change in angular error as a function of feature thresholding. During the bootstrapping procedure, we saturated features for both the training and testing datasets, and computed the change in angular error compared to no saturation. The shaded area represents the standard error of measurement. C: Examining the frequency of noise events. For each of the bootstrapped simulations, we counted the frequency at which each feature was incorporated into the decoder ($m = 40$), as well as the frequency at which the feature was observed to deviate by more than 2 z-scores.

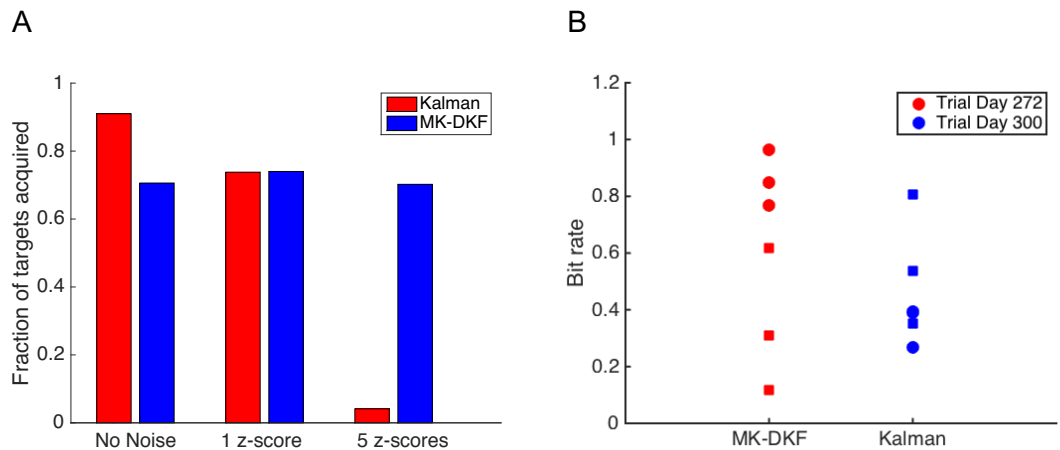

Figure 4.6 – A: Targets acquired during closed-loop Radial-8 control by T10. On research Days 259, 265, 272, and 300, T10 acquired targets in a Radial-8 task wherein the decoder (Kalman and MK-DKF) and the amount of noise (no noise, 1 z-score, 5 z-scores) were randomly selected. There was no statistically significant difference in performance across the noise injection trials for the MK-DKF decoder ($\chi^2$, $p = 0.81$) There was a statistically significant difference across conditions for the Kalman decoder ($\chi^2$, $p < 10^{-37}$). These conditions were performed with the kinematic parameters of the MK-DKF matched to the Kalman decoder. B: Performance of both the MK-DKF and Kalman decoders with optimal kinematic parameters. There was no statistically significant difference in bit rate between the two decoders (Trial Days 272 and 300, Wilcoxon rank-sum test p = 0.48).

# MORE ON DISCRIMINATIVE MODELING

> — Это водка? — слабо спросила Маргарита.
>
> Кот подпрыгнул на стуле от обиды.
>
> — Помилуйте, королева, — прохрипел он, — разве я позволил бы
>
> себе налить даме водки? Это чистый спирт!

> M. A. Bulgakov, *Master and Margarita*

## A.1 Introduction

Supervised training methods are often classified by their underlying probabilistic model. For inputs $x$ and outputs $y$, generative methods learn the joint distribution $p(x, y)$, discriminative methods learn the conditional distribution $p(y|x)$, and algorithmic methods learn the decision boundary directly.

## A.2 Nadaraya–Watson Kernel Regression

Given a dataset $\{(x_i, z_i)\}_{i=1}^n$, we can use Kernel Density Estimation (KDE) to model the joint density:

$$p(z, x) \approx \frac{1}{n} \sum_{i=1}^n \kappa_Z(z, Z_i) \kappa_X(x, X_i).$$

where $\kappa_Z(z, z')$ and $\kappa_X(x, x')$ are kernel functions (symmetric, positive-definite). It follows that the conditional distribution is then given by

$$p(z|x) = \frac{p(z, x)}{p(x)} \approx \frac{\sum_{i=1}^m \kappa_Z(z, Z_i) \kappa_X(x, X_i)}{\sum_{i=1}^m \kappa_X(x, X_i)}. \tag{A.1}$$

---

"Is that vodka?" asked Margarita weakly.

The cat took offense and jumped up on his chair.

"Excuse me, Your Majesty," he whined, "but how could I offer vodka to a lady? It's pure spirit!"

We then estimate $f(x) := \mathbb{E}[Z|X = x]$ as

$$f(x) = \int zp(z|x)dz \approx \frac{\sum_{i=1}^{m} \left( \int z\kappa_Z(z, Z_i)dz \right) \kappa_X(x, X_i)}{\sum_{i=1}^{m} \kappa_X(x, X_i)} \tag{A.2}$$

Because the kernel is symmetric, $\int z\kappa_Z(z, z')dz = z'$ so that

$$f(x) \approx \frac{\sum_{i=1}^{m} Z_i\kappa_X(x, X_i)}{\sum_{i=1}^{m} \kappa_X(x, X_i)} \tag{A.3}$$

This is the well-known Nadaraya-Watson kernel regression estimator [Nad64; Wat64]. In the same vein, we may estimate

$$\mathbb{E}[Z^\intercal Z|X = x] \approx \frac{\sum_{i=1}^{m} \left( \int z^\intercal z\kappa_Z(z, Z_i)dz \right) \kappa_X(x, X_i)}{\sum_{i=1}^{m} \kappa_X(x, X_i)} \tag{A.4}$$

If the kernel is both symmetric and stationary, so that $\int z^\intercal z\kappa_Z(z, z')dz = \Sigma_Z + (z')^\intercal z'$ for some fixed $\Sigma_Z \in \mathbb{S}_d$, then

$$\mathbb{E}[Z^\intercal Z|X = x] \approx \Sigma_Z + \frac{\sum_{i=1}^{m} Z_i^\intercal Z_i\kappa_X(x, X_i)}{\sum_{i=1}^{m} \kappa_X(x, X_i)} \tag{A.5}$$

If Equations A.3 and A.5 hold, we can then form an estimate for $Q(x) := \mathbb{V}[Z|X = x]$ as follows

$$Q(x) \approx \Sigma_Z + \frac{\sum_{i=1}^{m} Z_i^\intercal Z_i\kappa_X(x, X_i)}{\sum_{i=1}^{m} \kappa_X(x, X_i)} - (f(x))^\intercal f(x) \tag{A.6}$$

### A.2.1 Learning

Learning kernel bandwidth parameters can be done via cross-validation or a rule of thumb (e.g. Silverman's rule).

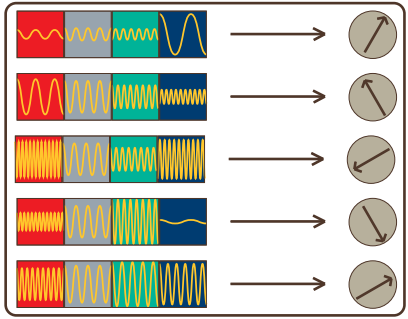

Figure A.1 – The Nadaraya–Watson kernel regression estimate takes a training dataset $\mathcal{D}$ (left) and for a testpoint $x_*$ returns (up to renormalization) an estimate for the corresponding $z_*$-value that is the weighted sum of the training values, where the weights for each $z_i$ are determined by how close the corresponding $x_i$ is to $x_*$ (as determined by the kernel):

## A.3 Neural Network Regression with Homoskedastic Gaussian Noise

Given a dataset $\{(x_i, y_i)\}_{i=1}^{n}$, we can model

$$Y_i = f(X_i) + \epsilon_i \tag{A.7}$$

where $\epsilon_i \sim^{\text{i.i.d.}} \mathcal{N}(\vec{0}, \Gamma)$ and $f$ is some unknown nonlinear function. We can learn $f$ from training data as a neural network. The idea is to form f by recursively composing a nonlinear activation function $a$ with linear combinations of inputs

$$f_1(x) = a(L_0 x) \tag{A.8}$$

$$f_2(x) = a(L_1 f_1(x)) \tag{A.9}$$

$$\vdots \tag{A.10}$$

$$f_m(x) = L_m a(L_{m-1} f_{m-1}(x)) \tag{A.11}$$

where at each step, the input is first multiplied by some matrix $L_i$ and then the function $a$ is applied to each coordinate of the output.

### A.3.1 Learning

Training a neural network then corresponds to solving the optimization problem:

$$\hat{L}_0, \dots, \hat{L}_m = \underset{L_0,\dots,L_m}{\arg\min} \|Y_{1:n} - f_m(X_{1:n})\| \tag{A.12}$$

Any functional relationship can be approximated arbitrarily well (if $m \geq 1$, i.e. if there is at least one hidden layer) by the above approach as the size of $L_0, L_1$ become arbitrarily large; this is known as the Universal Approximation Theorem [Hor91].

### A.3.2 Issues and Innovations

Choosing the architecture; i.e. the sizes for the $L_i$ (sizes of the hidden layers) and the nature of the activation functions (max-pooling, dropout, convolution, batch normalization) is amazingly challenging and oftentimes done painstakingly by hand [Rea+17; ZL17].

## A.4 Gaussian Process Regression

Given a dataset $\{(x_i, y_i)\}_{i=1}^n$, a Gaussian process regression models the random variables $Y_i$ as joint Gaussian with covariance

$$\mathrm{cov}(Y_i, Y_j) = \kappa_\theta(x_i, x_j) \tag{A.13}$$

where $\kappa$ is a kernel function (symmetric, positive-definite) dependent on a set $\theta$ of tunable hyperparameters. The model is then

$$\begin{bmatrix} Y_{1:n} \\ Y_* \end{bmatrix} \sim \mathcal{N}\left(\vec{0}, \begin{bmatrix} \kappa_\theta(x_{1:n}, x_{1:n}) & \kappa_\theta(x_{1:n}, x_*) \\ \kappa_\theta(x_*, x_{1:n}) & \kappa_\theta(x_*, x_*) \end{bmatrix}\right) \tag{A.14}$$

where $(x_*, Y_*)$ denotes the random value of $Y_*$ at a deterministic point $x_*$. We let $K := \kappa_\theta(x_{1:n}, x_{1:n})$ denote the matrix given by $K_{ij} = \kappa_\theta(x_i, x_j)$. Under this notation, we have also $K_* := \kappa_\theta(x_{1:n}, x_*)$ and $K_{**} := \kappa_\theta(x_*, x_*)$, so we may re-write eq. (A.14) as

$$\begin{bmatrix} Y_{1:n} \\ Y_* \end{bmatrix} \sim \mathcal{N}\left(\vec{0}, \begin{bmatrix} K & K_* \\ K_*^\mathsf{T} & K_{**} \end{bmatrix}\right) \tag{A.15}$$

This implies

$$Y_* | x_{1:n}, Y_{1:n}, x_* \sim \mathcal{N}(K_*^\intercal K^{-1} Y_{1:n}, K_{**} - K_*^\intercal K^{-1} K_*) \tag{A.16}$$

When the model specifies noisy observations, so that eq. (A.13) becomes

$$\mathrm{cov}(Y_i, Y_j) = \kappa_\theta(x_i, x_j) + \sigma^2 \delta_{\{i=j\}} \tag{A.17}$$

the inference eq. (A.16) becomes

$$Y_* | x_{1:n}, Y_{1:n}, x_* \sim \mathcal{N}(K_*^\intercal (K + \sigma^2 I)^{-1} Y_{1:n}, K_{**} - K_*^\intercal (K + \sigma^2 I)^{-1} K_*) \tag{A.18}$$

where $I = I_n$ denotes the $n$-dimensional identity matrix. It's worth noting that the mean function for a Gaussian process is a linear combination of kernel functions, i.e.

$$\mathbb{E}[Y_* | x_{1:n}, Y_{1:n}, x_*] = \sum_{i=1}^{n} \alpha_i \kappa_\theta(x_i, x_*) \tag{A.19}$$

where $\alpha = (K + \sigma^2 I)^{-1} Y_{1:n}$.

### A.4.1  Learning

To learn the model eq. (A.18) (which reduces to eq. (A.16) when $\sigma^2 = 0$), we usually consider the log-likelihood of the data

$$\log p(Y_{1:n} | x_{1:n}) = -\frac{1}{2} Y_{1:n}^\intercal (K + \sigma^2 I)^{-1} Y_{1:n} - \frac{1}{2} \log \left| K + \sigma^2 I \right| - \frac{n}{2} \log 2\pi \tag{A.20}$$

The most common method of choosing hypeparameters is by maximizing the MLE:

$$\hat{\theta}_{MLE}, \hat{\sigma}_{MLE} = \arg\max_{\theta, \sigma} \log p(Y_{1:n} | x_{1:n}) \tag{A.21}$$

where the dependence on $\theta$ occurs through the matrix $K$. See [RW06] for other hyperparameter learning methods.

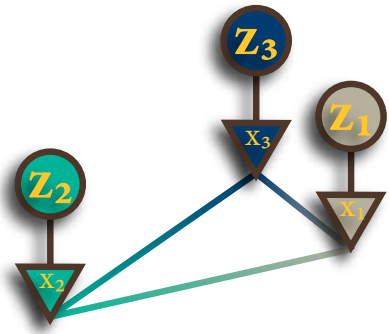

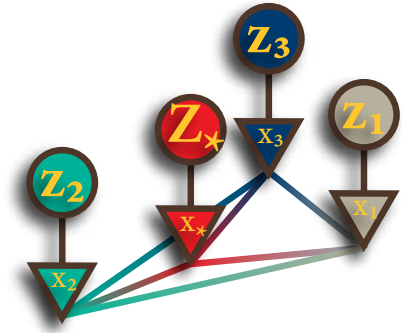

Figure A.2 – GP model. Given a dataset $\mathcal{D} = \{(x_i, z_i)\}_{i=1}^{n}$, the GP model specifies that

$$\begin{bmatrix} z_1 \\ z_2 \\ z_3 \end{bmatrix} \sim \mathcal{N}\left( \begin{bmatrix} 0 \\ 0 \\ 0 \end{bmatrix}, \begin{bmatrix} k_{11} & k_{12} & k_{13} \\ k_{21} & k_{22} & k_{23} \\ k_{31} & k_{32} & k_{33} \end{bmatrix} + \sigma^2 I \right)$$

where $k_{ij} = k_\theta(x_i, x_j)$ and $\sigma^2$ accounts for noisy measurements. The model specifies that more similar $x$-values (as determined by the kernel) should correspond to more correlated $z$-values.

Figure A.3 – GP Inference. Given some test value $x_*$, we now have

$$\begin{bmatrix} \top \\ z \\ \bot \\ Z_* \end{bmatrix} \sim \mathcal{N}\left( \begin{bmatrix} \top \\ 0 \\ \bot \\ 0 \end{bmatrix}, \begin{bmatrix} \ulcorner & & \urcorner & \top \\ & K + \sigma^2 I & & k_* \\ \llcorner & & \lrcorner & \bot \\ \vdash & k_*^\top & \dashv & k_{**} \end{bmatrix} \right).$$

From this, $Z_* | \{X_* = x_*, \mathcal{D}\}$ can be calculated. This is the prediction given in eq. (A.16).

### A.4.2 Issues and Innovations

GP prediction scales $O(n^3)$ in computational cost with the size of the training set because the full covariance matrix $K$ must be inverted. The issue of requiring the entire training dataset in order to perform prediction proves a particular downside to nonparametric models, including GP's. (We can contrast this to, e.g. linear regression, where the learned coefficients are sufficient for prediction.) To this end, some researchers asked *could one replace the original training set with a much smaller (optimized) pseudo-training set, with only minimal alteration to the predictions?* From the different ways to measure alteration and perform optimization arise a family of methods known collectively as sparse Gaussian processes [QR05]. The optimized pseudo-training points are variously referred to as pseudo-inputs, inducing points, or support points. It's possible the sparse GP's may provide other incidental benefits with respect to robustness: [SG05] remarked that

> although the [sparse GP] is not specifically designed to model nonstationarity, the extra flexibility associated with moving pseudo-inputs around can actually achieve this to a certain extent.

## A.5   Random Forests

We can also approximate $p(z_t|x_t)$ as a random forest, one type of ensemble method. The idea behind ensemble methods is to agglomerate a set of weak learners (learners that may be only slightly better than chance) into a strong learner (one that becomes arbitrarily good as the number of its constituents grows). In the case of random forests, the weak learner is a regression tree. A regression tree partitions the $x_t$ space and assigns a constant value to each of the partition members.

$$f_j(x) = \sum_{i=1}^{N_j} c_i^j \mathbb{1}_{C_i^j}(x) \tag{A.22}$$

Numerous algorithms exist to optimize these choices. A regression forest then combines many of these trees $f_1, \ldots, f_T$ into a single estimator. Under the model for $Z|X$ that first draws a tree randomly from $\{1, \ldots, T\}$ and then assigns the estimate from that tree, we have the model

$$p(z|x) = \begin{cases} 1/N_j, & z = c_{\{i:x \in C_i^j\}}^j \\ 0, & \text{otherwise} \end{cases} \tag{A.23}$$

Under this model it follows that

$$\mathbb{E}[Z|X = x] = \frac{1}{N_j} \sum_{j=1}^{N_j} c_{\{i:x \in C_i^j\}}^j = \frac{1}{N_j} \sum_{j=1}^{N_j} f_j(x) \tag{A.24}$$

and

$$\mathbb{E}[ZZ^\intercal|X = x] = \frac{1}{N_j} \sum_{j=1}^{N_j} f_j(x)(f_j(x))^\intercal \tag{A.25}$$

# REFERENCES

*«σὺ δὲ ὅταν προσεύχῃ, εἴσελθε εἰς τὸ ταμεῖόν σου καὶ κλείσας τὴν*
*θύραν σου πρόσευξαι τῷ πατρί σου τῷ ἐν τῷ κρυπτῷ· καὶ ὁ πατήρ*
*σου ὁ βλέπων ἐν τῷ κρυπτῷ ἀποδώσει σοι.»*

Matthew 6:6

---

"But you, when you pray, enter into your inner chamber, and having shut your door, pray to your Father who is in secret, and your Father who sees in secret will reward you openly."

# INDEX